%% file: neurips_2026.tex
\PassOptionsToPackage{table,xcdraw}{xcolor}
\documentclass[10pt,logo,copyright]{nvidiatechreport}

\usepackage[utf8]{inputenc} 
\usepackage[T1]{fontenc}    
\usepackage{hyperref}       
\usepackage{url}            
\usepackage{booktabs}       
\usepackage{amsfonts}       
\usepackage{nicefrac}       
\usepackage{microtype}      
\usepackage{xcolor}         
\usepackage{pifont}
\usepackage{multirow}
\usepackage{amsmath}
\usepackage{listings}
\usepackage[utf8]{inputenc} 
\usepackage{subcaption}
\usepackage{graphicx}
\usepackage{tabularx}
\usepackage{array}
\usepackage{makecell}
\usepackage[export]{adjustbox}
\newcolumntype{L}[1]{>{\raggedright\arraybackslash}m{#1}}
\newcolumntype{C}[1]{>{\centering\arraybackslash}m{#1}}

\usepackage{hyperref}
\usepackage{cleveref}
\usepackage{wrapfig}
\usepackage{float}

\usepackage{longtable}
\usepackage{enumitem}

\title{APE: Agentic Prompt Enhancer for Image Generation and Editing}

\author{%
Zijian Huang\textsuperscript{1,2} \quad 
Jay Zhangjie Wu\textsuperscript{1} \quad 
Zian Wang\textsuperscript{1} \quad 
Tianshi Cao\textsuperscript{1} \quad 
Jiasi Chen\textsuperscript{2} \quad 
Sanja Fidler\textsuperscript{1} \quad 
Huan Ling\textsuperscript{1} \newline 
Xuanchi Ren\textsuperscript{1} \\
\small \textsuperscript{1}NVIDIA, \textsuperscript{2}University of Michigan\\
\small \href{https://research.nvidia.com/labs/sil/projects/ape/}{https://research.nvidia.com/labs/sil/projects/ape/}
}

\begin{document}

\maketitle
\suppressfloats[t] 

\begin{abstract}
Natural language has become a powerful interface for image generation and editing, yet text-guided visual systems remain highly sensitive to prompt formulation. Semantically similar requests can produce different outputs depending on wording, specificity, and how explicitly visual constraints are stated, motivating prompt enhancement as a trainable component rather than a peripheral user choice. Existing strong enhancers often rely on large, proprietary LLMs such as ChatGPT or Gemini, adding cost, latency, and deployment dependence to the visual generation pipeline. We propose \textit{Agentic Prompt Enhancer (APE)}, a lightweight framework that post-trains small language models (SLMs) as prompt-enhancement agents. APE supports both single-agent rewriting and role-specialized multi-agent enhancement. Its single-agent instantiation, \textit{SAPE}, rewrites the prompt in one pass, while its multi-agent instantiation, \textit{MAPE}, decomposes enhancement into a \textit{router--rewriter--composer} process for handling compositional constraints over objects, attributes, spatial relations, and edits. With task-aware rewards and post-training protocols, APE improves visual alignment and prompt following without modifying the downstream visual model. 
Experiments on challenging image generation and editing benchmarks demonstrate that post-trained small prompt enhancers reliably outperform their base counterparts, narrowing the gap to closed-source prompt enhancers; in addition, MAPE proves particularly strong on complex compositional tasks within these benchmarks.
\end{abstract}

\vspace{-16pt}
\input{qualitative_mape}

\vspace{2pt}
\needspace{0.32\textheight}
\abscontent

\input{intro}

\input{related}

\input{method}

\input{experiment}

\input{conclusion}


\appendix
\clearpage
\section*{Appendix}

\input{appendix}


\bibliography{ref}
\bibliographystyle{plain}

\end{document}

%% file: qualitative_mape.tex
%

\definecolor{apeSoft}{HTML}{F5F7F0}
\definecolor{apeCard}{HTML}{FFFFFF}
\definecolor{apeQwenBg}{HTML}{F3F4F1}
\definecolor{apeMapeBg}{HTML}{EEF6E3}
\definecolor{apePromptBg}{HTML}{FAFBF7}
\definecolor{apeLine}{HTML}{C7D6AE}
\definecolor{apeMuted}{HTML}{58614C}
\definecolor{apeDark}{HTML}{151A11}

\newcommand{\mapegenimg}[2][\linewidth]{%
    \adjustbox{cfbox=black!28 0.4pt 0pt}{\includegraphics[width=#1]{#2}}%
}

\newcommand{\apepromptfont}{\fontsize{4.8}{4.75}\selectfont}
\newcommand{\apehl}[1]{{\fontsize{5.45}{4.6}\selectfont\bfseries #1}}
\newcommand{\apeorigfont}{\fontsize{5.8}{6.5}\selectfont}

\newcommand{\apemethodpanel}[4]{%
    \begingroup
    \setlength{\fboxsep}{2.1pt}%
    \noindent\colorbox{#4}{%
        \begin{minipage}{\dimexpr\linewidth-2\fboxsep\relax}
            {\scriptsize\bfseries\textcolor{apeDark}{#1}}\par\vspace{1.5pt}
            \begin{minipage}[t]{0.36\linewidth}\vspace{0pt}%
                \centering
                \mapegenimg[0.99\linewidth]{#2}
            \end{minipage}\hfill
            \begin{minipage}[t]{0.615\linewidth}\vspace{0pt}%
                {\scriptsize\bfseries\textcolor{apeMuted}{Enhanced prompt}}\par
                {\apepromptfont\color{apeDark}#3}
            \end{minipage}%
        \end{minipage}%
    }%
    \endgroup
}

\newcommand{\apemethodpanelwide}[4]{%
    \begingroup
    \setlength{\fboxsep}{2.1pt}%
    \noindent\colorbox{#4}{%
        \begin{minipage}{\dimexpr\linewidth-2\fboxsep\relax}
            {\scriptsize\bfseries\textcolor{apeDark}{#1}}\par\vspace{1.5pt}
            \begin{minipage}[t]{0.535\linewidth}\vspace{0pt}%
                \centering
                \mapegenimg[0.995\linewidth]{#2}
            \end{minipage}\hfill
            \begin{minipage}[t]{0.44\linewidth}\vspace{0pt}%
                {\scriptsize\bfseries\textcolor{apeMuted}{Enhanced prompt}}\par
                {\apepromptfont\color{apeDark}#3}
            \end{minipage}%
        \end{minipage}%
    }%
    \endgroup
}

\newcommand{\apepromptcase}[6]{%
    \par\noindent
    \begingroup
    \setlength{\fboxsep}{3.2pt}%
    \setlength{\fboxrule}{0.4pt}%
    \fcolorbox{apeLine}{apeCard}{%
        \begin{minipage}{\dimexpr\linewidth-2\fboxsep-2\fboxrule\relax}
            {\footnotesize\bfseries\textcolor{apeDark}{#1}}\par\vspace{1pt}
            \begingroup
            \setlength{\fboxsep}{3pt}%
            \noindent\colorbox{apePromptBg}{%
                \begin{minipage}{\dimexpr\linewidth-2\fboxsep\relax}
                    {\scriptsize\bfseries\textcolor{apeMuted}{Original user prompt}}\par
                    {\apeorigfont\itshape\color{apeDark}#2}
                \end{minipage}%
            }%
            \endgroup\par\vspace{1.6pt}
            \begin{minipage}[t]{0.486\linewidth}\vspace{0pt}%
                \apemethodpanel{Qwen3-4B}{#3}{#4}{apeQwenBg}
            \end{minipage}\hfill
            \begin{minipage}[t]{0.486\linewidth}\vspace{0pt}%
                \apemethodpanel{MAPE}{#5}{#6}{apeMapeBg}
            \end{minipage}
        \end{minipage}%
    }%
    \endgroup\par\vspace{2pt}
}

\newcommand{\apepromptcasewide}[6]{%
    \par\noindent
    \begingroup
    \setlength{\fboxsep}{3.2pt}%
    \setlength{\fboxrule}{0.4pt}%
    \fcolorbox{apeLine}{apeCard}{%
        \begin{minipage}{\dimexpr\linewidth-2\fboxsep-2\fboxrule\relax}
            {\footnotesize\bfseries\textcolor{apeDark}{#1}}\par\vspace{1pt}
            \begingroup
            \setlength{\fboxsep}{3pt}%
            \noindent\colorbox{apePromptBg}{%
                \begin{minipage}{\dimexpr\linewidth-2\fboxsep\relax}
                    {\scriptsize\bfseries\textcolor{apeMuted}{Original user prompt}}\par
                    {\apeorigfont\itshape\color{apeDark}#2}
                \end{minipage}%
            }%
            \endgroup\par\vspace{1.6pt}
            \begin{minipage}[t]{0.486\linewidth}\vspace{0pt}%
                \apemethodpanelwide{Qwen3-4B}{#3}{#4}{apeQwenBg}
            \end{minipage}\hfill
            \begin{minipage}[t]{0.486\linewidth}\vspace{0pt}%
                \apemethodpanelwide{MAPE}{#5}{#6}{apeMapeBg}
            \end{minipage}
        \end{minipage}%
    }%
    \endgroup\par\vspace{2pt}
}

\begingroup
\captionsetup{font=scriptsize,skip=1pt}
\setlength{\fboxsep}{3.6pt}
\setlength{\fboxrule}{0.6pt}
\noindent\fcolorbox{apeLine}{apeSoft}{%
    \begin{minipage}{\dimexpr\linewidth-2\fboxsep-2\fboxrule\relax}
        \centering
        {\large\bfseries\textcolor{apeDark}{Agentic prompt enhancement turns vague intent into grounded visual constraints}}\par\vspace{1pt}
        {\footnotesize\textcolor{apeMuted}{Each example keeps the enhanced prompt used to generate the image; \textbf{bolded} MAPE text marks the most informative additions.}}\par\vspace{1pt}
        \captionof{figure}{Prompt-aware qualitative teaser comparing one-shot Qwen3-4B with \textbf{MAPE (Qwen3-4B)} under the same image model.}
        \label{fig:qual_samples_mape}
    \end{minipage}%
}
\endgroup
\par\vspace{2pt}

\apepromptcase
    {Relational Detail}
    {The father and son, wearing the same plaid pajamas, slept in the same position on the sofa, looking peaceful.}
    {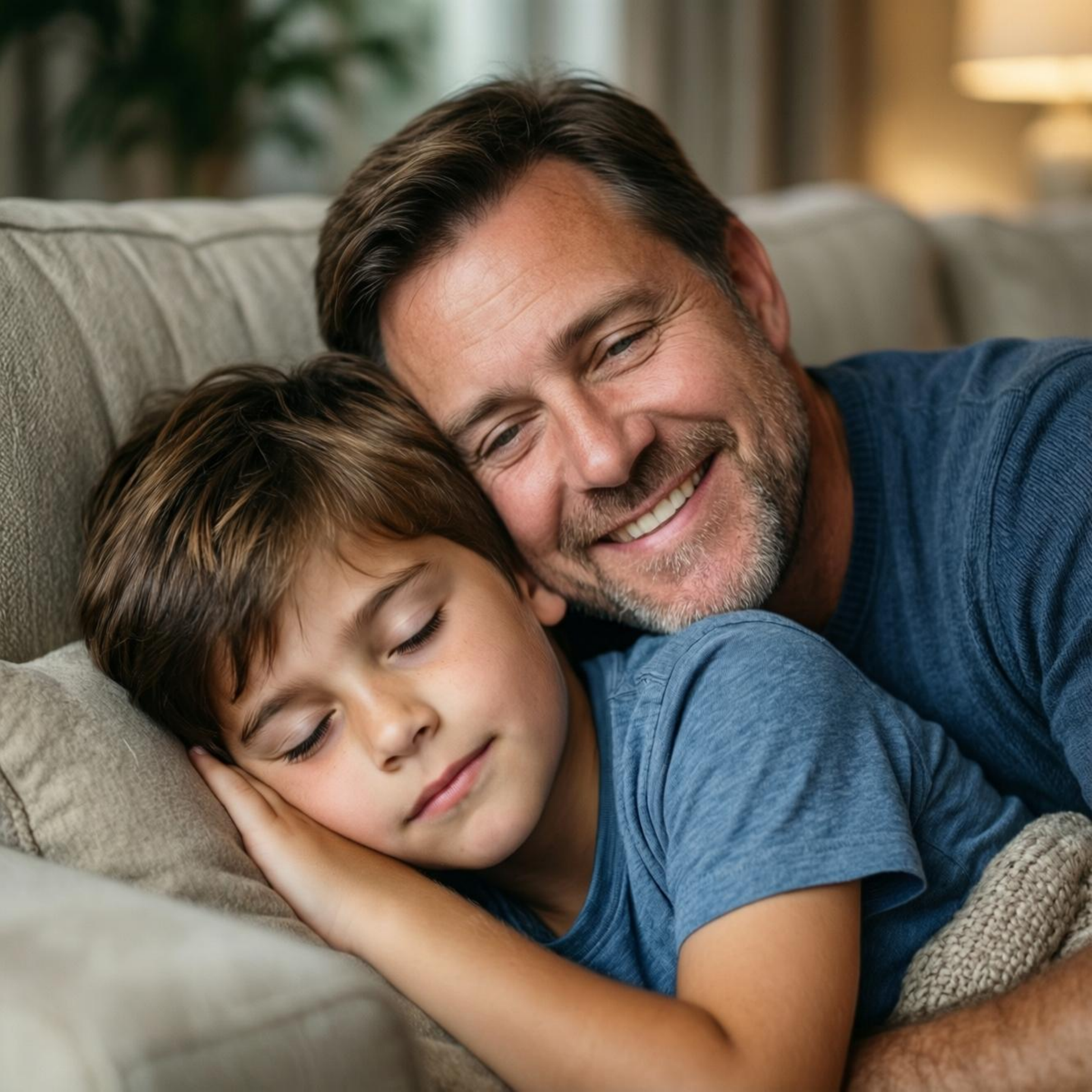}
    {A photorealistic portrait capturing a heartwarming moment of a middle-aged Caucasian father and his young adult son sleeping soundly on a plush beige sofa. Ultra HD, 4K, cinematic composition.}
    {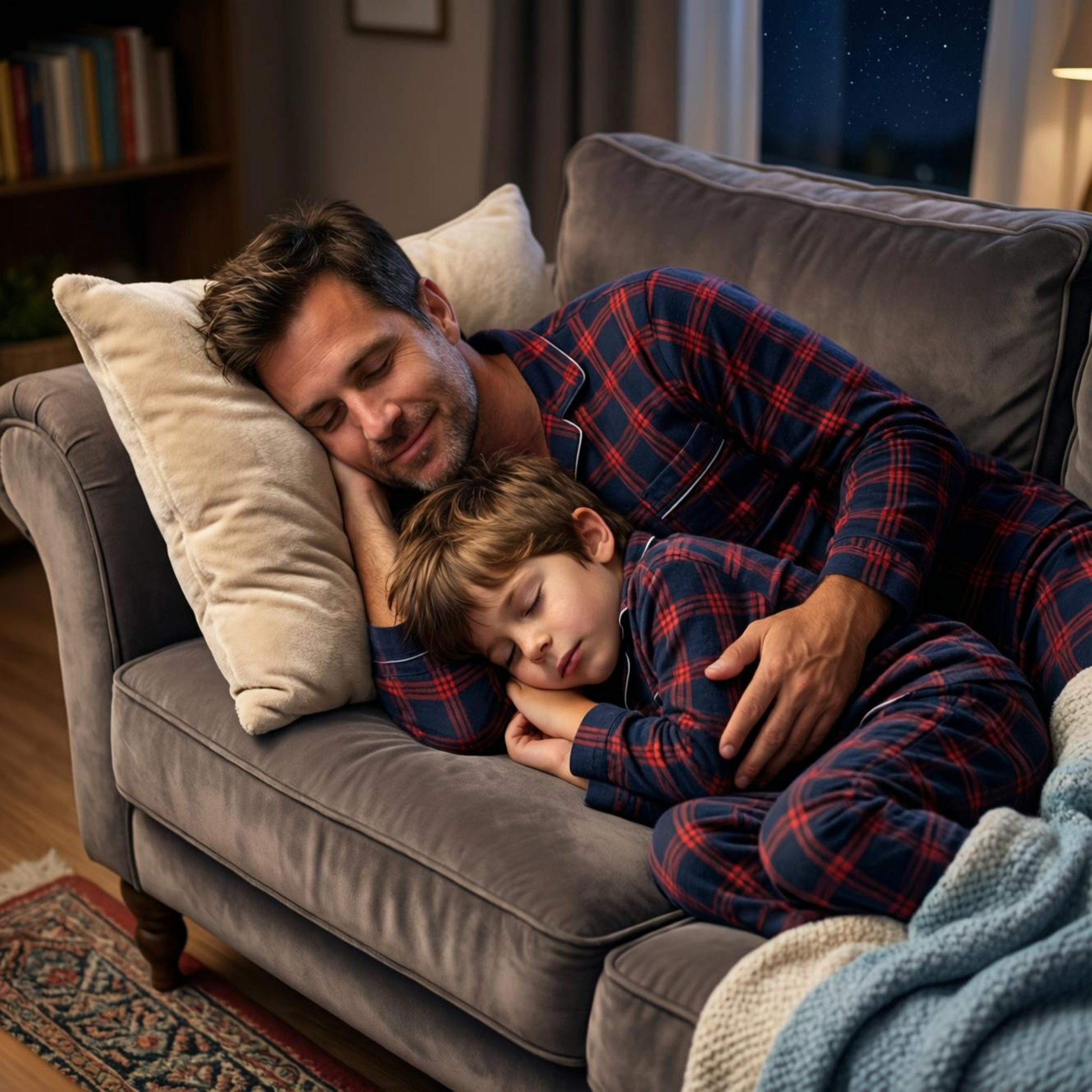}
    {A heartwarming, photorealistic medium shot of a father and his young son fast asleep on a plush, overstuffed velvet sofa, \apehl{their matching navy and red plaid pajama sets---complete with button-down shirts and comfortable pants---wrapping them in a cocoon of warmth.} The father, in his 30s or early 40s, lies center-left with a serene, slightly smiling expression, his arm loosely draped over the son in protection. The son, 5--7 years old, is curled against the father's chest in the same fetal position, eyes tightly shut. Soft ambient lighting from a nearby lamp casts gentle, elongated shadows across the cushions.}

\apepromptcase
    {Object Fidelity}
    {A Shiba Inu with a square watermelon body, its limbs are green vines, and its head is curiously tilted, playing with a trendy handmade style.}
    {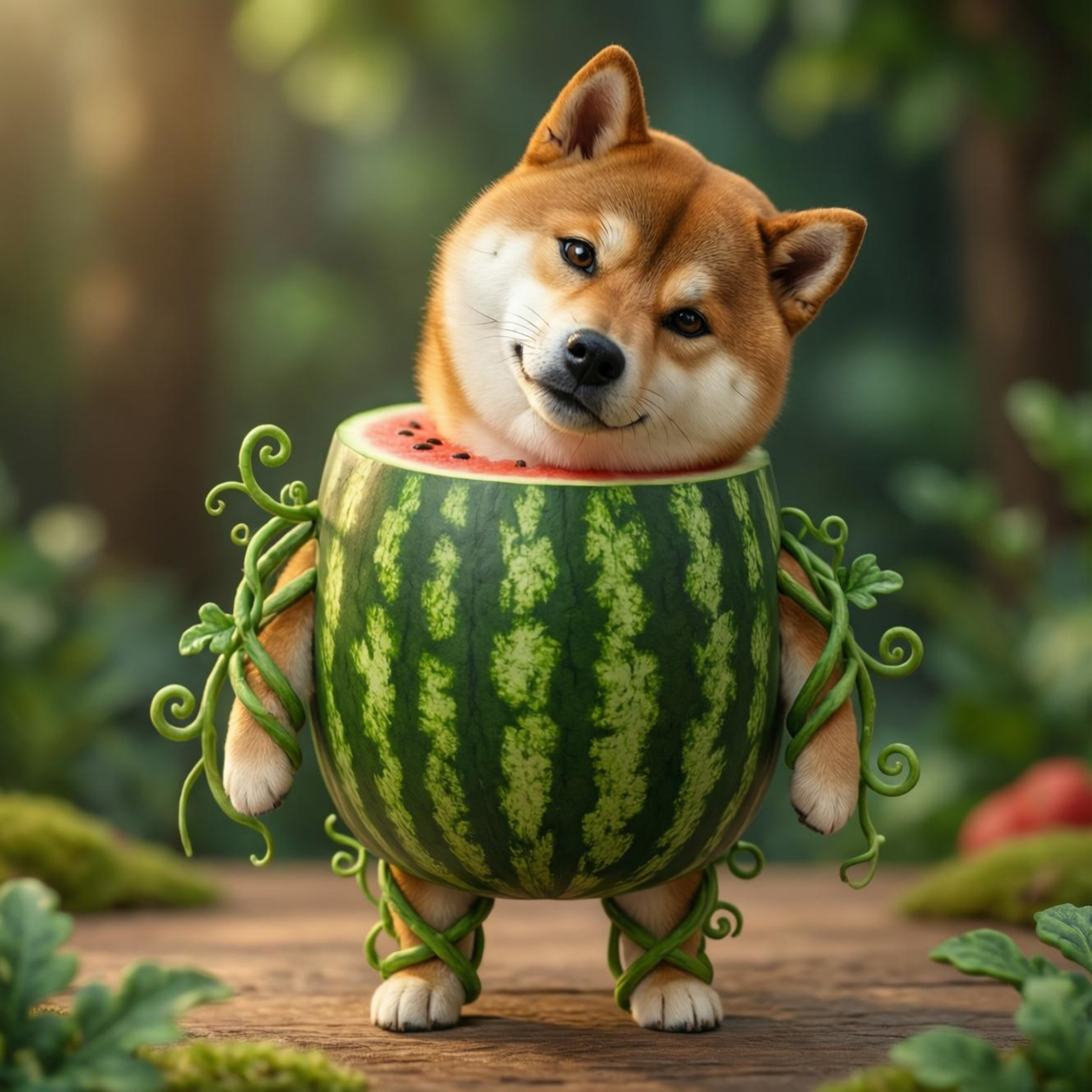}
    {A charming anthropomorphic Shiba Inu character stands with a square body made from a textured watermelon rind, its limbs composed of vibrant green vines curling naturally, head curiously tilted. The artistic style is a trendy handmade aesthetic resembling stop-motion claymation with soft, warm lighting. Ultra HD, 4K, cinematic composition.}
    {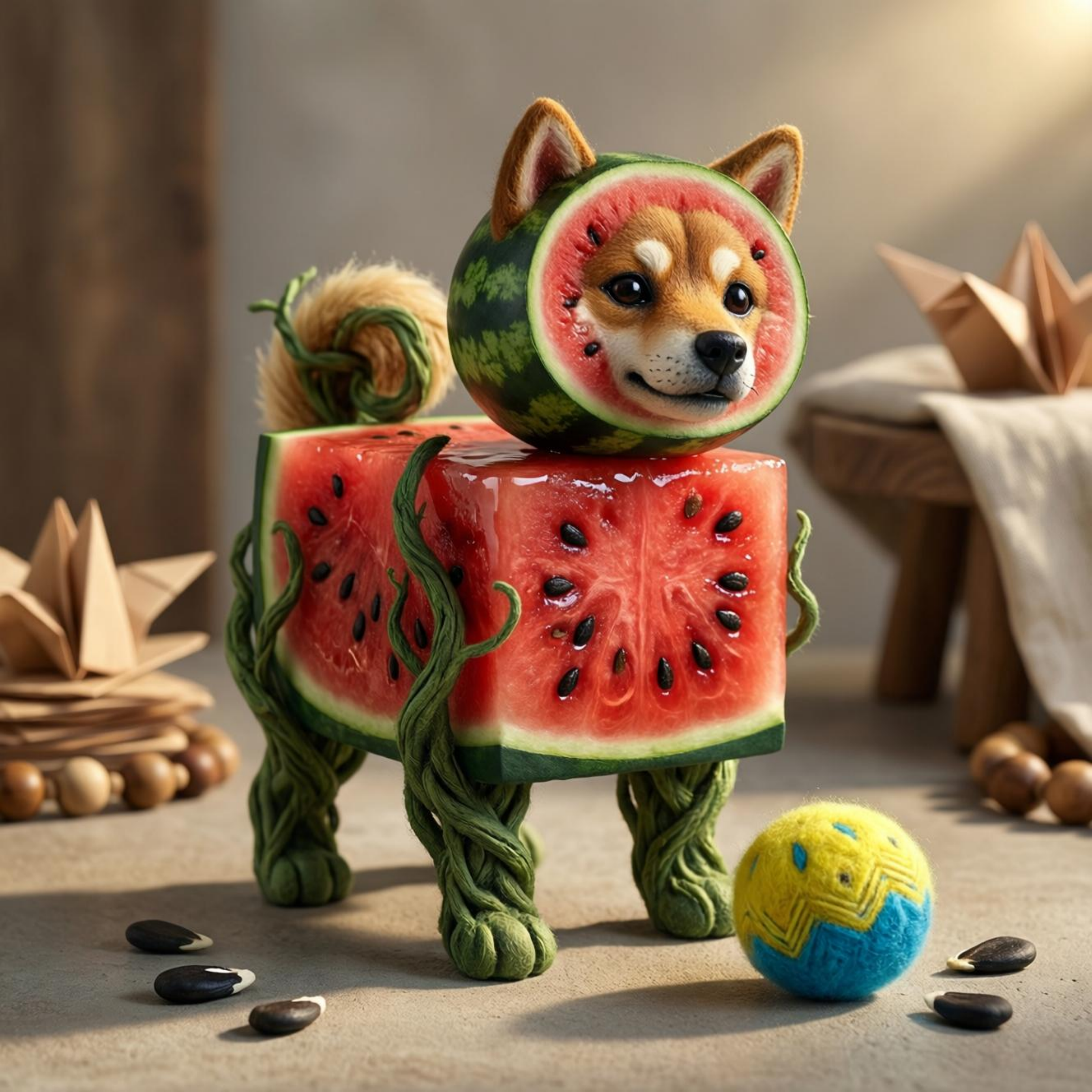}
    {A whimsical, handmade-style sculpture of a Shiba Inu stands center stage. \apehl{Its blocky, square torso is a vibrant red watermelon with a dark green rind, the flesh glistening with a glossy sheen while the vine-like limbs twist upward in organic, textured green tendrils resembling dried grapevine, their fibrous surfaces slightly bumpy and matte, supporting the heavy watermelon body.} The dog's head is a rounded watermelon slice, integrated into the torso, with carved black seeds for eyes and a stitched mouth, tilted curiously toward a small, fuzzy felt fidget toy in the foreground. Stray watermelon seeds and elongated shadows from warm, diffused studio lighting complete the scene.}

\apepromptcasewide
    {Text and Target}
    {An anthropomorphic cactus astronaut, floating in space, pointing to the distant earth, with the English words ``Home is where your heart is planted'' written on his helmet.}
    {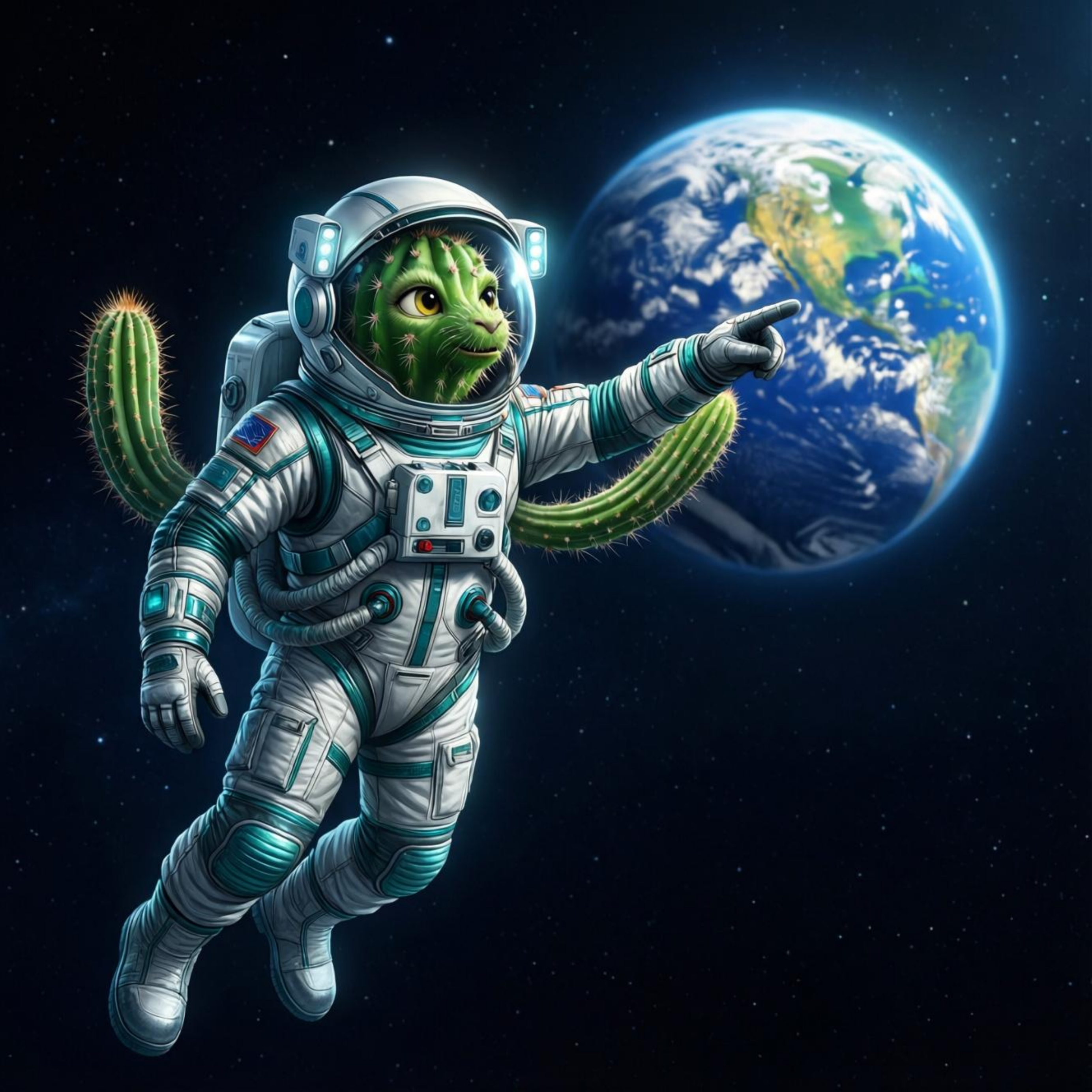}
    {A cinematic digital illustration of an anthropomorphic green cactus astronaut floating in the vast darkness of deep space, wearing a sleek white and metallic teal space suit. He has a humanoid stance with arms extended, pointing one long green arm toward the distant blue and white planet Earth. Ultra HD, 4K, cinematic composition.}
    {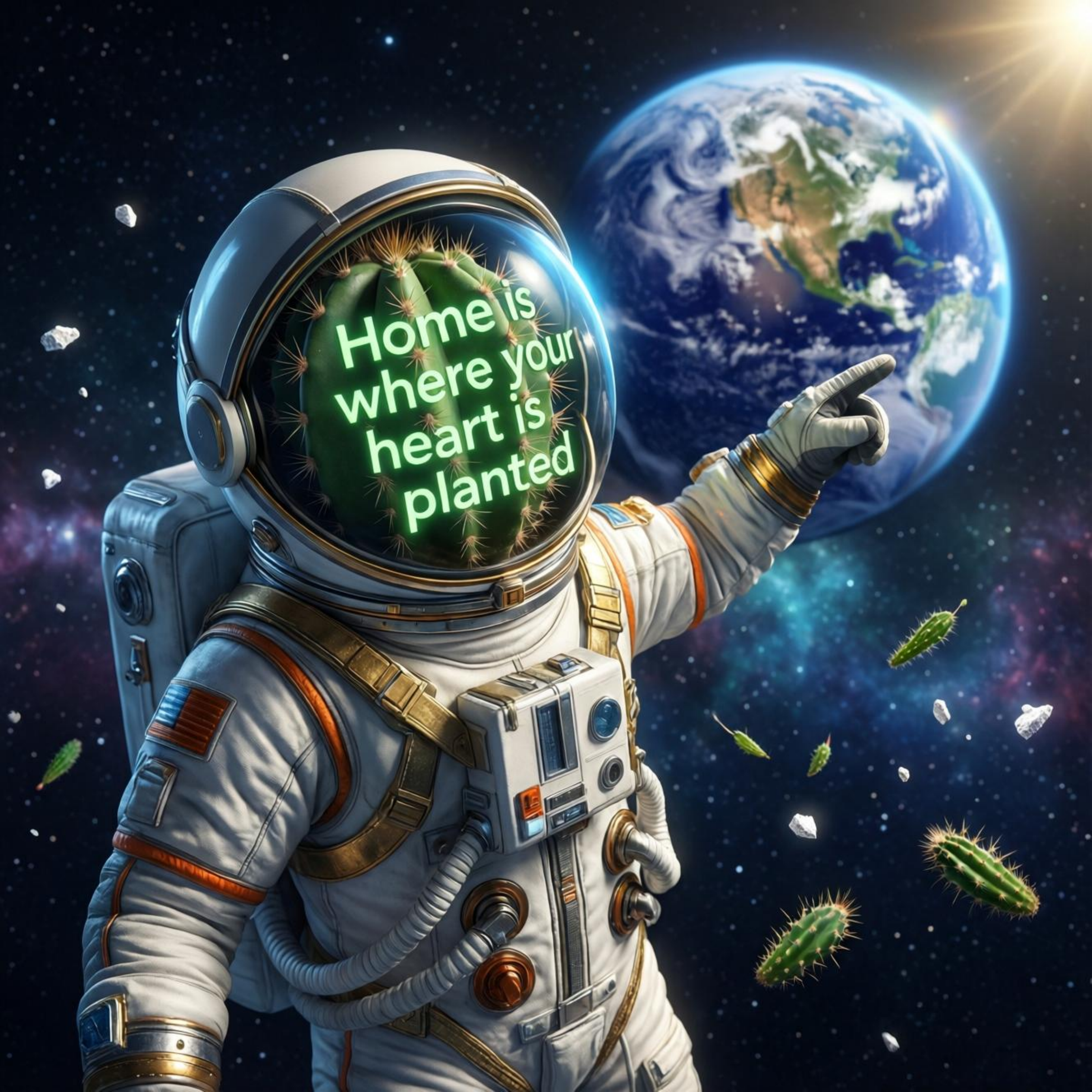}
    {A whimsical, anthropomorphic cactus astronaut floats weightlessly in the vast, star-speckled void of space, its vibrant sage-green, spiky body contrasting with a bulky, retro-futuristic white space suit with metallic gold accents. \apehl{The cactus's spherical, potted-like head is encased in a large, clear bubble helmet with a glossy, reflective visor, the front of which glows with neon-green, bold, futuristic sans-serif typography reading ``Home is where your heart is planted,'' slightly curved to follow the helmet's geometry.} The astronaut's gloved finger points toward a distant, glowing blue-and-white marble of Earth in the upper-right background, with a visible terminator line and soft atmospheric halo.}

\apepromptcasewide
    {Imaginative Structure}
    {A rock musician composed of old black vinyl records, wearing sunglasses, is passionately playing an electric guitar shaped like lightning.}
    {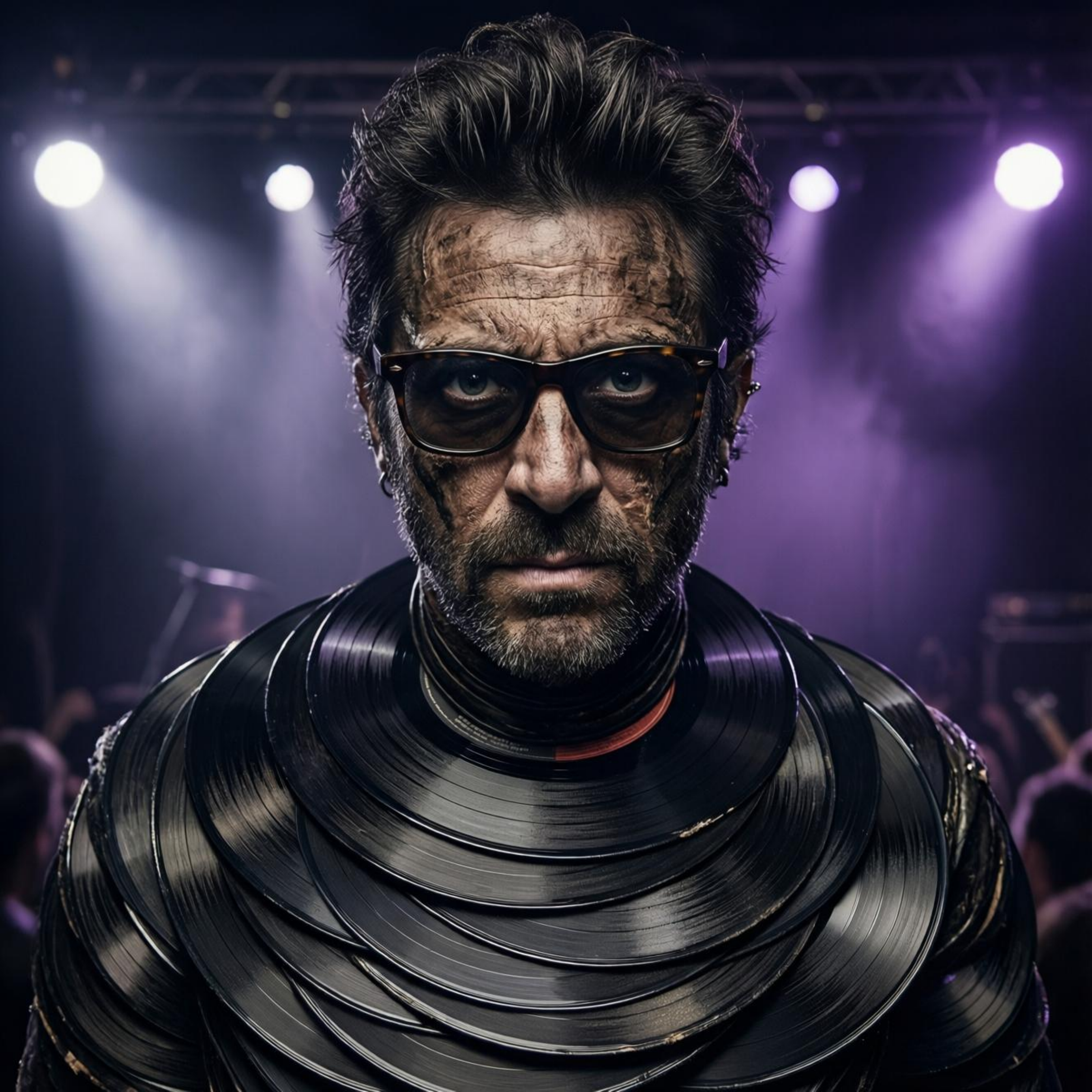}
    {A surreal portrait of a male rock musician constructed entirely from aged black vinyl records, with a rugged human face and dark sunglasses. He stands in a dramatic concert stage bathed in moody spotlights and purple ambient lighting, his body sculpted from stacked vinyl discs in a textured, circular pattern. Ultra HD, 4K, cinematic composition.}
    {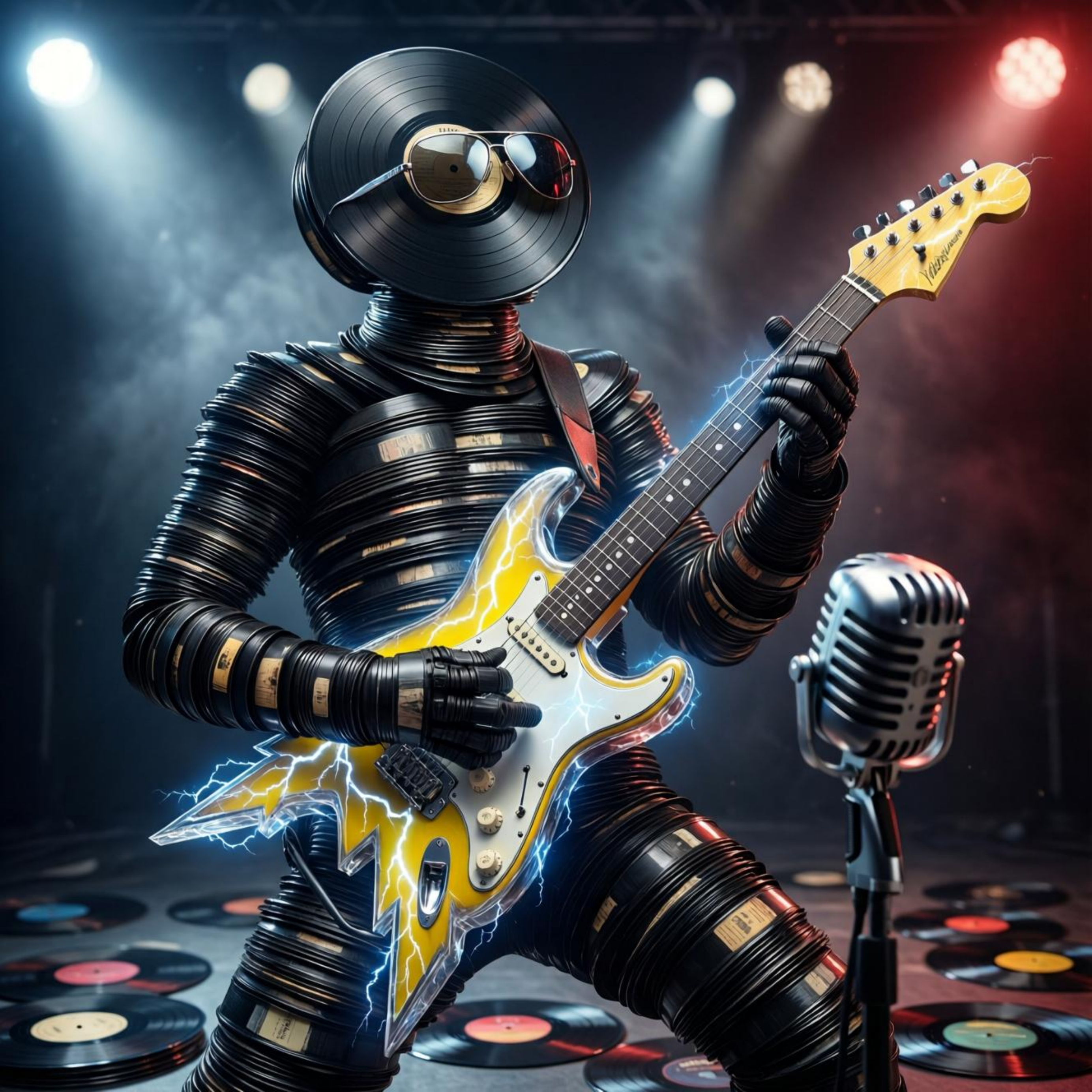}
    {A surreal, photorealistic medium shot of a rock musician whose entire body is constructed from stacked, vintage black vinyl records, their grooved plastic creating a textured, layered silhouette. \apehl{The musician's torso is a stack of smaller 7-inch singles, his neck a cluster of smaller discs, and his head a larger LP record. His right hand grips the neck of a jagged, stylized electric guitar shaped like a bolt of yellow and white lightning, its chrome and translucent acrylic body crackling with subtle magical energy, while his left hand strums the strings with intense passion.} Oversized aviator sunglasses reflect stage lights against a blurred, high-contrast concert stage with deep shadows and piercing spotlights.}

\apepromptcasewide
    {Style and Scene}
    {Generate pictures: Please design a poster for an environmental theme, showing a polar bear standing alone on a piece of ice floe that is about to melt, looking out at the silhouette of the industrial city in the distance, in Memphis style.}
    {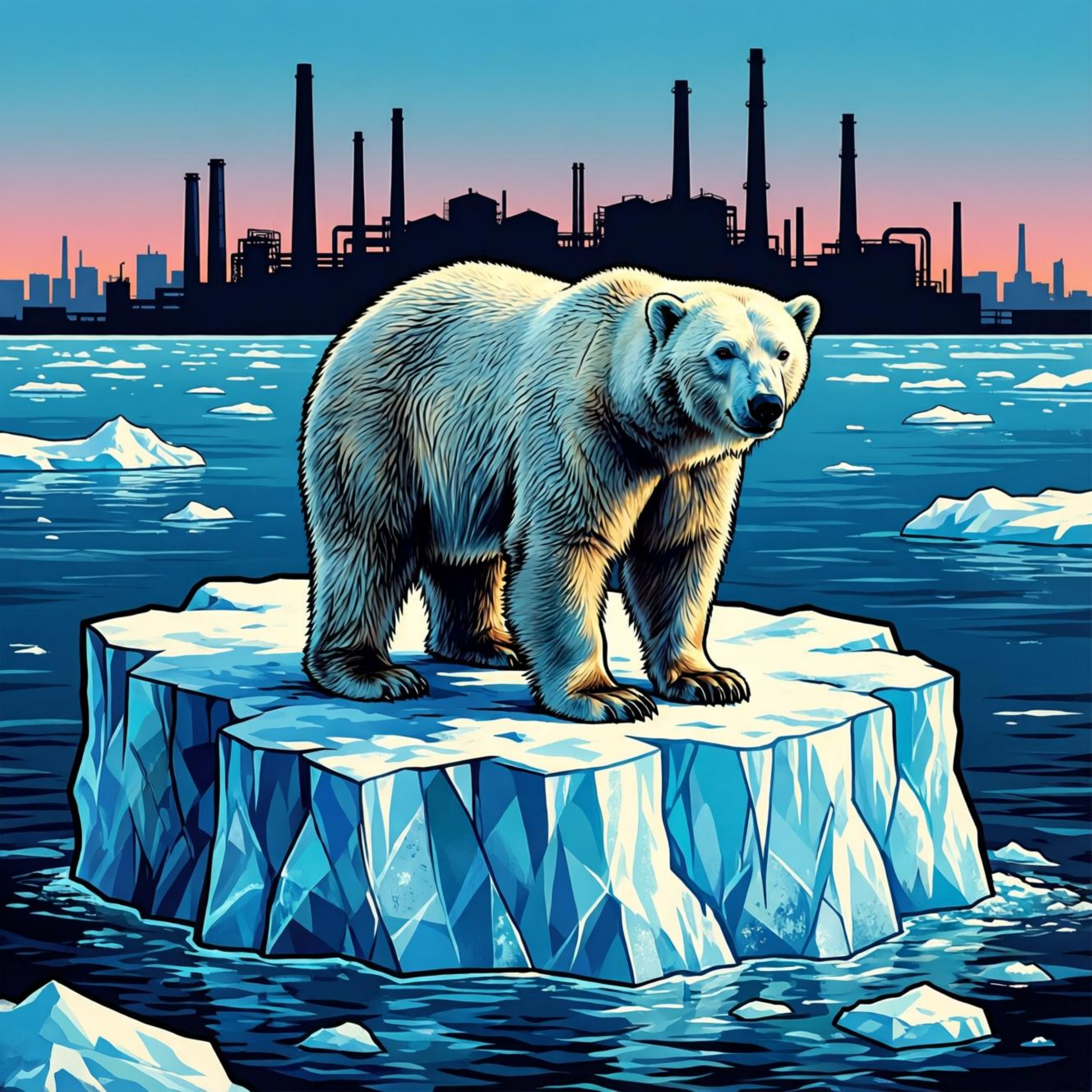}
    {A poster design featuring a polar bear standing alone on a large piece of ice floe that is visibly melting, looking toward an industrial city silhouette in the distance. Rendered in a Memphis style with bold geometric shapes, vibrant flat colors, and distinct black outlines. Ultra HD, 4K, cinematic composition.}
    {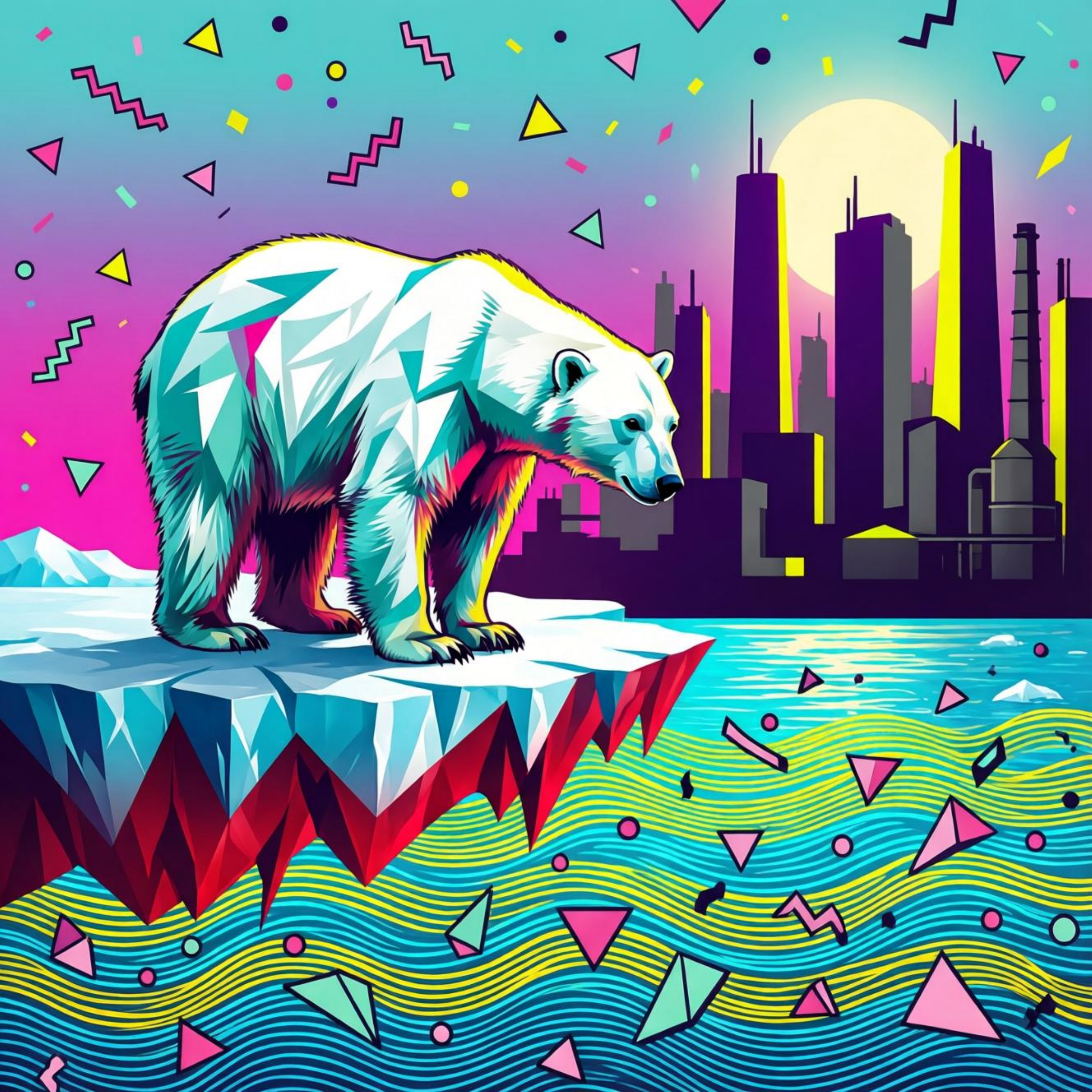}
    {A striking Memphis design poster features a solitary polar bear in bold, flat geometric shapes of stark white with accents of teal and magenta, standing center-left on a jagged, shrinking ice floe that fractures into sharp, angular lines fading to deep red at the edges. \apehl{Below, the ice floe floats on wavy, serpentine water lines in bright yellow and electric blue, surrounded by floating geometric debris---triangles, dots, and zig-zags in pastel pink, mint green, and black---that drift away, emphasizing melting.} An industrial city silhouette in dark purple, charcoal grey, and neon yellow geometric prisms dominates the background. \apehl{The sky transitions from pastel teal to vibrant magenta, punctuated by scattered Memphis shapes.}}

\apepromptcasewide
    {Motion Contrast}
    {Please draw a scene where a white-haired old swordsman is patiently guiding a young swordsman to practice sword. The movements of the two are in sharp contrast. Japanese animation style.}
    {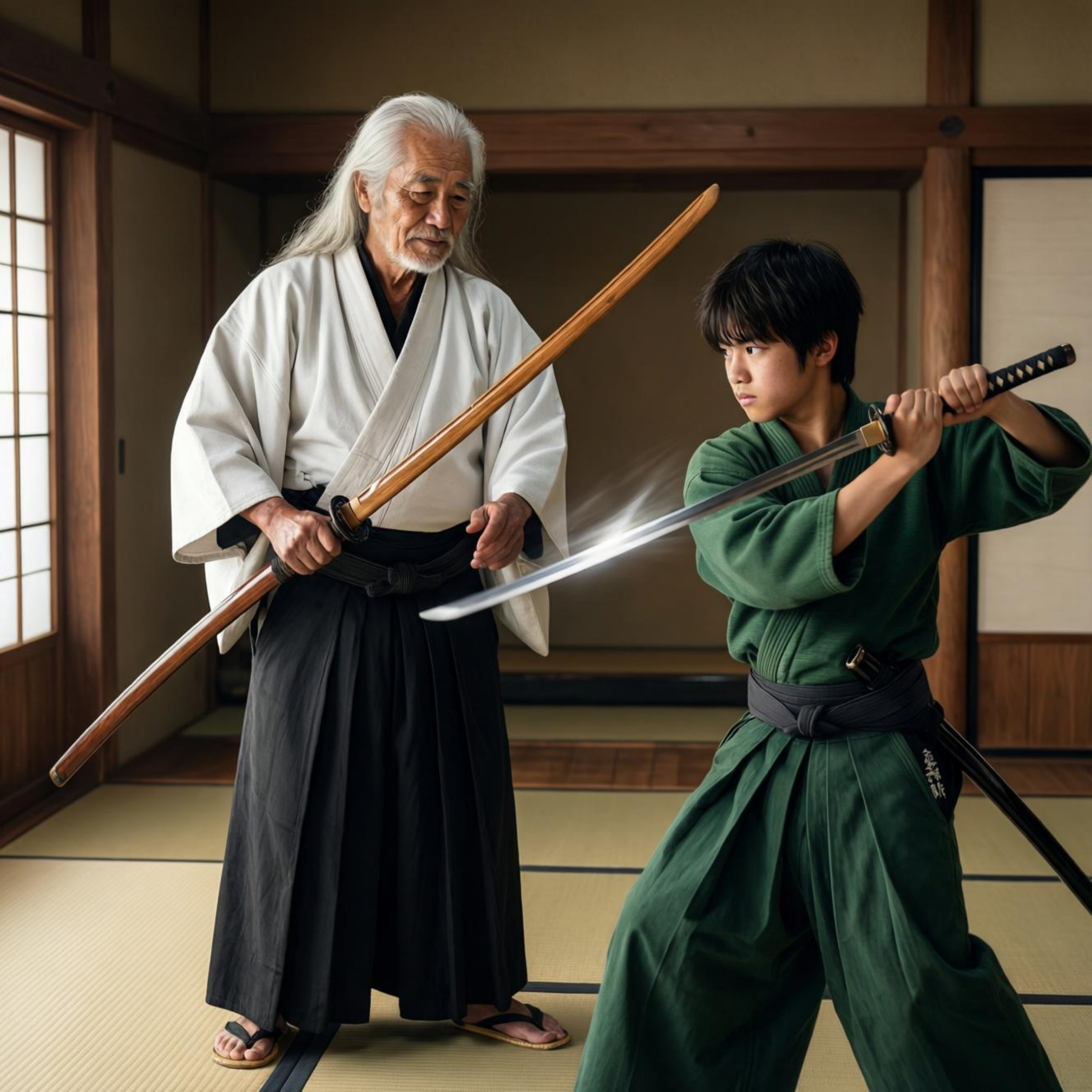}
    {Japanese animation style, an elderly Japanese swordsman with long white hair and a weathered face stands calmly while guiding a young apprentice. The old man wears a traditional white kimono with a black obi sash and holds a wooden training katana in a relaxed posture. Beside him, a young swordsman in his late teens wears a green training gi and practices a swift sword strike. They stand on a tatami mat floor in a traditional wooden dojo with soft morning light. Ultra HD, 4K, cinematic composition.}
    {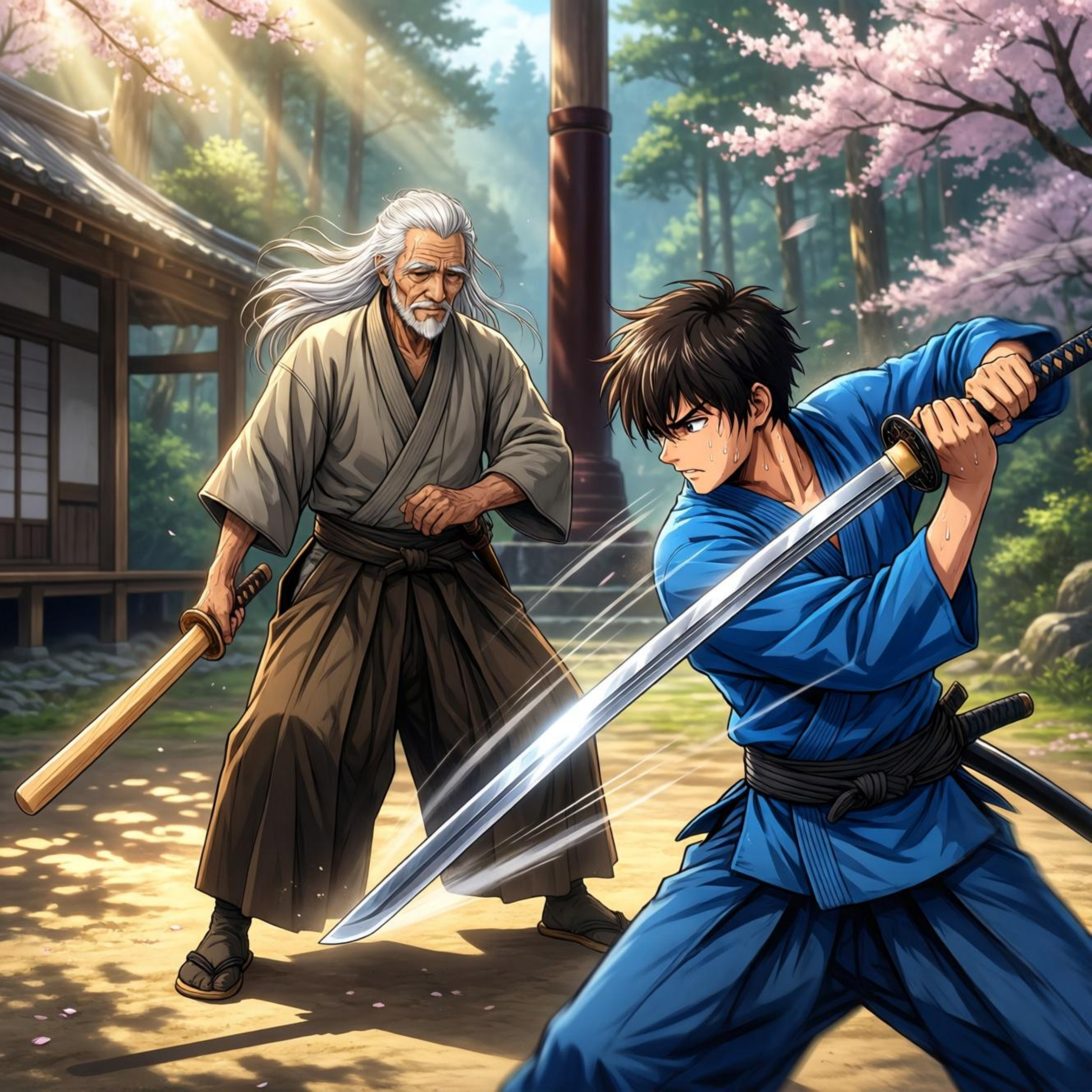}
    {A dynamic Japanese anime-style illustration of a sun-dappled mountain dojo. The elderly swordsman, in muted beige and brown samurai robes, holds a wooden bokken loosely in his left hand, his serene expression focused on the apprentice. The apprentice, dynamic and fluid, is mid-swing, his bright blue training kimono and gleaming silver katana emphasizing intense focus; motion lines and speed effects trail his blade's arc, while the old man's grounded shadow contrasts the kinetic energy. \apehl{The background features traditional wooden architecture, cherry blossom trees in full bloom, and a tall, dark-brown training post in the distance, bathed in the warm, golden light of late afternoon.}}

%% file: intro.tex
\section{Introduction}
Recent advances in generative visual systems have turned natural language into a powerful interface for image creation and editing. Users can now describe scenes, styles, objects, spatial relations, and editing intents in free-form text, and modern models translate these instructions into high-fidelity images or targeted transformations of existing images \cite{podellsdxl,rombach2022high}. As these systems become increasingly capable, the bottleneck is shifting from whether they can synthesize plausible visual content to whether they can faithfully interpret complex, compositional, and often underspecified user intent.

This interpretive bottleneck becomes apparent in the prompt sensitivity of text-guided generation and editing systems. Semantically similar requests can lead to substantially different outputs depending on wording, specificity, and how explicitly visual constraints are stated. For instance, models may satisfy individual object attributes while failing spatial relations, alter regions that should remain unchanged, or follow stylistic cues at the expense of compositional correctness. 

From \Cref{tab:qual_samples_teaser}, we observe that simple prompts can lead to model confusions and failures in areas such as positional reasoning, object count, aesthetics quality and faithfulness. Such failures suggest that prompt sensitivity is not merely a usability issue, but a symptom of a deeper gap between user intent and model interpretation. Recent work has shown that richer textual descriptions can improve visual quality and prompt adherence \cite{betker2023improving,gutflaish2025generating}, indicating that prompt formulation itself should be treated as an essential component of modern generative visual systems rather than a peripheral user choice.

\begin{table*}[!ht]
    \centering
    \scriptsize
    \setlength{\tabcolsep}{3pt}
    \renewcommand{\arraystretch}{1.05}
    \caption{Qualitative comparison between the original prompt and the enhanced prompt. The image model is Z-Image-turbo, and the enhancer LLM is Qwen3-4B after GRPO.}
    \label{tab:qual_samples_teaser}
    \begin{tabular}{L{0.14\textwidth} C{0.14\textwidth} L{0.53\textwidth} C{0.14\textwidth}}
        \toprule
        \textbf{Original Prompt} &
        \makecell[c]{\textbf{Image from}\\\textbf{Original Prompt}} &
        \textbf{Enhanced Prompt} &
        \makecell[c]{\textbf{Image from}\\\textbf{Enhanced Prompt}} \\
        \midrule

        a photo of a skateboard above a person. &
        \includegraphics[width=0.13\textwidth]{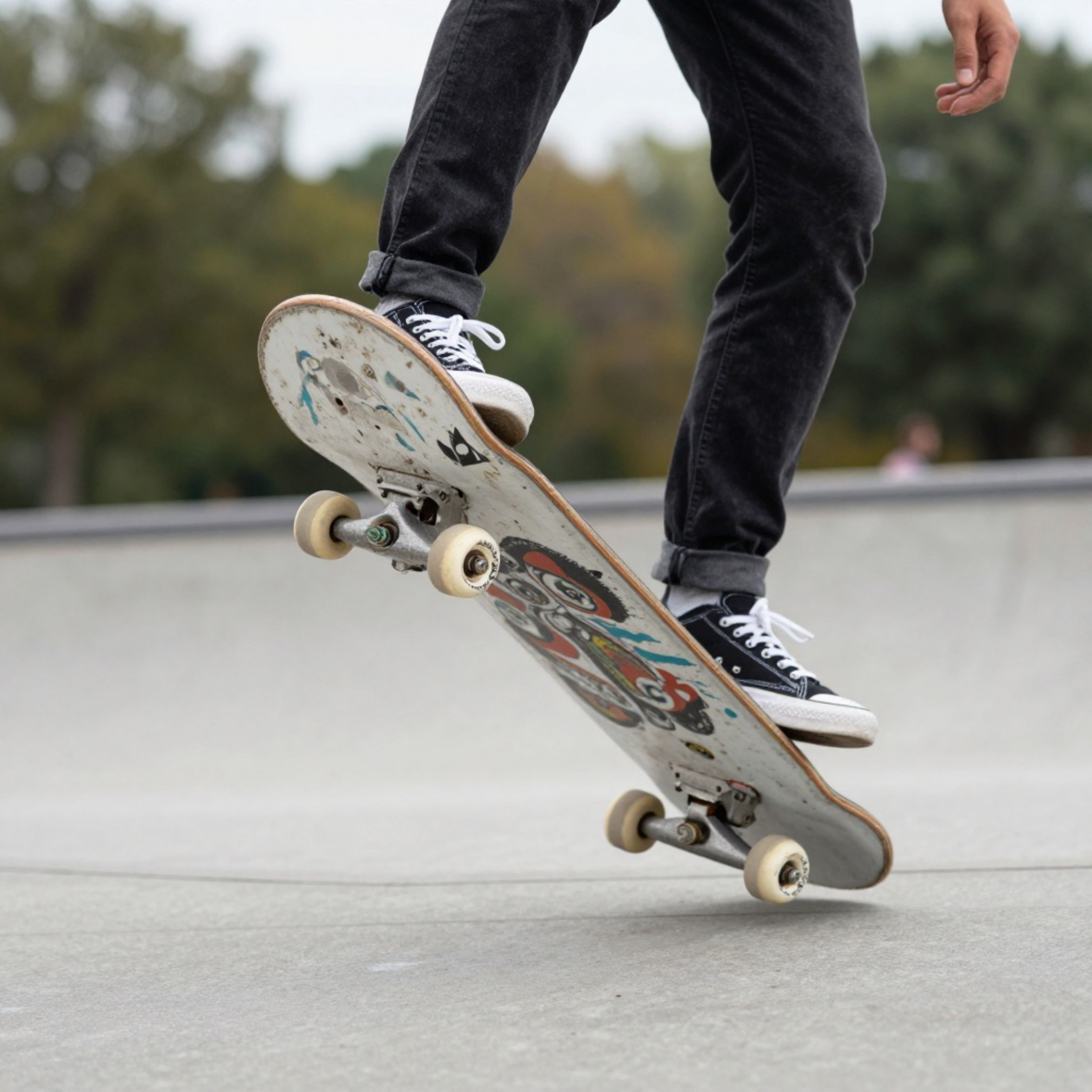} &
        A realistic photograph of a person standing on the ground, looking up in surprise, \textbf{with a skateboard floating just above their head}. The skateboard is bright red with white stripes, its wheels slightly spinning due to a gentle breeze. The person is wearing a light blue shirt and jeans, with a relaxed posture. The background is a clear sky with soft, fluffy clouds and a subtle gradient of blue and white. \textbf{The skateboard is positioned slightly above the person, creating a sense of depth and magic, with faint light rays spreading from it}. The overall scene captures a surreal, whimsical moment with natural lighting and subtle motion effects. &
        \includegraphics[width=0.13\textwidth]{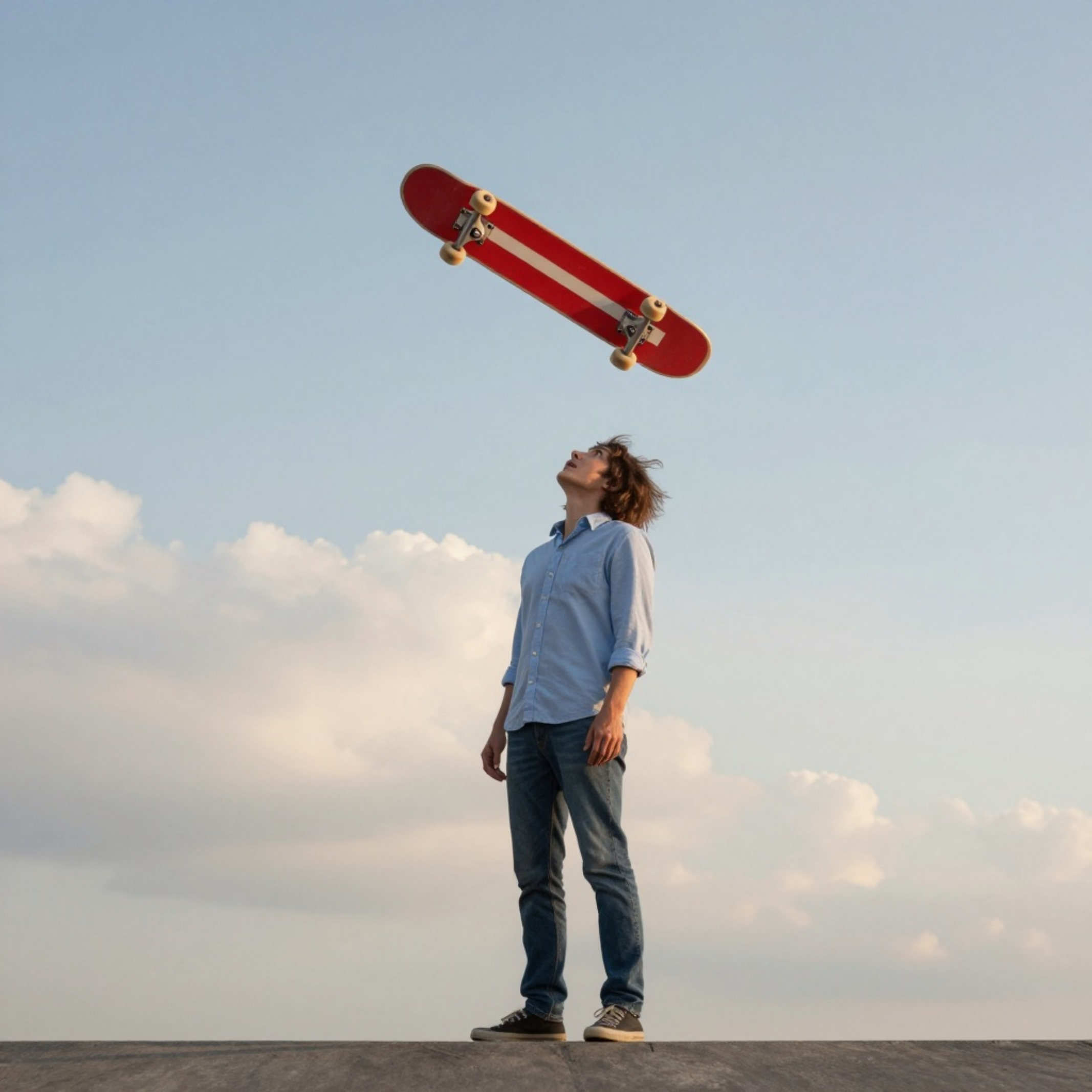} \\
        \midrule

        a photo of four donuts. &
        \includegraphics[width=0.13\textwidth]{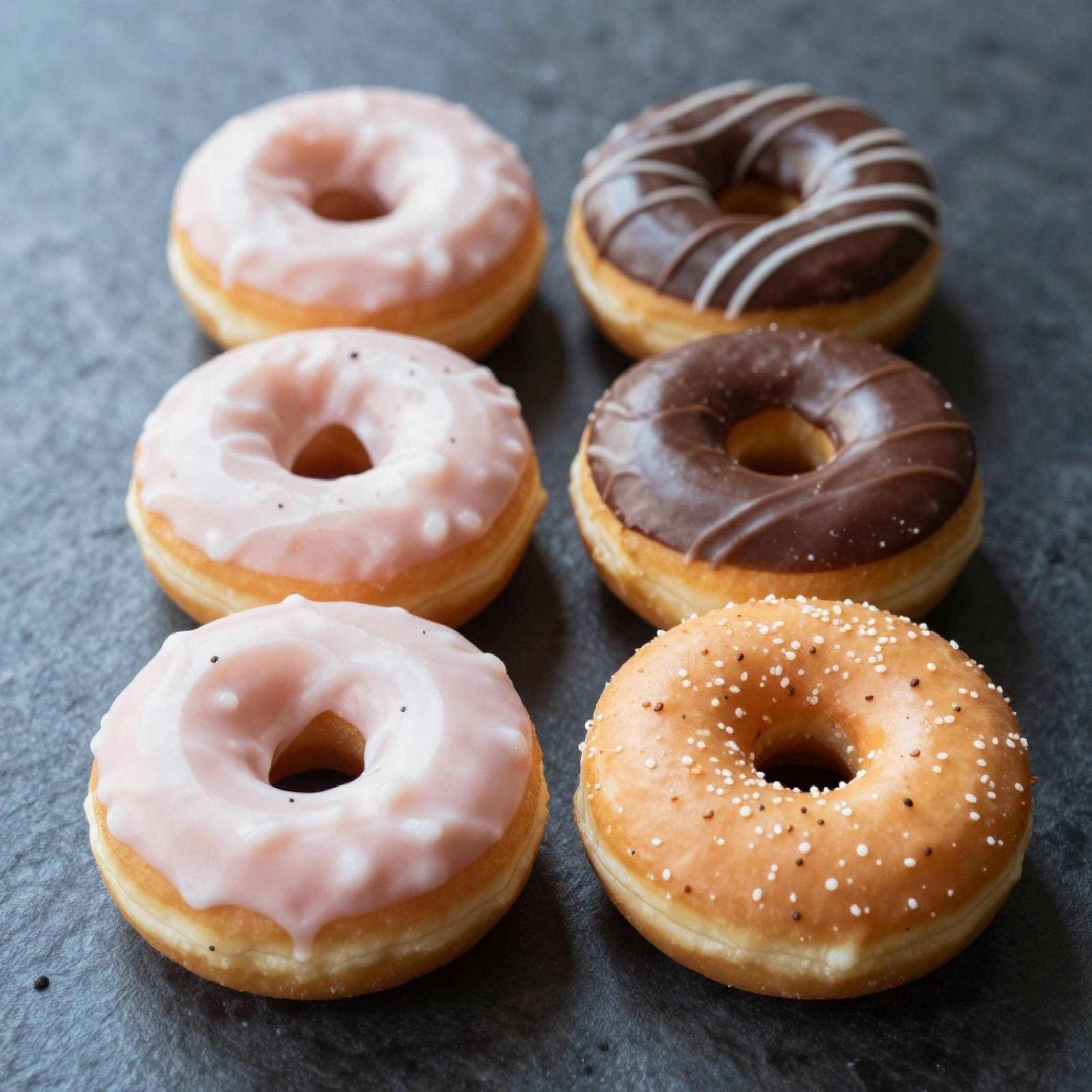} &
        A high-resolution photograph of four distinct donuts on a modern, clean kitchen counter under soft, natural morning light. \textbf{The first donut is a classic jelly donut with a pink, glossy surface and a slight golden crumbly texture, the second is a chocolate-hazelnut donut with a deep brown color and a textured, slightly cracked surface. The third is a vanilla donut with a light beige color and a smooth, slightly moist texture, and the fourth is a blueberry donut with a light blue color and a delicate, airy crumb. The donuts are arranged in a line, slightly varied in size and position to add visual interest, with a white, smooth counter background and subtle wooden grain texture.} The scene is bright, airy, and inviting, with a soft shadow cast by a window to add depth. &
        \includegraphics[width=0.13\textwidth]{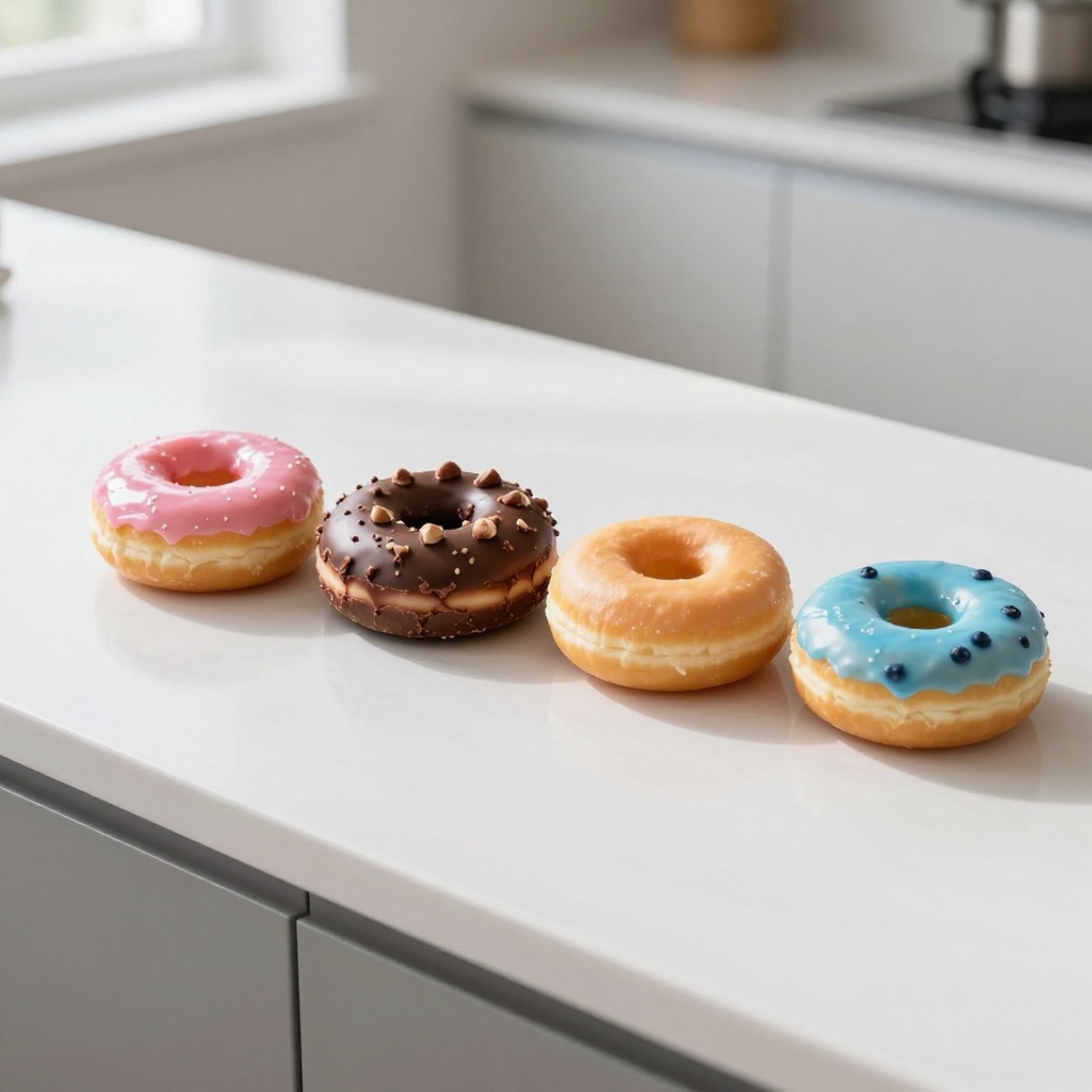} \\
        \midrule

        a photo of a green hot dog. &
        \includegraphics[width=0.13\textwidth]{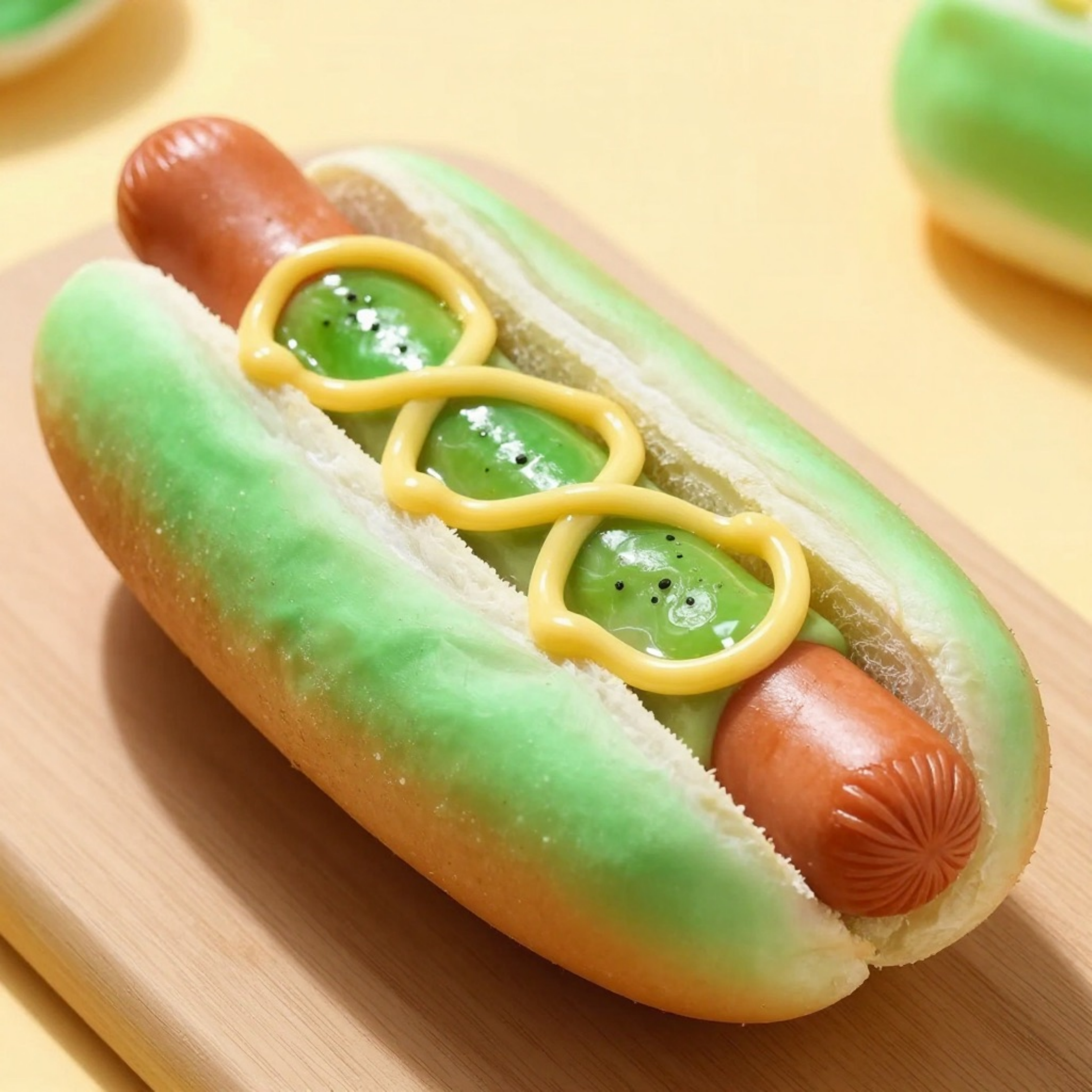} &
        A detailed photograph of a green hot dog on a white plate, with a crispy, golden-brown bun. The hot dog is slightly grilled, showing subtle smoke marks and a slight char on the exterior. \textbf{The sausage inside is a mild, green-hued variety, with a smooth, slightly juicy texture. The hot dog is topped with a creamy green relish, fresh lettuce leaves, a slice of tomato, and a small amount of melted cheddar cheese}. The background is a clean, bright kitchen counter with soft natural light coming through a window, creating a warm, inviting atmosphere. The colors are vibrant: the green of the hot dog and relish stand out against the white plate and light background, with subtle shadows adding depth to the texture of the bread and sausage. The overall scene is casual and appetizing, capturing the essence of a freshly made, green-hot dog meal. &
        \includegraphics[width=0.13\textwidth]{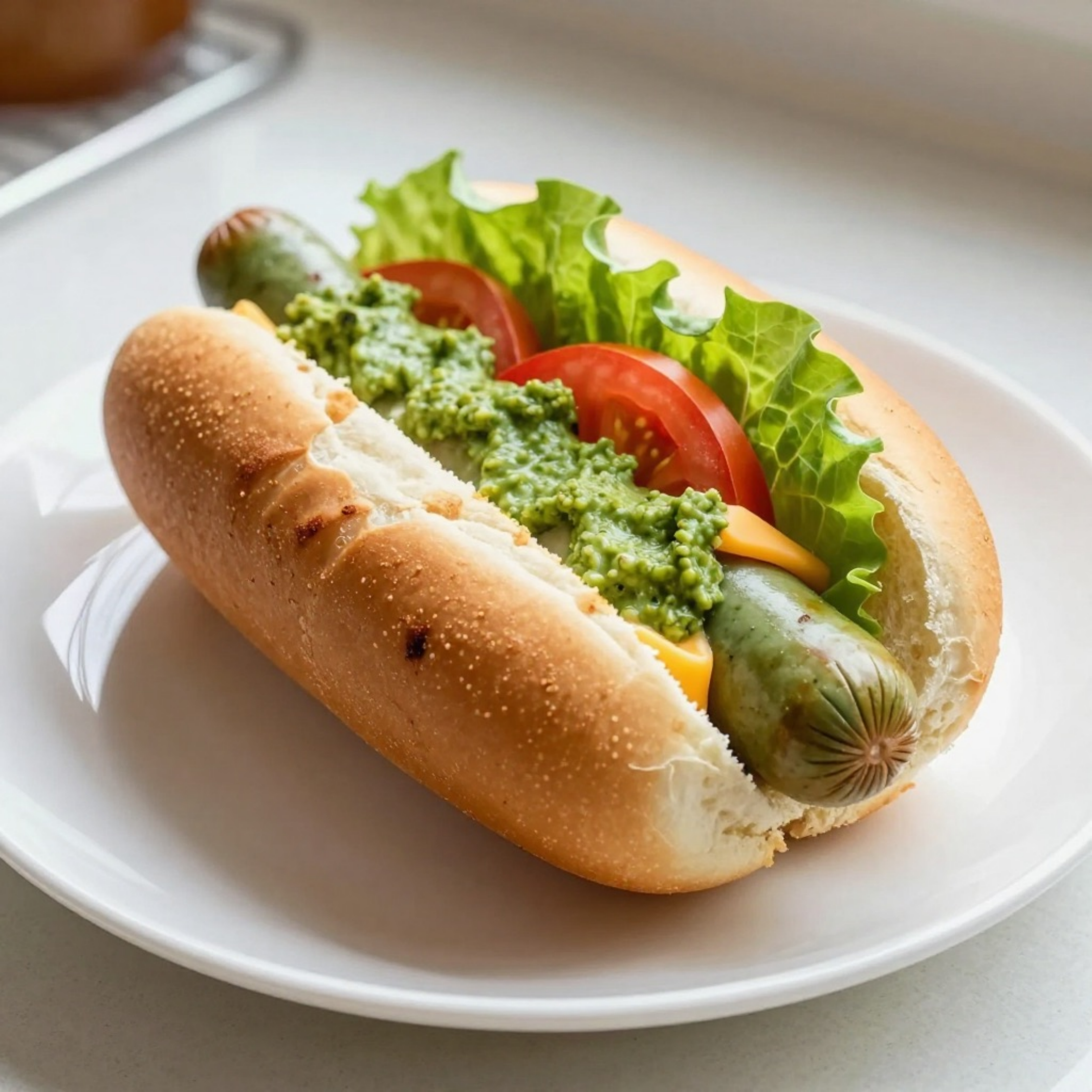} \\
        \bottomrule
    \end{tabular}
\end{table*}

A natural response is prompt enhancement: rewriting or enriching the user request before it reaches the visual model. In practice, strong enhancers often rely on large, proprietary LLMs such as ChatGPT or Gemini, adding cost, latency, and deployment dependence to the visual generation pipeline. This raises a simple question: \textit{can a much smaller language model be post-trained to serve as an effective prompt enhancer?} We address this with \textit{Agentic Prompt Enhancer (APE)}, a lightweight framework for post-training SLMs as prompt-enhancement agents. We first study a \textit{Single-Agent Prompt Enhancer (SAPE)}, where an SLM rewrites the input prompt in one pass. Surprisingly, reinforcement learning on this small enhancer alone yields substantial gains in visual alignment and prompt following without modifying the downstream visual model.

While SAPE demonstrates the promise of lightweight prompt enhancement, a single-pass rewriter can struggle with complex, compositional requests. Effective enhancement often requires deciding which aspects of the user intent need refinement, adding the right level of detail, and preserving the original intent when these refinements are combined. Instead of requiring one small model to make all of these decisions implicitly, \textit{Multi-Agent Prompt Enhancer (MAPE)} assigns them to separate role-specialized agents through a structured \textit{router--rewriter--composer} process. The router selects relevant semantic fields, specialized rewriters refine the selected fields, and a composer assembles the final natural-language prompt. By separating selection, refinement, and composition, MAPE better handles prompts with interacting constraints over objects, attributes, spatial relations, and edits, while maintaining a lightweight interface to the downstream visual model.

Empirically, we find that post-training small prompt-enhancement agents yields consistent gains in visual alignment and prompt following, and that the multi-agent design is especially effective on complex, compositional tasks. Across image generation and editing settings, with task-aware rewards and post-training protocols, APE narrows the gap between lightweight open models and powerful closed-source prompt enhancers, achieving comparable performance on challenging benchmarks, which is a gap reduction from $-9\%$ to the range of $(-3\%, +2\%)$. 
More broadly, our results suggest that prompt enhancement is not merely an interface trick, but a scalable and trainable component for improving generative visual systems.

Overall, our contributions are threefold:
\begin{itemize}[leftmargin=0pt, itemindent=1.5em, itemsep=0pt, topsep=0pt]
    \item We propose APE, a lightweight framework for post-training SLMs as prompt enhancers, instantiated as SAPE and MAPE for generation and editing.
    \item We show that RL improves both single-agent and multi-agent enhancers in visual alignment and prompt following without updating the downstream visual model.
    \item We develop task-aware post-training protocols and show that APE improves visual alignment and prompt following, with MAPE providing larger gains on complex compositional prompts.
\end{itemize}

%% file: related.tex
\section{Related Work}
\textbf{Generative Models for Image Synthesis.}
Image synthesis has evolved from autoregressive and GAN-based methods \cite{van2016conditional,parmar2018image,ramesh2021zero,goodfellow2020generative,karras2019style,radford2015unsupervised} to diffusion and flow-matching paradigms \cite{ho2020denoising,nichol2021improved,song2020score,lipman2022flow}. Recent latent, transformer-based, and flow-based generators such as LDMs, FLUX, Z-Image, and Qwen-Image have substantially improved instruction following and visual fidelity \cite{rombach2022high,peebles2023scalable,esser2024scaling,flux-2-2025,cai2025z,wu2025qwen}. Image editing has advanced in parallel through instruction-aligned editing backbones such as FLUX.1-Kontext, FLUX.2-klein, Qwen-Image-Edit, and ChronoEdit \cite{labs2025flux,flux-2-2025,wu2025qwen,wu2025chronoedit}. Across both generation and editing, prompt quality strongly affects semantic alignment and faithfulness \cite{betker2023improving,gutflaish2025generating}, motivating trainable prompt enhancement.

\textbf{Prompt Enhancement.}
Prompt enhancement improves generator performance without modifying the generator itself. Prior work in NLP studies automatic prompt optimization through search, ranking, or evolution \cite{zhou2022large,fernandopromptbreeder}, while recent text-to-image methods use LLMs or learned rewriters to expand underspecified user prompts into richer visual descriptions \cite{lianllm,wang2025promptenhancer}. However, most existing enhancers are single-stage and rely primarily on prompting or supervised rewriting. In contrast, APE treats SLMs as trainable prompt enhancers, and MAPE further decomposes enhancement into specialized stages for semantic-field selection, rewriting, and composition.

\textbf{Multi-Agent Systems.}
LLM-based multi-agent systems improve complex-task performance by decomposing problems into coordinated interactions among specialized agents \cite{guo2024large,li2024survey,tran2025multi}. Prior frameworks such as CAMEL, AutoGen, MetaGPT, ChatDev, and AgentVerse demonstrate the value of role specialization and structured coordination \cite{li2023camel,wu2024autogen,hong2023metagpt,qian2024chatdev,chen2023agentverse}. Related work on debate and collaborative reasoning further suggests that decomposed, diverse reasoning can improve solution quality \cite{hegazy2024diversity,hu2025multi}. Motivated by this literature, we cast prompt enhancement as a decomposable multi-agent problem with specialized roles for field selection, field-level rewriting, and final composition.

%% file: method.tex
\begin{figure}[t]
    \centering
    \includegraphics[width=0.95\linewidth]{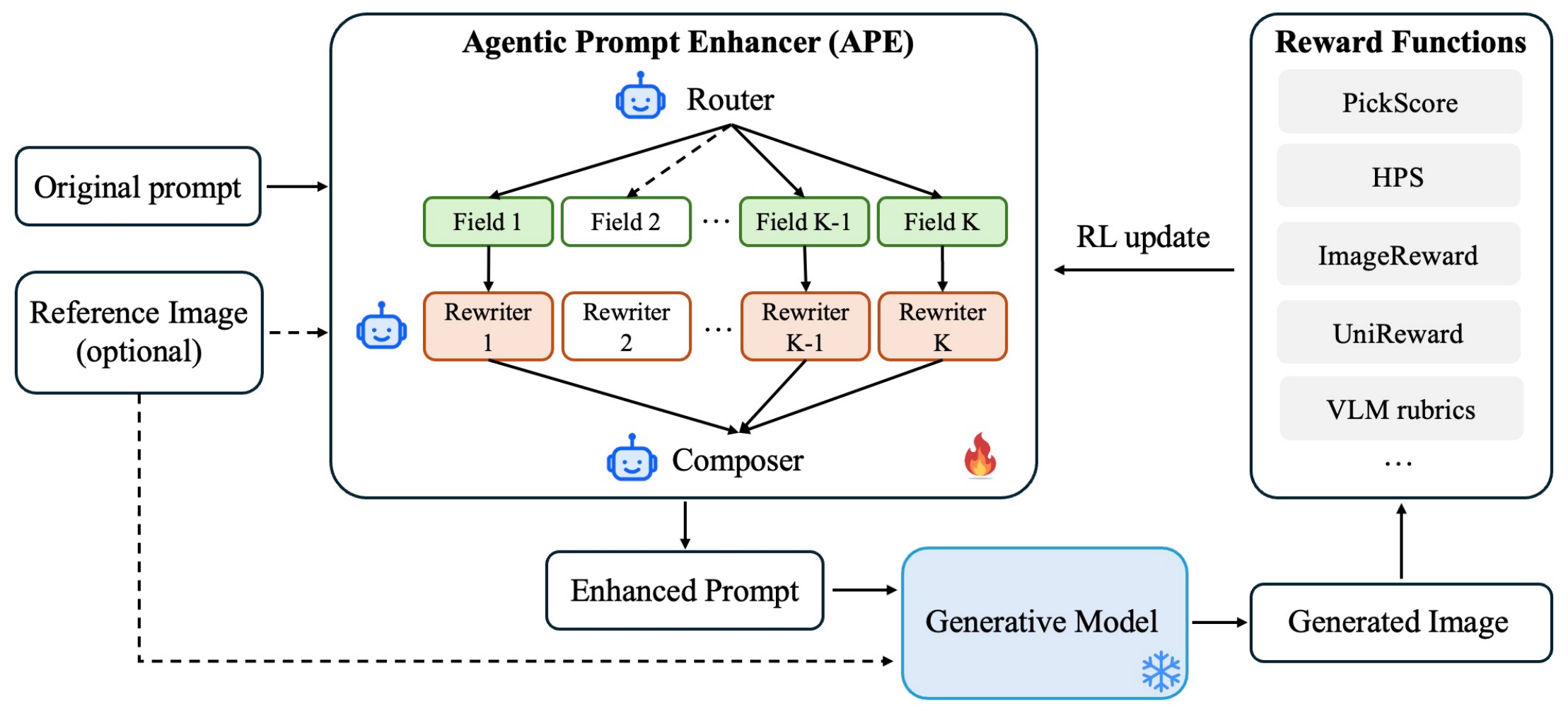}
    \caption{APE for image generation and image editing. APE rewrites user inputs with either SAPE, a single-pass prompt enhancer, or MAPE, a structured multi-agent pipeline in which a router selects fields, field rewriters refine selected fields, and a composer produces the final prompt. The rewritten prompt is then passed to a frozen image generator or editor, while only the prompt-enhancement modules are post-trained with GRPO or GDPO. For editing, the router may also choose not to rewrite simple instructions.}
    \label{fig:ape_framework}
\end{figure}

\section{APE: Agentic Prompt Enhancer}
\label{sec:ape}

We propose APE, a unified framework for prompt enhancement in both image generation and image editing tasks. 
Given a user instruction, and optionally a source image in the editing setting, APE learns a prompt enhancement policy that rewrites the input into a more effective natural-language prompt for a downstream generator or editor. 

We design this framework along two orthogonal axes: architecture and optimization. Architecturally, we study both a Single-Agent Prompt Enhancer (SAPE) and a Multi-Agent Prompt Enhancer (MAPE). For optimization, we use Group Relative Policy Optimization (GRPO) \cite{shao2024deepseekmath} when training feedback is scalar, and Group reward-Decoupled Normalization Policy Optimization (GDPO) \cite{liu2026gdpo} for multi-reward supervision. This decomposition yields a flexible design in which simpler settings may be handled by SAPE, while harder and more compositional settings benefit from MAPE. 

\subsection{Problem Formulation}
Let $x$ denote a user instruction. In image generation, the prompt enhancer receives only $x$. In image editing, it additionally receives a source image $I$. We denote the input state by
{
\begin{align}
    s = \begin{cases} 
x, & \text{for image generation,} \\
(x, I), & \text{for image editing.}
\end{cases}
\end{align}}
A prompt enhancer with parameters $\theta$ defines a conditional policy $\pi_\theta(p|s)$, where $p$ is the enhanced natural-language prompt. The downstream image model $G$ then produces an output image $y=G(s,p)$, where $G(s,p)$ stands for $G(p)$ in generation and $G(I,p)$ in editing.

Unlike standard supervised rewriting, prompt enhancement does not admit a unique ground-truth output. Multiple rewritten prompts may all be valid, and their quality is best evaluated through the resulting images rather than through textual similarity to a reference rewrite. We therefore optimize the prompt enhancer using reward signals computed from downstream outputs. Depending on the setting, the reward may be either a scalar $r\in\mathbb{R}$, or a reward vector $\mathbf{r}=(r^{(1)},r^{(2)},...,r^{(K)})\in\mathbb{R}^K$. Our objective is to learn $\pi_\theta$ so that the downstream outputs induced by its rewritten prompts better satisfy user intent.

\subsection{SAPE: Single-Agent Prompt Enhancer}
The simplest architecture is a Single-Agent Prompt Enhancer (SAPE), in which a single model directly rewrites the input into a final prompt in one step:
{
\begin{align}
    p\sim \pi_\theta^{\text{SAPE}}(\cdot|s)
\end{align}}
SAPE is simple, efficient, and suitable when prompt enhancement mainly requires local rewriting, stylistic polishing, or moderate attribute expansion. However, prompt enhancement becomes harder when the user request is underspecified, highly compositional, or constraint-heavy. In such cases, the enhancer must implicitly decide which semantic aspects need elaboration, how each aspect should be refined, and how these refinements should be expressed coherently in the final prompt. Solving all of these subproblems in a single generation step can lead to entangled reasoning and weaker controllability.

\subsection{MAPE: Multi-Agent Prompt Enhancer}
MAPE addresses a key limitation of SAPE: a single-pass rewriter must implicitly decide what to refine, how to refine it, and how to compose the refinements. MAPE instead introduces a structured intermediate representation over semantic fields, such as subject attributes, scene context, composition, style, and edit-specific constraints. It decomposes prompt enhancement into three role-specialized components: a \textit{router}, field-specific \textit{rewriters}, and a \textit{composer}. The router selects which fields should be enhanced, the rewriters generate refined content for those fields, and the composer integrates the field-level rewrites into the final natural-language prompt. This design is motivated by the observation that high-quality prompts are often compositional, with different inputs requiring refinement of different aspects. Recent structured-prompting work, such as FIBO \cite{gutflaish2025generating}, similarly shows that expanding short free-form prompts into richer structured descriptions can improve controllability and prompt adherence. 

Let $\mathcal{F}=\{f_1,f_2,...,f_M\}$ be a predefined set of candidate prompt fields. For image generation, these may include subject, appearance, background, composition, lighting, and style. For image editing, they may additionally include edit operation, preserved content, modified region, and locality constraints. Given input state $s$, the router predicts a subset of fields $\mathcal{S}\subseteq\mathcal{F}$, or equivalently a binary selection vector $\mathbf{m}\in\{0,1\}^M$, where $m_j=1$ indicates that field $f_j$ should be rewritten. We denote the router policy by $\pi_{\theta_r}(\mathcal{S}|s)$. For each selected field $f_j\in\mathcal{S}$, a field-specific rewriter generates a refined field description $z_j\sim \pi_{\theta_j}^{\text{rw}}(\cdot|s,f_j)$. The composer then converts the selected field rewrites into the final natural-language prompt: $p\sim\pi_{\theta_c}^\text{comp}(\cdot|s,\{(f_j,z_j):f_j\in\mathcal{S}\})$. The overall MAPE policy can thus be written as
{
\begin{align}
    \pi_\theta(p|s)=\pi_{\theta_r}(\mathcal{S}|s)\prod_{f_j\in\mathcal{S}}\pi_{\theta_j}^{\text{rw}}(z_j|s,f_j)\cdot\pi_{\theta_c}^{\text{comp}}(p|s,\{(f_j,z_j)\}_{f_j\in\mathcal{S}})
\end{align}}
where $\theta=\{\theta_r,\theta_1,...,\theta_M,\theta_c\}$. Compared with SAPE, MAPE imposes a stronger inductive bias for harder tasks. This factorization makes the intermediate enhancement decisions explicit, separating field selection, field-level elaboration, and final prompt composition. As a result, MAPE can reduce interference across heterogeneous sub-decisions and provide more targeted enhancement for underspecified or constraint-heavy requests.

\subsection{GRPO and GDPO for Prompt Enhancer Optimization}
We optimize the prompt enhancer with reinforcement learning, where prompt quality is defined by the downstream image produced by the rewritten prompt rather than by a reference rewrite. Given an input state $s$, we sample a group of candidate prompts $p_1,\ldots,p_G \sim \pi_{\theta_{\text{old}}}(\cdot \mid s)$, execute them with the downstream model to obtain outputs $y_i = G(s,p_i)$, and compute rewards from the resulting images.

When supervision is scalar, we use \textit{Group Relative Policy Optimization (GRPO)}. Let $r_i = R(y_i,s)$ denote the scalar reward for sample $i$. GRPO computes a group-normalized advantage 
$
    A_i = \frac{r_i - \mu}{\sigma}, \quad\mu = \frac{1}{G}\sum_{i=1}^G r_i,\quad\sigma = \sqrt{\frac{1}{G}\sum_{i=1}^G (r_i-\mu)^2 + \epsilon},
$
and optimizes the clipped policy objective 
{\small
\begin{align}
    \mathcal{J}_{\text{GRPO}}(\theta; s)
    =
    \frac{1}{G}\sum_{i=1}^G
    \min\!\Big(
        \rho_i(\theta)A_i,\,
        \mathrm{clip}(\rho_i(\theta),1-\epsilon,1+\epsilon)A_i
    \Big)
    - \beta \,\mathrm{KL}\!\left(\pi_\theta(\cdot\mid s)\,\|\,\pi_{\text{ref}}(\cdot\mid s)\right),
\end{align}
}
where $\rho_i(\theta)=\frac{\pi_\theta(p_i\mid s)}{\pi_{\theta_{\text{old}}}(p_i\mid s)}$. If multiple scalar metrics are available, we combine them into a single reward before GRPO.

When supervision is multi-reward, we use \textit{Group reward-Decoupled Normalization Policy Optimization (GDPO)}. Let $\mathbf{r}_i = (r_i^{(1)},\dots,r_i^{(K)})$ denote the reward vector. GDPO first normalizes each reward dimension independently within the group,
$
    A_i^{(k)} = \frac{r_i^{(k)}-\mu^{(k)}}{\sigma^{(k)}},
$
then aggregates them as $\tilde{A}_i = \sum_{k=1}^K w_k A_i^{(k)}$, followed by an additional batch-level normalization $\hat{A}_i = (\tilde{A}_i-\mu_{\text{batch}})/\sigma_{\text{batch}}$. The final clipped objective has the same form as GRPO, but uses $\hat{A}_i$ in place of $A_i$:
{\small
\begin{align}
    \mathcal{J}_{\text{GDPO}}(\theta; s)
    =
    \frac{1}{G}\sum_{i=1}^G
    \min\!\Big(
        \rho_i(\theta)\hat{A}_i,\,
        \mathrm{clip}(\rho_i(\theta),1-\epsilon,1+\epsilon)\hat{A}_i
    \Big)
    - \beta \,\mathrm{KL}\!\left(\pi_\theta(\cdot\mid s)\,\|\,\pi_{\text{ref}}(\cdot\mid s)\right).
\end{align}
}

GRPO and GDPO share the same policy-optimization backbone and differ mainly in how the advantage signal is constructed: GRPO uses a single scalar reward, while GDPO decouples normalization across reward dimensions before aggregation.
By default, SAPE and MAPE refer to the post-trained variants under the above objectives. For MAPE, users may choose which modules in the router--rewriter--composer pipeline are optimized. We additionally refer to the untrained counterparts of these variants as: a one-shot SLM (denoted by its model name) for the single-agent setting, and multi-step prompting (MSP) — the MAPE pipeline applied without any post-training — for the multi-agent setting. Full derivations and algorithmic details are provided in Appendix~\ref{appendix:rl_detail}.

%% file: experiment.tex
\section{Experiments}
In this section, we evaluate APE from three complementary perspectives. We begin with SAPE for image generation, where the goal is to isolate the effect of reinforcement learning and show that post-training the prompt enhancer alone already yields substantial gains. We then study MAPE for image generation, focusing on more challenging and compositional settings where single-agent rewriting is insufficient, and where the effectiveness of the multi-agent design depends critically on architectural choices, field decomposition, and post-training strategy. 
Finally, we investigate how well the framework generalizes beyond generation by applying MAPE to image editing. We demonstrate that deciding which fields to rewrite is especially important in editing tasks, where both preserving the original content and precisely controlling localized changes are essential.

\begin{table*}[t]
    \centering
    \scriptsize
    \renewcommand{\arraystretch}{1.05}
    \caption{Qualitative comparison of prompt enhancement before and after RL.} 
    \label{tab:qual_samples_single}
    \begin{tabular}{L{0.08\textwidth} L{0.28\textwidth} C{0.1\textwidth} L{0.28\textwidth} C{0.1\textwidth}}
        \toprule
        \textbf{Original Prompt} &
        \makecell[c]{\textbf{Enhancer Prompt}} &
        \makecell[c]{\textbf{Generated}\\\textbf{Image}} &
        \makecell[c]{\textbf{Enhancer Prompt}\\\textbf{after GRPO}} &
        \makecell[c]{\textbf{Generated Image}\\\textbf{after GRPO}} \\
        \midrule

        A large motor vehicle carrying passengers by road, typically one serving the public on a fixed route and for a fare. &
        A large, modern sedan with a sleek, polished body, its frame outlined in a dark blue, reflecting the golden light of the road. A muscular driver, wearing a tailored suit, stands at the front, their face lit by the warm glow of a single streetlamp. Passengers, dressed in comfortable, casual attire, are seated in a row, their faces lit by the dim lighting. The vehicle is on a scenic, two-lane road, its wheels crunching through the soft grass of the countryside. The scene is vibrant with the sounds of passing cars and the distant hum of traffic, creating a dynamic and peaceful atmosphere. &
        \includegraphics[width=0.12\textwidth]{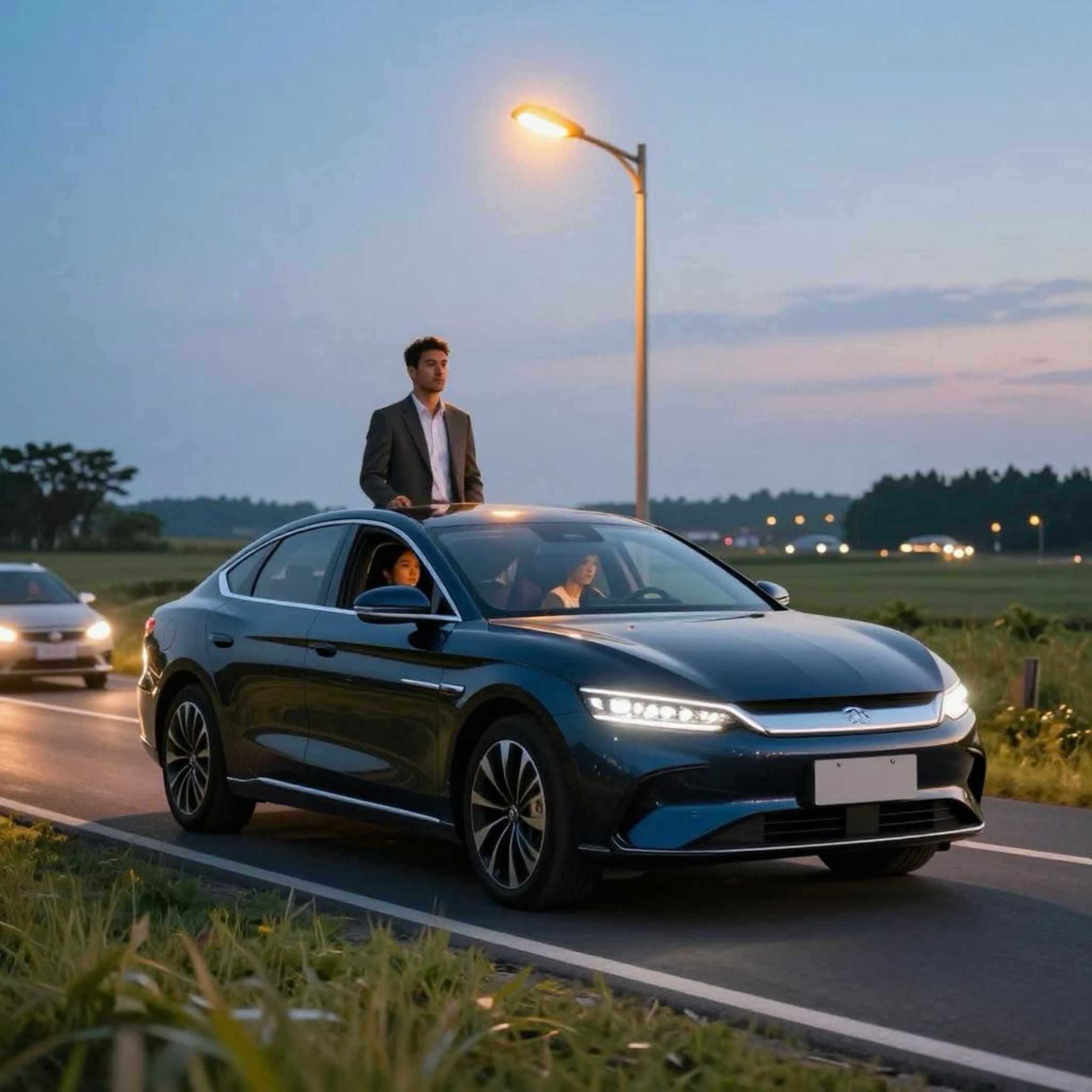} &
        A large motor vehicle carrying passengers by road, typically \textbf{one serving the public on a fixed route and for a fare, with a well-lit, scenic highway in a bustling city or rural area, featuring passengers in their mid- or late 20s, adults, children, or elderly individuals, wearing appropriate attire, and a fare sticker prominently displayed on the vehicle}, emphasizing the public transportation aspect. The vehicle is described with detailed features such as its color, size, and design, while the scene captures the dynamic interaction between the vehicle and its passengers, highlighting the fixed route and the fare system. The setting adds context, making the image more immersive and visually appealing. &
        \includegraphics[width=0.12\textwidth]{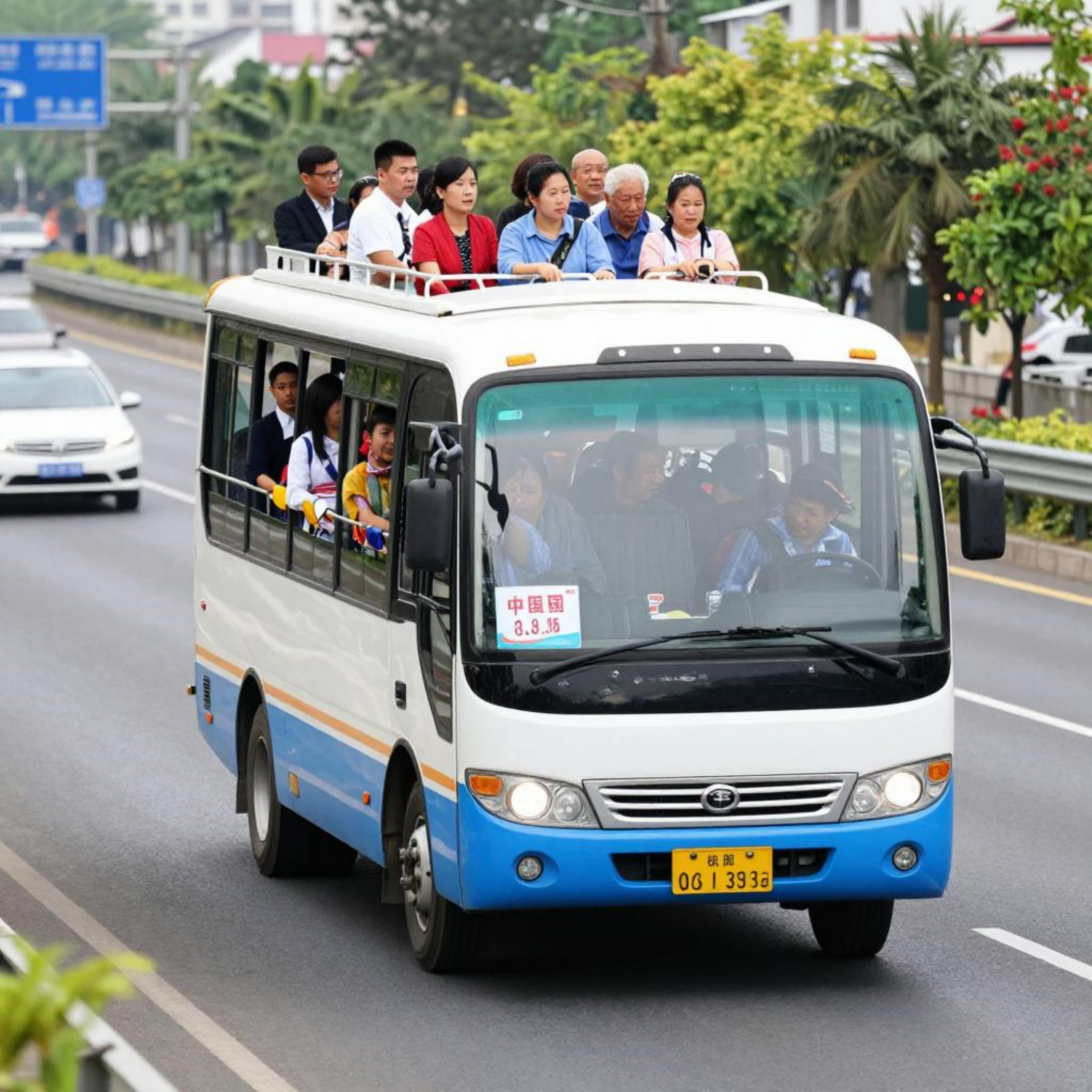} \\
        \midrule

        A donut underneath a toilet. &
        A donut-shaped object sits under a toilet, its smooth, rounded surface reflecting a warm, ambient light. The donut\textbackslash 's interior features a subtle, circular hole, while the toilet\textbackslash 's surface is adorned with decorative tiles or fixtures. The scene is set in a modern bathroom, with soft lighting casting a golden hue and a cozy atmosphere. The donut\textbackslash 's texture is described as smooth and slightly glossy, while the toilet\textbackslash 's surface is warm and inviting, adding to the cozy ambiance. &
        \includegraphics[width=0.12\textwidth]{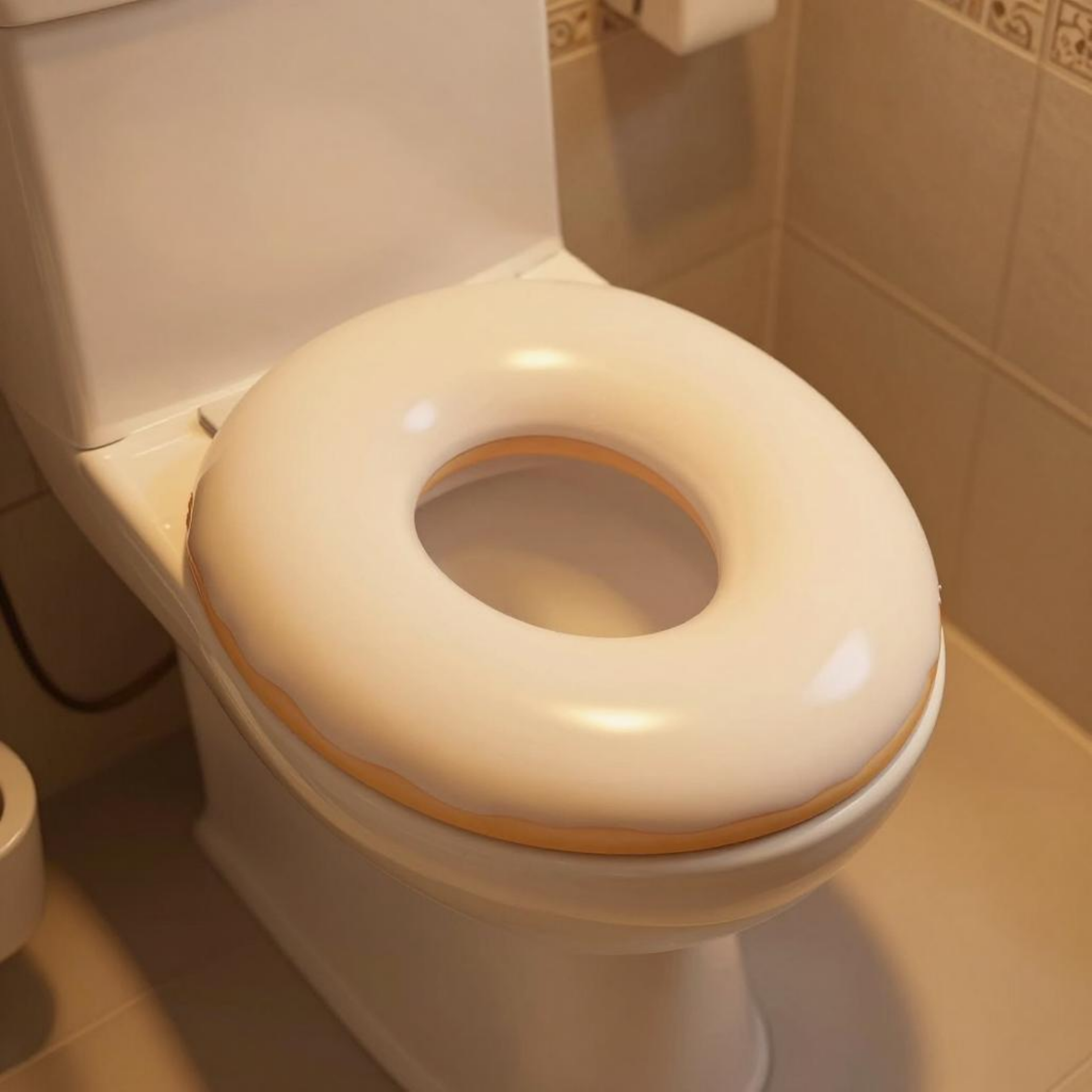} &
        A donut nestled beneath a toilet, its smooth, round shape blending seamlessly with the rustic design of the toilet, with a soft, warm color palette to highlight its texture and form. \textbf{The donut is surrounded by elements of the bathroom environment, such as a sink or tiles, creating a cohesive and detailed scene.} The overall atmosphere is inviting and cozy, emphasizing the donut's presence under the toilet while adding visual interest through its shape and color. &
        \includegraphics[width=0.12\textwidth]{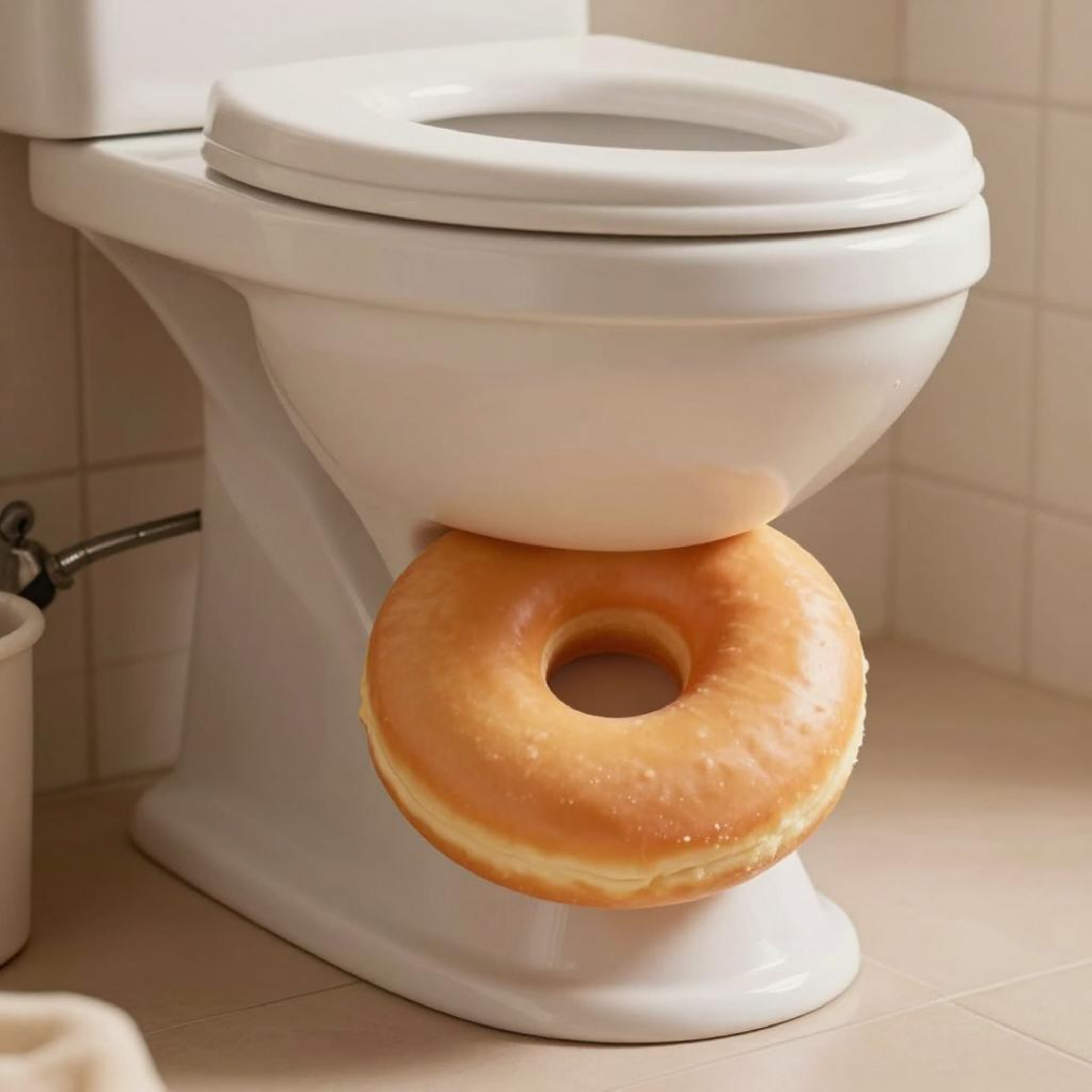} \\
        \bottomrule
    \end{tabular}
\end{table*}

\subsection{Single Agent Prompt Enhancer for Image Generation}
\textbf{Setting.} We study whether reinforcement learning can improve a SAPE while keeping the downstream text-to-image model fixed. We instantiate SAPE with SLMs and apply GRPO only to the prompt enhancer, without updating the image generator. For single-reward training, we consider HPSv2.1 \cite{wu2023human} and PickScore \cite{kirstain2023pick} separately as the optimization target, trained on Pick-a-Pic \cite{kirstain2023pick} and evaluated on DrawBench \cite{saharia2022photorealistic}, as shown in \Cref{tab:hpsv2_1_single} and \Cref{tab:pickscore_single}. For multi-reward training, we use a combined reward of PickScore + CLIPScore \cite{hessel2021clipscore} + HPSv2.1 during GRPO training on Pick-a-Pic, and report PickScore, CLIPScore, HPSv2.1, Aesthetic, ImgRwd \cite{xu2023imagereward}, and UniRwd \cite{wang2025unified} on DrawBench, as shown in \Cref{tab:sape_multi}. This setting aligns with the reward construction in \cite{liuflow,zheng2026diffusionnft}. The image generation model is Z-Image-turbo \cite{cai2025z}, and the SLMs are Qwen3-0.6B and Qwen3-1.7B \cite{yang2025qwen3}. 

\textbf{Result and Analysis.} The results show a clear and consistent pattern: reinforcement learning can substantially improve the prompt enhancer itself, even when the image backbone is frozen. We find that naively applying off-the-shelf SLMs as prompt enhancer deteriorates generation quality. In contrast, under single-reward training, GRPO reliably strengthens SAPE compared with its pre-trained counterpart, showing that a large portion of improvement can be achieved by post-training the SLM prompt enhancer alone. Under multi-reward training, the GRPO-trained SAPE further exhibits broader gains across multiple evaluation metrics, suggesting that the learned improvement is not limited to a single reward signal but transfers to overall prompt quality. 

The qualitative visualizations in \Cref{tab:qual_samples_single} further clarify what the trained prompt enhancer is learning. Compared with the original model, the GRPO-trained enhancer produces prompts that are better at resolving vagueness, grounding abstract descriptions into visually realizable content, and avoiding overthinking or semantically drifting elaborations. For example, instead of expanding a vague prompt into a long but weakly grounded scene that misses the user’s real intent, the trained enhancer is more likely to identify the central concept and rewrite it into a prompt that is visually concrete and directly actionable for the image model. It also reduces cases where the enhancer adds excessive narrative detail or irrelevant associations that distract from the requested concept. In this sense, RL improves not only metric performance but also the behavior of the prompt enhancer: it learns to make prompts clearer, better grounded, and more faithful to the intended semantics, rather than merely longer or more descriptive. Overall, these results establish SAPE as a strong baseline and show that RL on a prompt-enhancing SLM is already highly effective, while also motivating the need for a more structured multi-agent design on harder benchmarks.

\begin{table}[ht]
    \centering
    \footnotesize
    \caption{Single-reward GRPO results for the single-agent prompt enhancer (SAPE) on image generation. GRPO is applied only to the SLM prompt enhancer, while the downstream text-to-image model is kept frozen. Training is performed on Pick-a-Pic and evaluation on DrawBench.}
    \label{tab:single_reward}

    \begin{subtable}[t]{0.48\textwidth}
        \centering
        \setlength{\tabcolsep}{6pt}
        \renewcommand{\arraystretch}{1.08}
        \caption{HPSv2.1-only reward training and evaluation.}
        \label{tab:hpsv2_1_single}
        \begin{tabular}{llc}
            \toprule
            \textbf{T2I Model} & \textbf{Prompt Enhancer} & \textbf{HPSv2.1} \\
            \midrule
            \multirow{5}{*}{Z-Image-turbo}
              & \ding{55}             & 0.2998 \\
              \cmidrule(l){2-3}
              & Qwen3-0.6B            & 0.2801 \\
              & SAPE (Qwen3-0.6B)     & \textbf{0.3234} \\
              \cmidrule(l){2-3}
              & Qwen3-1.7B            & 0.3015 \\
              & SAPE (Qwen3-1.7B)     & \textbf{0.3141} \\
            \bottomrule
        \end{tabular}
    \end{subtable}
    \hfill
    \begin{subtable}[t]{0.48\textwidth}
        \centering
        \setlength{\tabcolsep}{6pt}
        \renewcommand{\arraystretch}{1.08}
        \caption{PickScore-only reward training and evaluation.}
        \label{tab:pickscore_single}
        \begin{tabular}{llc}
            \toprule
            \textbf{T2I Model} & \textbf{Prompt Enhancer} & \textbf{PickScore} \\
            \midrule
            \multirow{5}{*}{Z-Image-turbo}
              & \ding{55}             & 23.0484 \\
              \cmidrule(l){2-3}
              & Qwen3-0.6B            & 22.2293 \\
              & SAPE (Qwen3-0.6B)     & \textbf{23.2233} \\
              \cmidrule(l){2-3}
              & Qwen3-1.7B            & 22.9124 \\
              & SAPE (Qwen3-1.7B)     & \textbf{23.2066} \\
            \bottomrule
        \end{tabular}
    \end{subtable}
\end{table}

\begin{table}[ht]
    \centering
    \footnotesize
    \setlength{\tabcolsep}{4.5pt}
    \renewcommand{\arraystretch}{1.08}
    \caption{Multi-reward GRPO results for the SAPE on image generation. GRPO uses a combined reward of PickScore, CLIPScore, and HPSv2.1 during training on Pick-a-Pic, while evaluation on DrawBench reports a broader set of metrics, including PickScore, CLIPScore, HPSv2.1, Aesthetic, ImgRwd, and UniRwd. All metrics are higher-is-better. Results show that post-training the prompt enhancer alone can improve prompt quality under multiple evaluation criteria.}
    \label{tab:sape_multi}
    \begin{tabular}{llcccccc}
        \toprule
        \multirow{2}{*}{\textbf{T2I Model}} & \multirow{2}{*}{\textbf{Prompt Enhancer}}
          & \multicolumn{3}{c}{\textbf{Training-reward metrics}}
          & \multicolumn{3}{c}{\textbf{Held-out metrics}} \\
        \cmidrule(lr){3-5} \cmidrule(l){6-8}
        & & \textbf{PickScore} & \textbf{ClipScore} & \textbf{HPSv2.1}
          & \textbf{Aesthetic} & \textbf{ImgRwd} & \textbf{UniRwd} \\
        \midrule
        \multirow{5}{*}{Z-Image-turbo}
          & \ding{55}             & 23.0484 & 0.2831 & 0.2998 & 5.3670 & 1.0623 & 3.4161 \\
          \cmidrule(l){2-8}
          & Qwen3-0.6B            & 22.2293 & 0.2524 & 0.2801 & \textbf{5.6232} & 0.6035 & 3.0676 \\
          & SAPE (Qwen3-0.6B)     & \textbf{23.1789} & \textbf{0.2814} & \textbf{0.3128} & 5.4807 & \textbf{1.1915} & \textbf{3.4721} \\
          \cmidrule(l){2-8}
          & Qwen3-1.7B            & 22.9124 & 0.2718 & 0.3015 & \textbf{5.5635} & 0.9297 & 3.3736 \\
          & SAPE (Qwen3-1.7B)     & \textbf{23.0528} & \textbf{0.2768} & \textbf{0.3051} & 5.4951 & \textbf{1.0433} & \textbf{3.4803} \\
        \bottomrule
    \end{tabular}
\end{table}

\subsection{Multi-Agent Prompt Enhancer for Image Generation}

\textbf{Setting.} We next study Multi-Agent Prompt Enhancer (MAPE) on image generation, to test whether multi-agent prompt enhancement is more effective than single-agent rewriting on harder and more compositional tasks. We instantiate the language-side enhancer with Qwen3-1.7B and Qwen3-4B, and evaluate across multiple downstream image generators, including Qwen-Image-2512~\cite{wu2025qwen}, Z-Image-turbo, and FLUX.2-klein-4B/9B~\cite{flux-2-2025}. The field space consists of 10 semantic domains borrowed from FIBO, providing a structured decomposition of prompt content. In implementation, we disable thinking mode for the router and rewriters to keep field selection and field-level rewriting lightweight, while enabling thinking mode for the composer so that it can better integrate multiple rewritten fields into a coherent final prompt. We collect 10,000 SFT examples together with scoring rubrics from Gemini-3.1-Pro~\cite{gemini2026pro}, and use UniGenBench~\cite{wang2025unigenbenchpp} as the main benchmark. Evaluation is conducted on both short and long test samples. We first perform 3 epochs of SFT, and then apply 30 GDPO steps for post-training, using a batch size of 128 and a group size of 8.

\begin{table}[!htbp]
    \centering
    \footnotesize
    \setlength{\tabcolsep}{6pt}
    \renewcommand{\arraystretch}{1.08}
    \caption{UniGenBench results. MAPE consistently matches or surpasses the strong Gemini-3.1-Pro~(MSP) baseline across four T2I backbones, while substantially improving over the corresponding off-the-shelf Qwen3 enhancers. Bold marks the best score within each Qwen3 base-vs-MAPE pair. \textbf{(b)} isolates the contribution of multi-step prompting~(MSP), SFT, and RL on Z-Image-turbo.}
    \label{tab:unigenbench_combined}

    \begin{subtable}{\linewidth}
        \centering
        \caption{Main performance.}
        \label{tab:unigenbench}
        \begin{tabular}{llcc}
            \toprule
            \textbf{T2I Model} & \textbf{Prompt Enhancer} & \textbf{UniGen Short} & \textbf{UniGen Long} \\
            \midrule
            \multirow{6}{*}{Qwen-Image-2512}
              & \ding{55}              & 0.7493 & 0.8869 \\
              & Gemini-3.1-Pro (MSP)   & 0.8669 & 0.8758 \\
              \cmidrule(l){2-4}
              & Qwen3-1.7B             & 0.7645 & 0.8719 \\
              & MAPE (Qwen3-1.7B)      & \textbf{0.8334} & \textbf{0.8624} \\
              \cmidrule(l){2-4}
              & Qwen3-4B               & 0.7055 & 0.7233 \\
              & MAPE (Qwen3-4B)        & \textbf{0.8539} & \textbf{0.8923} \\
            \midrule
            \multirow{6}{*}{Z-Image-turbo}
              & \ding{55}              & 0.6931 & 0.8170 \\
              & Gemini-3.1-Pro (MSP)   & 0.7501 & 0.7674 \\
              \cmidrule(l){2-4}
              & Qwen3-1.7B             & 0.6878 & 0.7947 \\
              & MAPE (Qwen3-1.7B)      & \textbf{0.7716} & \textbf{0.8405} \\
              \cmidrule(l){2-4}
              & Qwen3-4B               & 0.7221 & 0.8081 \\
              & MAPE (Qwen3-4B)        & \textbf{0.8356} & \textbf{0.8512} \\
            \midrule
            \multirow{6}{*}{FLUX.2-klein-4B}
              & \ding{55}              & 0.7489 & 0.8464 \\
              & Gemini-3.1-Pro (MSP)   & 0.8263 & 0.8392 \\
              \cmidrule(l){2-4}
              & Qwen3-1.7B             & 0.6790 & 0.8170 \\
              & MAPE (Qwen3-1.7B)      & \textbf{0.7710} & \textbf{0.8275} \\
              \cmidrule(l){2-4}
              & Qwen3-4B               & 0.4817 & 0.5037 \\
              & MAPE (Qwen3-4B)        & \textbf{0.8042} & \textbf{0.8539} \\
            \midrule
            \multirow{6}{*}{FLUX.2-klein-9B}
              & \ding{55}              & 0.8058 & 0.8792 \\
              & Gemini-3.1-Pro (MSP)   & 0.8553 & 0.8704 \\
              \cmidrule(l){2-4}
              & Qwen3-1.7B             & 0.7093 & 0.8518 \\
              & MAPE (Qwen3-1.7B)      & \textbf{0.8268} & \textbf{0.8578} \\
              \cmidrule(l){2-4}
              & Qwen3-4B               & 0.4777 & 0.5154 \\
              & MAPE (Qwen3-4B)        & \textbf{0.8460} & \textbf{0.8641} \\
            \bottomrule
        \end{tabular}
    \end{subtable}

    \vspace{8pt}

    \begin{subtable}{\linewidth}
        \centering
        \caption{Ablation study on Z-Image-turbo.}
        \label{tab:unigenbench_ablation_partial}
        \begin{tabular}{llcc}
            \toprule
            \textbf{T2I Model} & \textbf{Prompt Enhancer} & \textbf{UniGen Short} & \textbf{UniGen Long} \\
            \midrule
            \multirow{11}{*}{Z-Image-turbo}
              & \ding{55}                  & 0.6931 & 0.8170 \\
              & Gemini-3.1-Pro             & 0.6949 & 0.8170 \\
              & Gemini-3.1-Pro (MSP)       & 0.7501 & 0.7674 \\
              \cmidrule(l){2-4}
              & Qwen3-1.7B                 & 0.6878 & 0.7947 \\
              & Qwen3-1.7B (MSP)           & 0.7383 & 0.8231 \\
              & MAPE $-$ RL (Qwen3-1.7B)   & 0.7623 & 0.8321 \\
              & MAPE (Qwen3-1.7B)          & \textbf{0.7716} & \textbf{0.8405} \\
              \cmidrule(l){2-4}
              & Qwen3-4B                   & 0.7221 & 0.8081 \\
              & Qwen3-4B (MSP)             & 0.7721 & 0.8290 \\
              & MAPE $-$ RL (Qwen3-4B)     & 0.8233 & 0.8502 \\
              & MAPE (Qwen3-4B)            & \textbf{0.8356} & \textbf{0.8512} \\
            \bottomrule
        \end{tabular}
    \end{subtable}
\end{table}

\textbf{Result and Analysis.} Quantitative results of MAPE on the challenging UniGenBench dataset are detailed in \Cref{tab:unigenbench}. 
We find again that deploying off-the-shelf SLMs as prompt enhancers typically degrades the performance of most T2I models. In contrast, MAPE consistently out-performs both un-enhanced baselines and SLM baselines, achieving parity with the Gemini-3.1-Pro strong baseline. For a rigorous comparison, the Gemini-3.1-Pro row in \Cref{tab:unigenbench} uses our decomposed router--rewriter--composer prompting (MSP) rather than one-shot prompting; the ablation in \Cref{tab:unigenbench_ablation_partial} confirms that this decomposition empirically outperforms one-shot prompting for both Gemini-3.1-Pro and Qwen3, justifying its use as the strong baseline. This is a central empirical result of the paper: with appropriate agent decomposition and post-training, an SLM-based prompt enhancer can approach the behavior of much stronger closed-source prompt enhancers on challenging image-generation tasks. A complete breakdown across all T2I models, including both one-shot and MSP variants of Gemini-3.1-Pro and Qwen3, is provided in Appendix~\ref{appendix:ablation}.

\Cref{tab:unigenbench_ablation_partial} presents an ablation study using the Z-Image-turbo text-to-image model to isolate the effects of MAPE's architectural and training components. Performance gains are observed when transitioning from the one-shot Qwen3 baseline to a decomposed prompting structure (MSP, i.e., MAPE's router--rewriter--composer pipeline without any post-training), with further improvements provided by supervised fine-tuning (MAPE~$-$~RL) and reinforcement learning (full MAPE). 
In Appendix~\ref{appendix:distribution}, we show that SFT injects more details into the prompts, while RL can help the models/pipelines adapt to the downstream image models.

The qualitative results in \Cref{fig:qual_samples_mape} at the front of the paper provide additional insight into why MAPE helps. Compared with single-pass SLMs, MAPE is better at grounding fine-grained details into the prompt. This is particularly important for images that require imagination, where the user request does not directly specify all visual content and the enhancer must infer plausible but relevant details. MAPE also shows advantages on prompts involving spatial reasoning and logical reasoning, where multiple semantic constraints must be retained and combined consistently. Another common benefit is that MAPE is less likely to forget key information from the original prompt when elaborating additional content. Together, these patterns suggest that the main value of MAPE is not simply producing longer prompts, but producing prompts whose structure better preserves constraints, grounds missing details, and supports more coherent reasoning about the desired image. Overall, these results show that multi-agent prompt enhancement becomes increasingly important as image-generation tasks grow more complex, and that careful architectural design together with post-training is crucial for realizing its benefits. This result echoes recent findings on test-time compute scaling \cite{snell2024scaling}. 

\subsection{Multi-Agent Prompt Enhancer for Image Editing}

\textbf{Settings.} We finally extend APE to image editing to test whether the multi-agent prompt-enhancement framework generalizes beyond image generation. We instantiate the vision-language prompt enhancer with Qwen3-VL-4B-Instruct, and evaluate across three downstream image editing models: FLUX.2-klein-4B, FLUX.2-klein-9B, and Qwen-Image-Edit. We use ImgEdit \cite{yeimgedit} as the benchmark. As in the image-generation setting, we compare both one-shot SLMs and MSP. MAPE follows the same router–rewriter–composer design, but the editing scenario adds the original image to the input of each step, and places additional emphasis on deciding whether prompt rewriting is needed at all, since some edit instructions are already explicit while others require substantial grounding and elaboration. We then apply GDPO post-training to improve the small-model MAPE and better align it with the multi-dimensional reward structure of image editing. The rewards are based on the testpoints given by ImgEdit, evaluated by GPT-4.1 \cite{OpenAI_GPT4_2026}.

\textbf{Result and Analysis.} Overall, we find that MSP (i.e., MAPE’s router-rewriter–composer pipeline) generally improves image editing performance, showing that prompt enhancement also transfers well to the editing setting. However, the relative behavior of different prompt enhancers is more nuanced than in image generation. In particular, MSP with Gemini-3.1-Pro performs worse than the one-shot Gemini-3.1-Pro, suggesting that a more complex multi-agent decomposition does not automatically help when the underlying enhancer is already very strong. A plausible reason is that multi-stage processing can introduce error propagation and information confusion, especially when the edit instruction is already sufficiently clear and decomposition adds unnecessary intermediate transformations.

In contrast, the MSP with Qwen3-VL-4B-Instruct is more beneficial, especially on more complicated editing tasks such as Extract, Adjust, and Style. These categories typically require richer grounding, more detailed reasoning about the desired transformation, and better preservation of relevant visual context, all of which benefit from explicit decomposition into field selection, field-specific rewriting, and composition. However, indiscriminately invoking the multi-agent pipeline can hurt on simpler categories such as Remove, where the edit intent is often direct and local, and excessive rewriting may introduce redundancy or distract from the core operation. Specifically, we test the MSP variant of our agentic framework, which forces rewriting on every input with Qwen3-VL-4B and FLUX.2-klein-4B, and we find that even though MSP can improve the "Extract", "Style" and "Adjust" performance (+1.67, +0.22, +0.06), it can hurt on some other tasks such as "Remove", and "Action" (-0.9, -0.1). This observation highlights a key difference between generation and editing: in editing, more prompt enhancement is not always better, and the system must decide when rewriting is necessary and when it should remain minimal.

This motivates the design choice of allowing the router to choose whether to rewrite or not, rather than forcing multi-agent rewriting on every input. With this more flexible design, GDPO post-training further improves the performance of MSP with SLMs, while also helping to balance performance across editing categories. Instead of over-optimizing for difficult tasks at the expense of simpler ones, the post-trained model learns a better trade-off between categories that need more grounding and categories that benefit from restraint. Overall, these results demonstrate two important points: first, the proposed multi-agent prompt-enhancement framework generalizes effectively to image editing; and second, in editing tasks, rewrite-field selection and rewrite necessity are especially important, making routing decisions a central part of successful prompt enhancement. The complete qualitative examples are shown in Appendix~\ref{appendix:visual_mape_edit}.

\begin{table}[t!]
    \centering
    \scriptsize
    \renewcommand{\arraystretch}{1.05}
    \caption{Qualitative comparison of prompt enhancement by one-shot Qwen3-VL-4B and MAPE (Qwen3-VL-4B) for the image editing task. More examples in Appendix~\ref{appendix:visual_mape_edit}}
    \label{tab:qual_samples_multi_edit_main}
    \begin{tabular}{L{0.08\textwidth} C{0.10\textwidth} L{0.16\textwidth} C{0.10\textwidth} L{0.24\textwidth} C{0.10\textwidth}}
        \toprule
        \multirow{2}{*}{\textbf{Instruction}} &
        \multirow{2}{*}{\makecell[c]{\textbf{Original}\\\textbf{Image}}} &
        \multicolumn{2}{c}{\textbf{Qwen3-VL-4B}} &
        \multicolumn{2}{c}{\textbf{MAPE (Qwen3-VL-4B)}} \\
        \cmidrule(lr){3-4} \cmidrule(lr){5-6}
        & & \multicolumn{1}{c}{\textbf{Prompt}} & \textbf{Image} & \multicolumn{1}{c}{\textbf{Prompt}} & \textbf{Image} \\
        \midrule

        Change the house in the picture to a coastal setting with a beach and ocean in the background. &
        \includegraphics[width=0.10\textwidth]{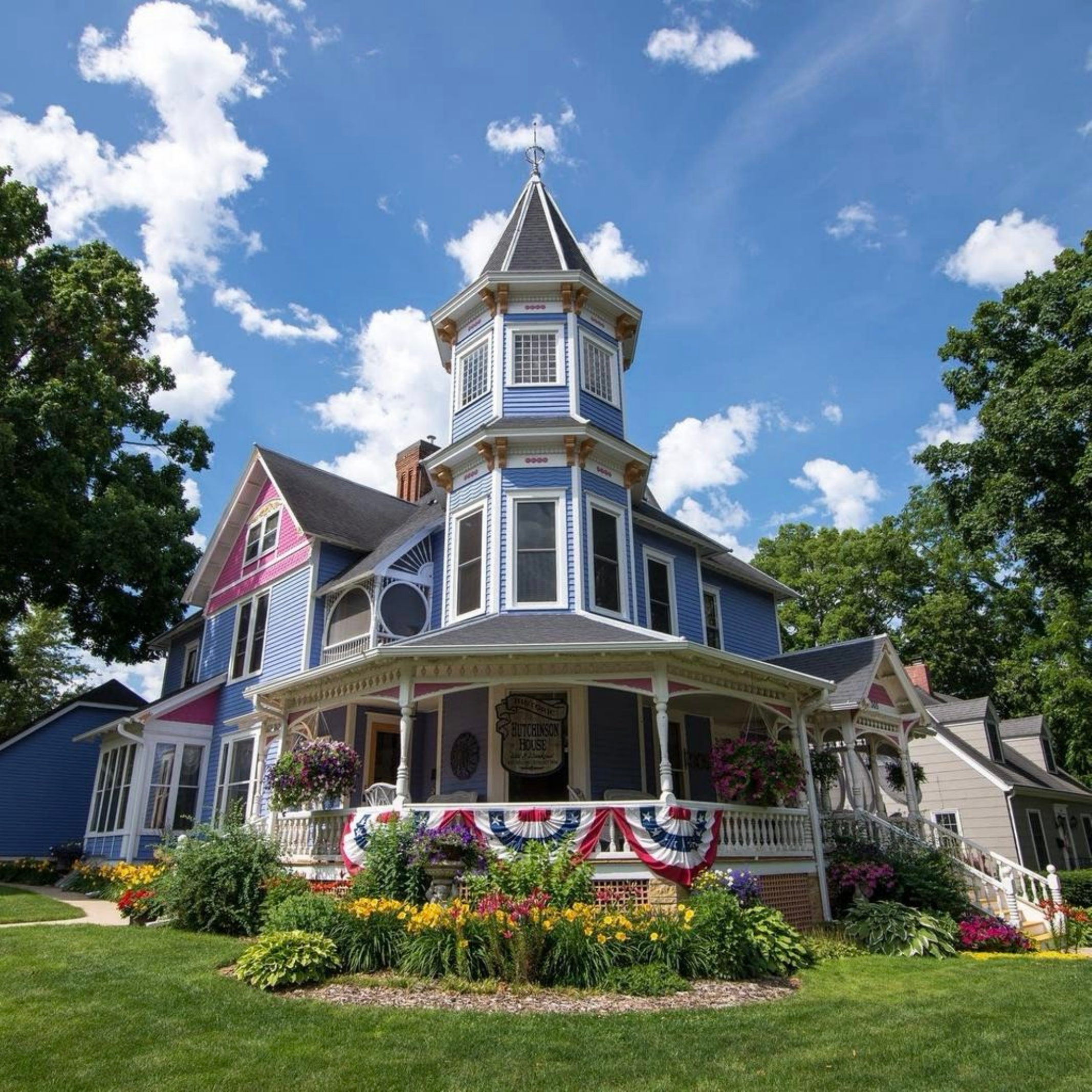} &
        Replace the current Victorian house with a coastal cottage featuring whitewashed walls and a sloped roof, set against a sandy beach and ocean backdrop. Keep the vibrant blue sky, fluffy clouds, and surrounding green trees. Maintain the porch with hanging flowers and red, white, and blue bunting. &
        \includegraphics[width=0.10\textwidth]{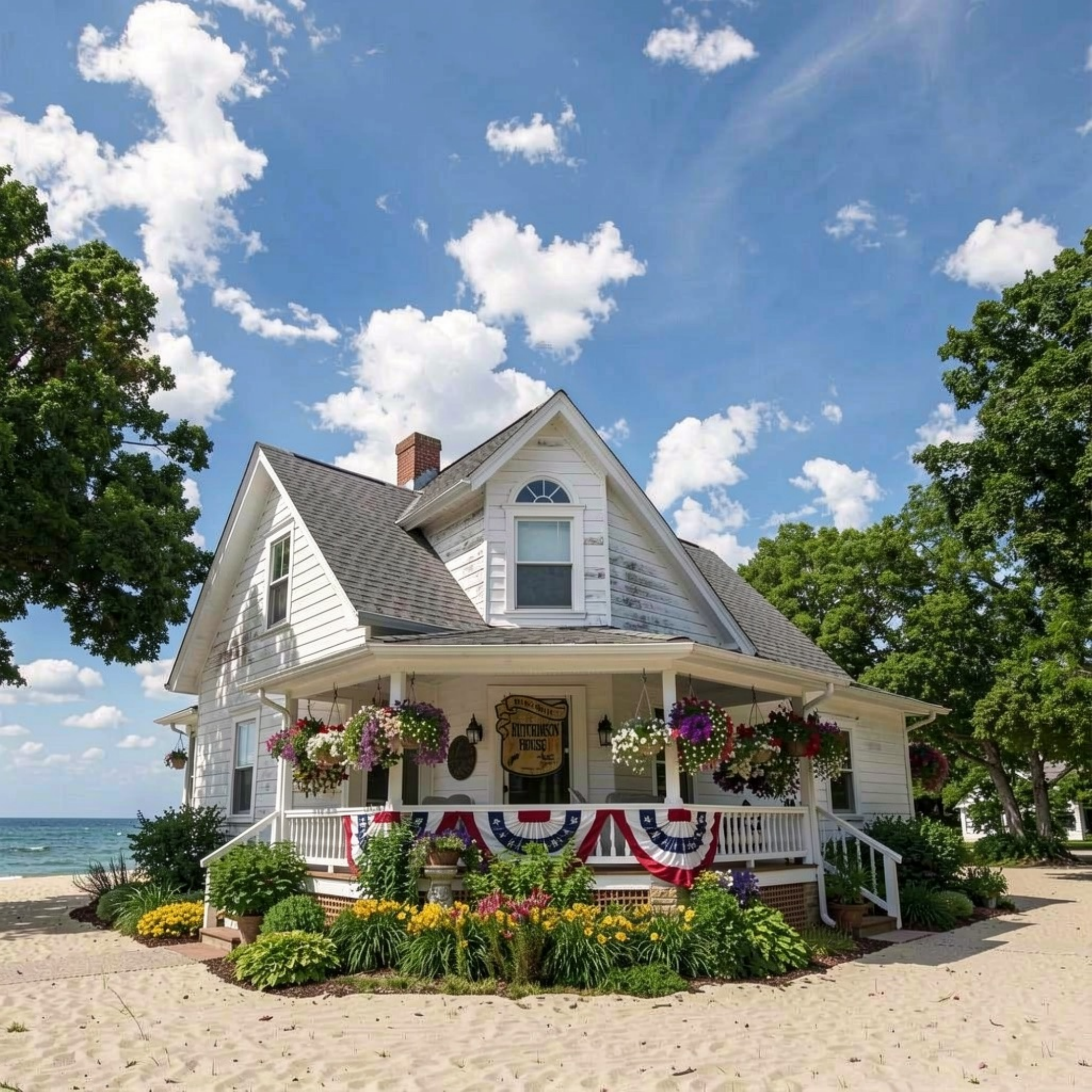} &
        Replace the entire background behind and around the Victorian house with a vibrant coastal scene: wide sandy beach meeting calm blue ocean under bright sky with scattered clouds, distant sea foam and small boats, sparse salt-tolerant vegetation and driftwood. \textbf{Keep the house, porch, flag bunting, flower arrangements, and garden beds unchanged in position and detail}. Match lighting, shadows, and color grading to original, preserving shallow depth of field and spatial perspective. &
        \includegraphics[width=0.10\textwidth]{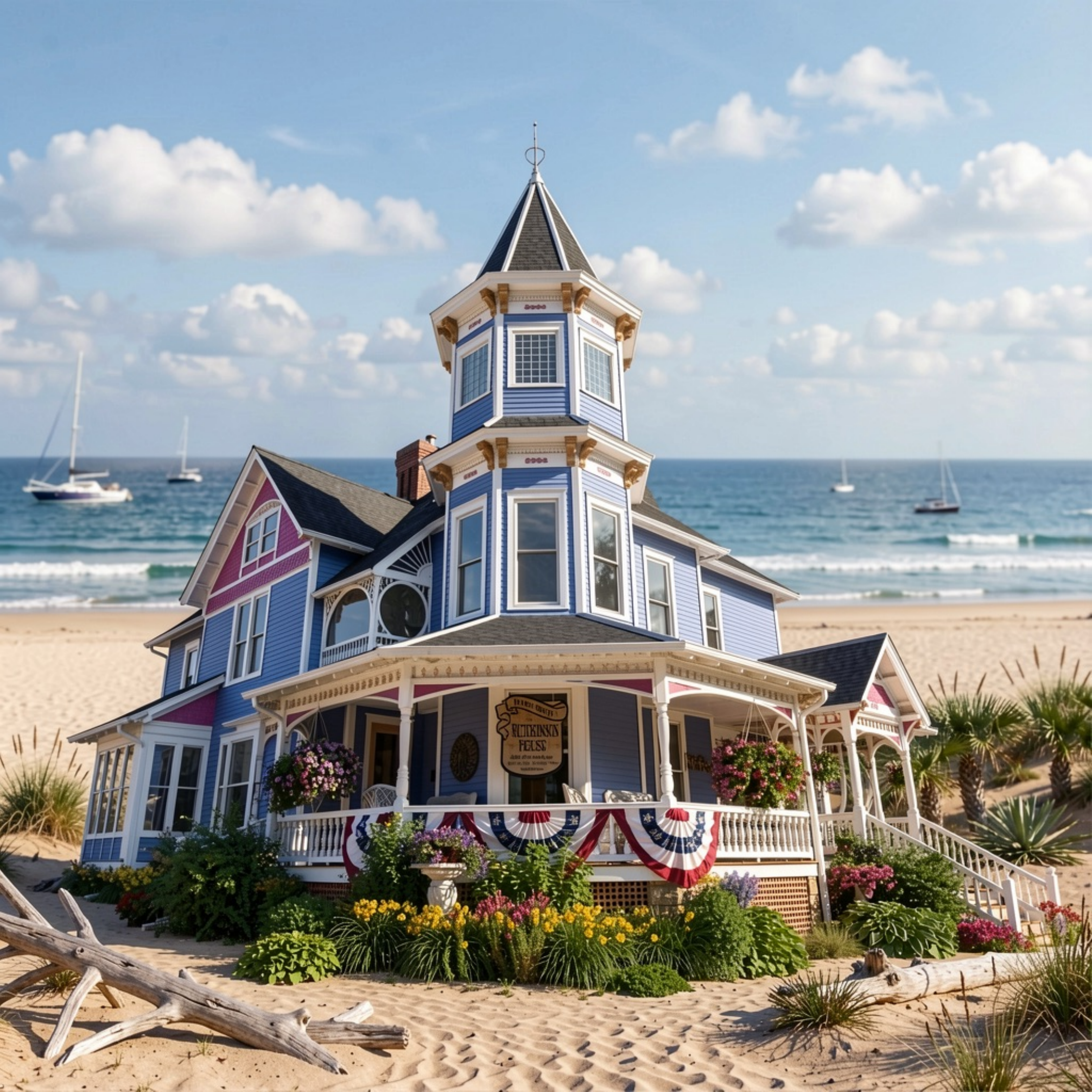} \\
        \bottomrule
    \end{tabular}
\end{table}

\begin{table}[H]
    \centering
    \tiny
    \caption{ImgEdit performance.}
    \label{tab:imgedit}
    \renewcommand{\arraystretch}{1.05}
    \resizebox{\linewidth}{!}{%
    \begin{tabular}{c|c|c|c|c|c|c|c|c|c|c|c}
        \hline
        I2I Model & Prompt Enhancer & Overall & Add & Adjust & Remove & Replace & Style & Action & Extract & Background & Compose \\
        \hline
        \multirow{5}{*}{FLUX.2-klein-4B} & \ding{55} & 3.85 & 4.28 & 4.24 & 3.52 & 4.19 & 4.91 & 4.69 & 1.94 & 4.2 & 2.71 \\
        & Gemini-3.1-Pro & 4.38 & 4.48 & 4.44 & 4.66 & 4.53 & 4.72 & 4.58 & 3.82 & 4.4 & 3.12 \\
        & Gemini-3.1-Pro (MSP) & 3.85 & 4.39 & 4.07 & 3.75 & 4.28 & 4.79 & 4.49 & 1.87 & 4.17 & 3.22 \\\cline{2-12}
        & Qwen3-VL-4B & 3.87 & 4.36 & 4.17 & 4.26 & 4.11 & 4.67 & 4.41 & 1.91 & 4.12 & 2.91 \\
        & Qwen3-VL-4B (MSP) & 4.06 & 4.35 & 4.23 & 3.36 & 3.98 & 4.89 & 4.31 & 3.58 & 4.2 & 2.91 \\
        & MAPE (Qwen3-VL-4B) & \textbf{4.15} & 4.46 & 4.25 & 4.19 & 4.02 & 4.81 & 4.57 & 3.58 & 4.12 & 2.77 \\
        \hline
        \multirow{6}{*}{FLUX.2-klein-9B} & \ding{55} & 4.07 & 4.44 & 4.22 & 4.33 & 4.54 & 4.94 & 4.32 & 2.23 & 4.39 & 3.14 \\
        & Gemini-3.1-Pro & 4.46 & 4.55 & 4.31 & 4.62 & 4.76 & 4.94 & 4.82 & 3.96 & 4.43 & 2.9 \\
        & Gemini-3.1-Pro (MSP) & 4.08 & 4.4 & 4.33 & 4.43 & 4.45 & 4.91 & 4.39 & 2.34 & 4.31 & 2.88 \\\cline{2-12}
        & Qwen3-VL-4B & 4.03 & 4.52 & 4.27 & 4.53 & 4.44 & 4.88 & 4.41 & 1.98 & 4.21 & 2.87 \\
        & Qwen3-VL-4B (MSP) & 4.25 & 4.39 & 4.24 & 4.26 & 4.39 & 4.91 & 4.48 & 3.61 & 4.26 & 3.19 \\
        & MAPE (Qwen3-VL-4B) & \textbf{4.32} & 4.32 & 4.21 & 4.28 & 4.55 & 4.84 & 4.68 & 4.01 & 4.22 & 3.0 \\
        \hline
        \multirow{6}{*}{Qwen-Image-Edit} & \ding{55} & 3.98 & 4.32 & 3.66 & 3.61 & 4.15 & 4.56 & 3.9 & 4.04 & 3.78 & 2.64 \\
        & Gemini-3.1-Pro & 4.0 & 4.33 & 3.87 & 3.63 & 4.33 & 4.66 & 4.05 & 3.63 & 3.85 & 2.9 \\
        & Gemini-3.1-Pro (MSP) & 3.99 & 4.22 & 3.71 & 3.67 & 4.14 & 4.58 & 4.02 & 3.87 & 3.96 & 2.91 \\\cline{2-12}
        & Qwen3-VL-4B & 3.93 & 4.11 & 3.85 & 3.67 & 4.24 & 4.65 & 3.84 & 3.5 & 3.87 & 2.71 \\
        & Qwen3-VL-4B (MSP) & 4.0 & 4.35 & 3.7 & 3.71 & 4.35 & 4.57 & 3.79 & 3.79 & 3.94 & 2.58 \\
        & MAPE (Qwen3-VL-4B) & \textbf{4.01} & 4.28 & 3.8 & 3.63 & 4.34 & 4.58 & 4.14 & 3.83 & 3.9 & 2.64 \\
        \hline
    \end{tabular}%
    }
\end{table}

%% file: conclusion.tex
\section{Conclusion}
We presented APE, a unified prompt-enhancement framework for image generation and editing, with SAPE for single-agent rewriting and MAPE for router–rewriter–composer enhancement. Experiments show that post-training only the prompt enhancer, while keeping the image model frozen, yields strong gains; for harder compositional tasks, MAPE further improves detail grounding, information preservation, and reasoning over complex prompts. APE demonstrates that small, carefully designed prompt enhancers can achieve strong performance, sometimes approaching closed-source systems. A key limitation is that GRPO and GDPO use only final downstream rewards; future work will explore finer-grained credit assignment across pipeline stages and extensions to video generation and editing.

%% file: appendix.tex
The appendix is organized as follows:
\begin{itemize}
    \item Appendix~\ref{appendix:visual} shows more qualitative examples of our APE method, where Appendix~\ref{appendix:visual_rl_gen} shows the effectiveness of RL and Appendix~\ref{appendix:visual_mape_edit} shows the visualization of MAPE on the ImgEdit image editing tasks. Qualitative MAPE results on UniGenBench image generation are shown at the front of the paper in \Cref{fig:qual_samples_mape}.
    \item Appendix~\ref{appendix:distribution} shows the prompt distribution difference between prompt enhancer and the effectiveness of different training stages for MAPE. 
    \item Appendix~\ref{appendix:ablation} shows the ablation studies on the image generation task with all 4 image generation models, and demonstrates the benefit of our framework and training method.
    \item Appendix~\ref{appendix:rl_detail} explains the details about RL methods, including the mathematical formulation of GRPO (Appendix~\ref{appendix:grpo}) and GDPO (Appendix~\ref{appendix:gdpo}) with VLM scoring.
    \item Appendix~\ref{appendix:prompts} lists the detailed prompts used for our MAPE framework, including the router, rewriter, composer for image generation (Appendix~\ref{appendix:prompts_gen}) and image editing (Appendix~\ref{appendix:prompts_edit}). 
    \item Appendix~\ref{appendix:related} goes through more related works with more details.
\end{itemize}
\section{Visualization}
\label{appendix:visual}

\subsection{Visualization of RL effect for Image Generation}
\label{appendix:visual_rl_gen}
\begin{table}[H]
    \centering
    \scriptsize
    \setlength{\tabcolsep}{3pt}
    \renewcommand{\arraystretch}{1.05}
    \caption{Qualitative comparison of prompt enhancement before and after GRPO.}
    \label{tab:qual_samples_single_appendix}
    \begin{tabular}{L{0.10\textwidth} L{0.28\textwidth} C{0.12\textwidth} L{0.28\textwidth} C{0.12\textwidth}}
        \toprule
        \textbf{Original Prompt} &
        \makecell[c]{\textbf{Enhancer Prompt}} &
        \makecell[c]{\textbf{Generated}\\\textbf{Image}} &
        \makecell[c]{\textbf{Enhancer Prompt}\\\textbf{after GRPO}} &
        \makecell[c]{\textbf{Generated Image}\\\textbf{after GRPO}} \\
        \midrule

        A medieval painting of the wifi not working. &
        A medieval villagescape bathed in a dim, shadowy glow, with tall, weathered stone towers and intricate archways. A lone figure, cloaked in tattered robes, is holding a flickering Wi-Fi signal, their hands trembling as a device on the table blinks erratically. The air is thick with the scent of parchment and damp earth, and a lone torch casts long shadows across the cobblestone path. Below, a group of villagers huddle near a flickering lamp, their faces lit by the dim light, while a distant signal crackles through the trees. The scene is set in a time of uncertainty, capturing both the medieval setting and the technical struggle of a modern problem. &
        \includegraphics[width=0.12\textwidth]{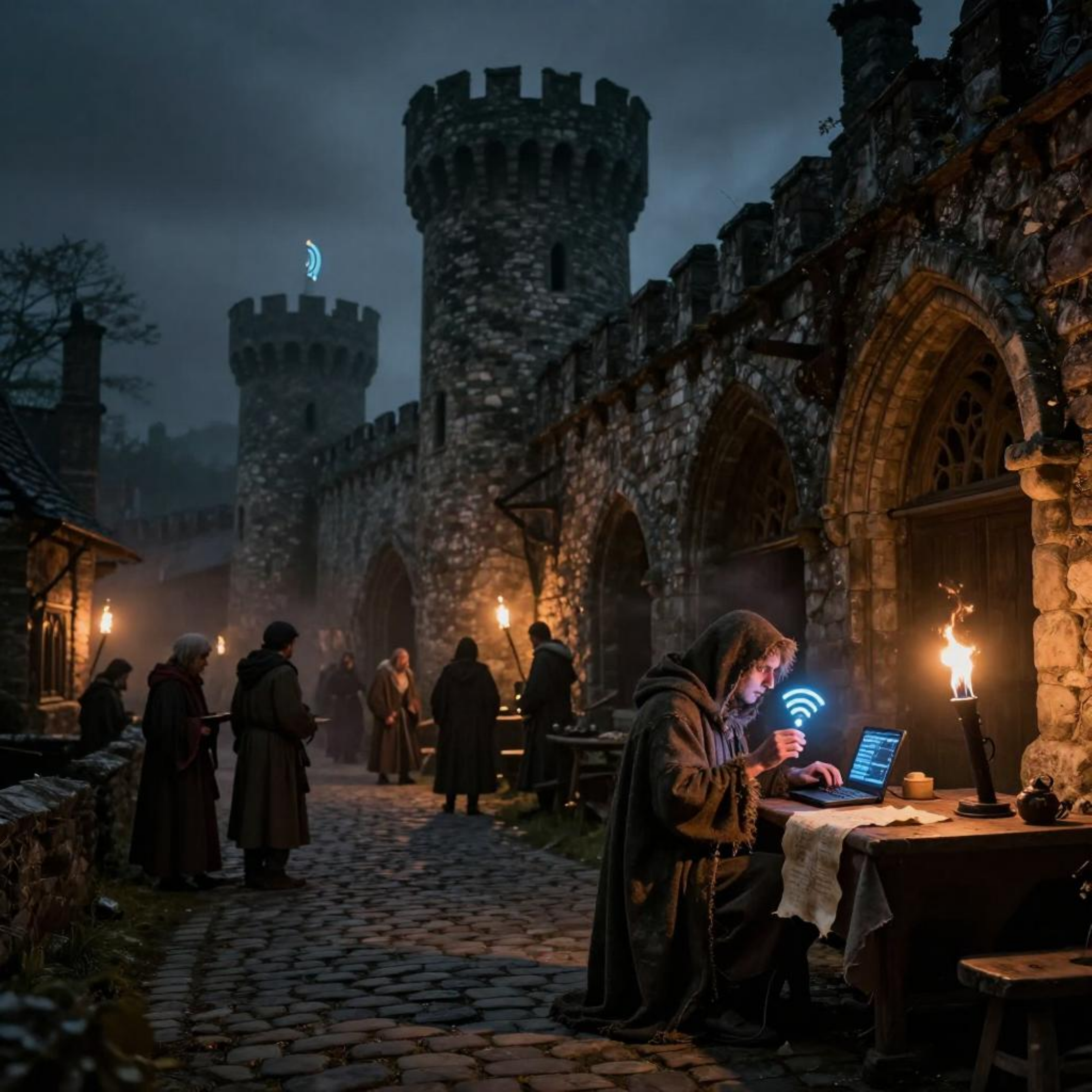} &
        A medieval painting of the Wi-Fi not working, set in a dark, atmospheric setting, featuring a medieval-style figure struggling with a broken Wi-Fi signal. \textbf{The scene captures the technical frustration of the Wi-Fi not working, with details such as the signal strength, the environment, and the characters \textbackslash ' reactions. The painting emphasizes the challenges of modern technology in a medieval context, blending historical elements with the technical issue of connectivity.} &
        \includegraphics[width=0.12\textwidth]{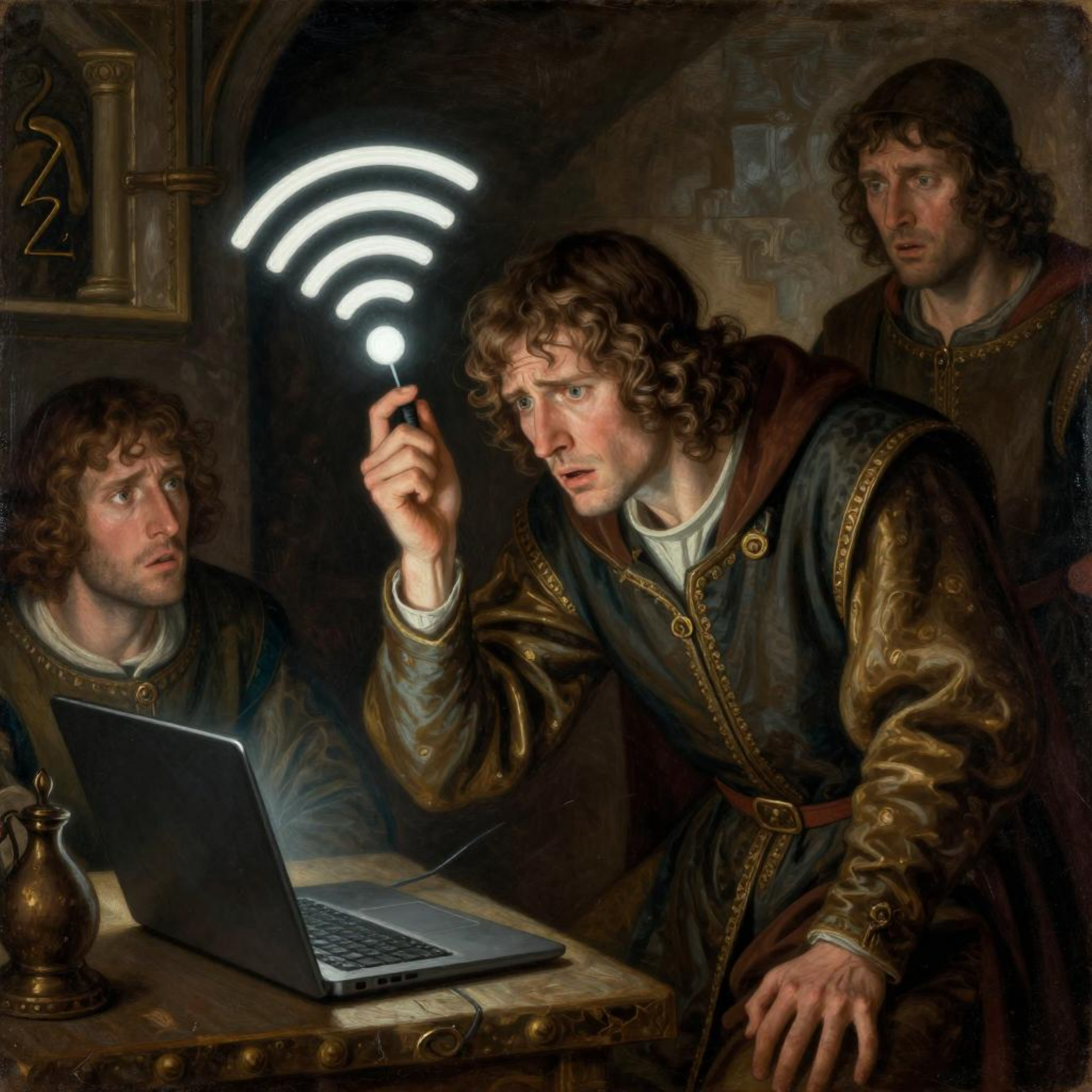} \\
        \midrule

        A tennis racket underneath a traffic light. &
        A tennis racket is displayed on a red traffic light at a bustling city street, its polished leather surface catching the golden sunlight. The light, a large circular prism with a red dot at its center, casts soft shadows across the pavement. The racket is positioned directly above the light, its frame slightly bent at the wrist, while a group of pedestrians passes by, their laughter echoing against the rhythmic motion of the racket. The scene is captured in a dynamic urban setting with vibrant colors and a sense of motion. &
        \includegraphics[width=0.12\textwidth]{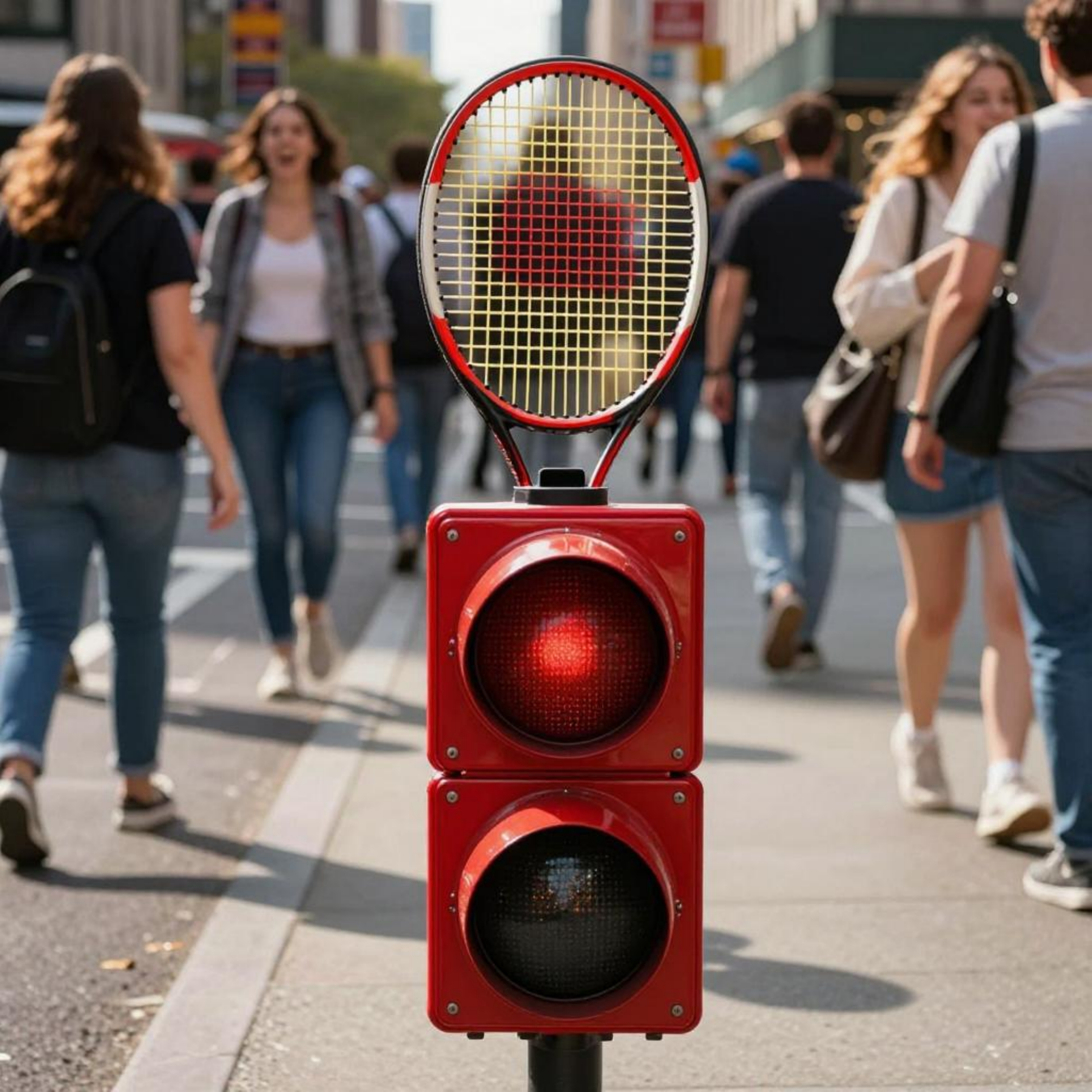} &
        \textbf{A tennis racket lies beneath a traffic light}, casting a warm, reflective glow over the surroundings. The racket\textbackslash 's sleek, modern design contrasts with the traffic light\textbackslash 's intricate details, creating a dynamic visual harmony. The scene is set in a bustling urban environment, emphasizing the tension between the racket and the traffic light as key elements of the scene. &
        \includegraphics[width=0.12\textwidth]{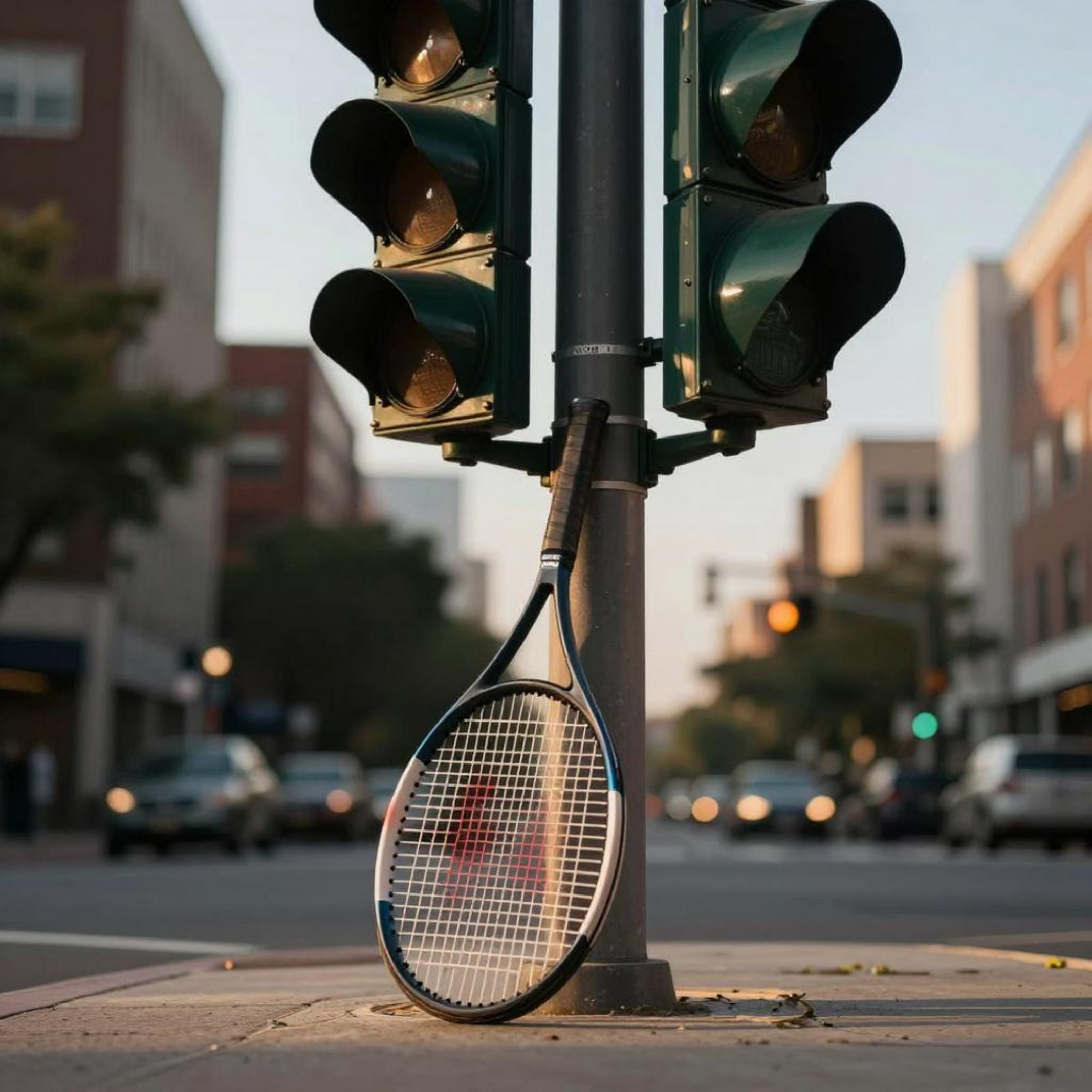} \\
        \midrule

        A pizza on the right of a suitcase. &
        A pizza on the right side of a suitcase, its plate filled with golden crust and toppings, while a backpack rests beside it. The suitcase is a sleek black leather bag with a vintage logo, and the room is a cozy living room with soft lighting. The pizza is centered on a wooden table, its edges slightly curved, and the toppings are colorful, adding to the visual appeal. The background features a dimly lit kitchen with scattered items, creating a warm and inviting atmosphere. &
        \includegraphics[width=0.12\textwidth]{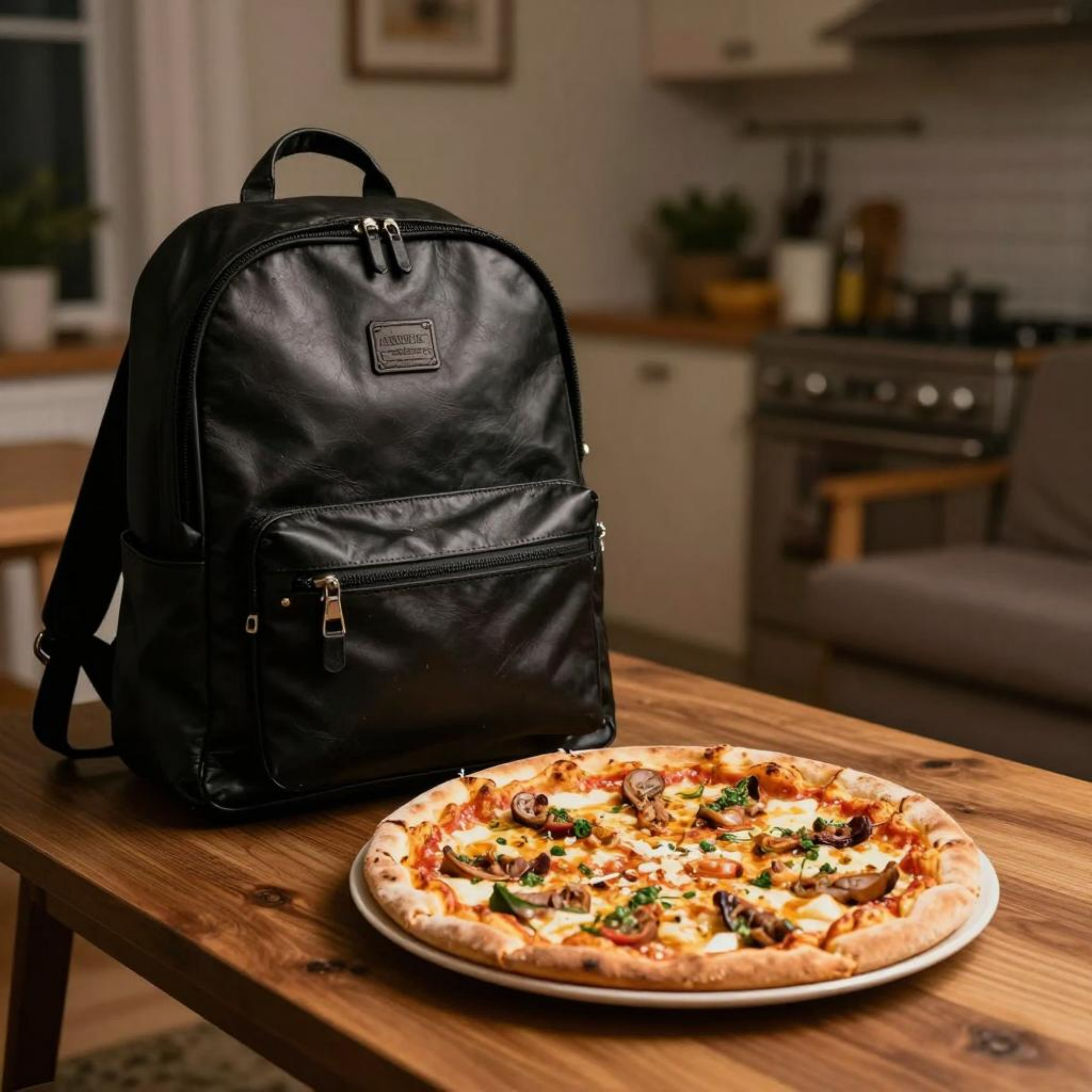} &
        A pizza on the right side of a suitcase, \textbf{placed on a suitcase rack or in a travel bag, with the pizza's plate and toppings visible}. The suitcase is described as a travel item, possibly on a bed or in a suitcase, and the scene is set in a casual, everyday setting. The pizza is highlighted with its texture and presentation, while the suitcase adds a stylish or practical element to the image. &
        \includegraphics[width=0.12\textwidth]{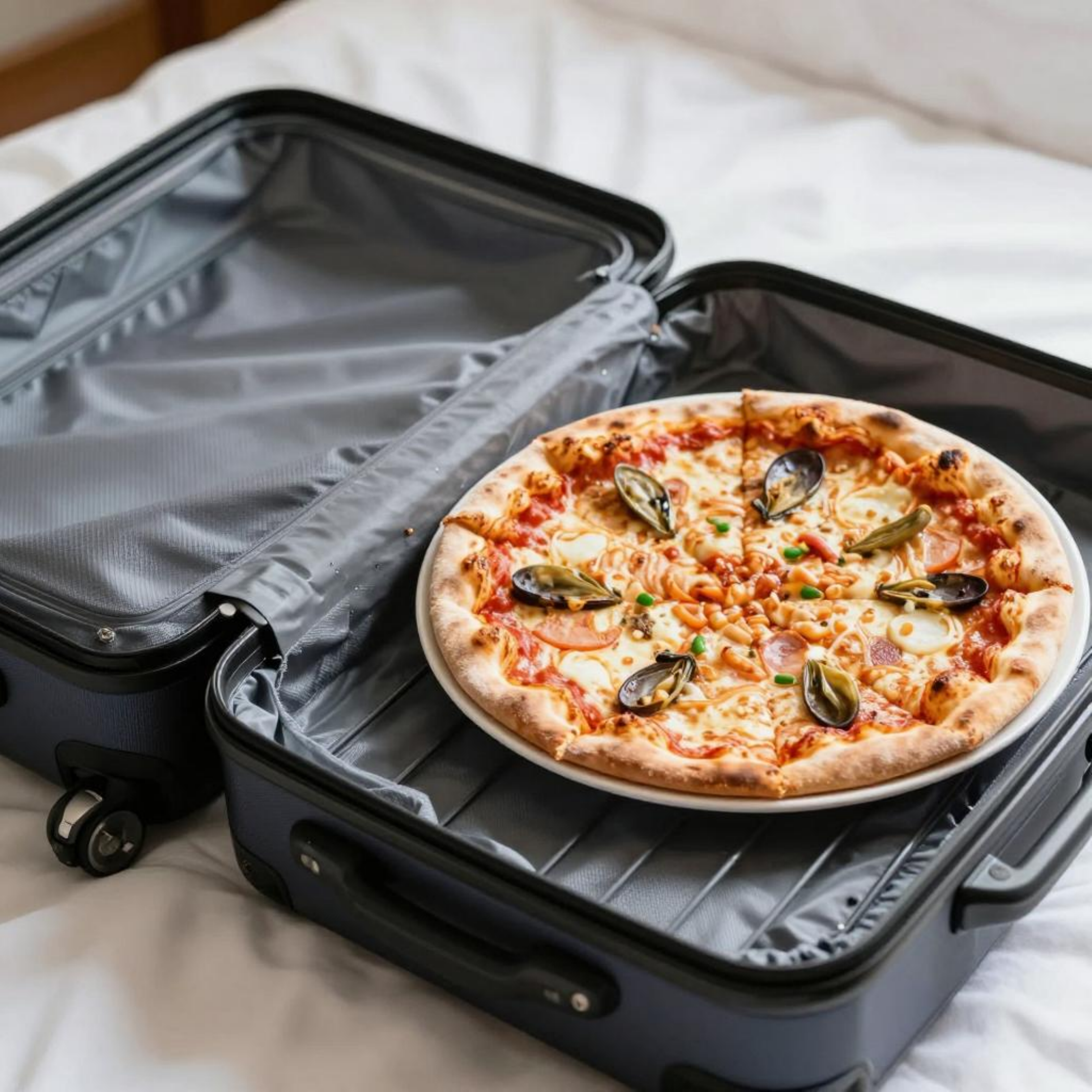} \\
        \bottomrule
    \end{tabular}
\end{table}

\clearpage
\subsection{Visualization of Image Editing by MAPE}
\label{appendix:visual_mape_edit}

\begin{table}[H]
    \centering
    \scriptsize
    \setlength{\tabcolsep}{3pt}
    \renewcommand{\arraystretch}{1.05}
    \caption{Qualitative comparison of prompt enhancement by one-shot Qwen3-VL-4B and MAPE (Qwen3-VL-4B) for the image editing task. (Part I.)}
    \label{tab:qual_samples_multi_edit_appendix}
    \begin{tabular}{L{0.12\textwidth} C{0.12\textwidth} L{0.16\textwidth} C{0.12\textwidth} L{0.24\textwidth} C{0.12\textwidth}}
        \toprule
        \multirow{2}{*}{\textbf{Instruction}} &
        \multirow{2}{*}{\makecell[c]{\textbf{Original}\\\textbf{Image}}} &
        \multicolumn{2}{c}{\textbf{Qwen3-VL-4B}} &
        \multicolumn{2}{c}{\textbf{MAPE (Qwen3-VL-4B)}} \\
        \cmidrule(lr){3-4} \cmidrule(lr){5-6}
        & & \multicolumn{1}{c}{\textbf{Prompt}} & \textbf{Image} & \multicolumn{1}{c}{\textbf{Prompt}} & \textbf{Image} \\
        \midrule

        Transfer the image into a stained-glass cathedral-window style. &
        \includegraphics[width=0.11\textwidth]{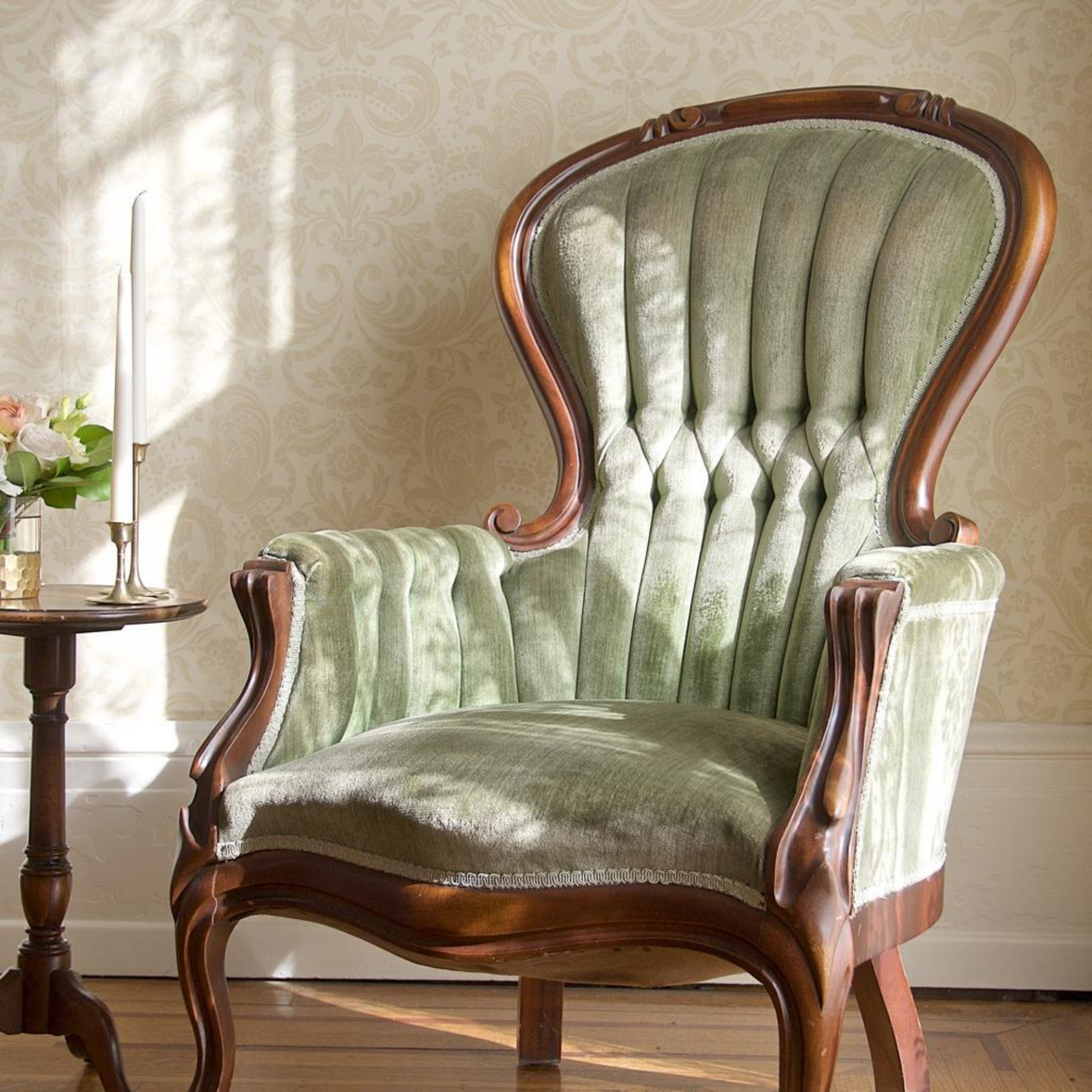} &
        Render the green velvet armchair and patterned wallpaper as stained-glass cathedral windows: preserve the chair's wooden frame and tufted upholstery as leaded glass panes, convert the wallpaper into intricate, colorful stained-glass patterns, and maintain the side table with candles as a complementary stained-glass element. &
        \includegraphics[width=0.11\textwidth]{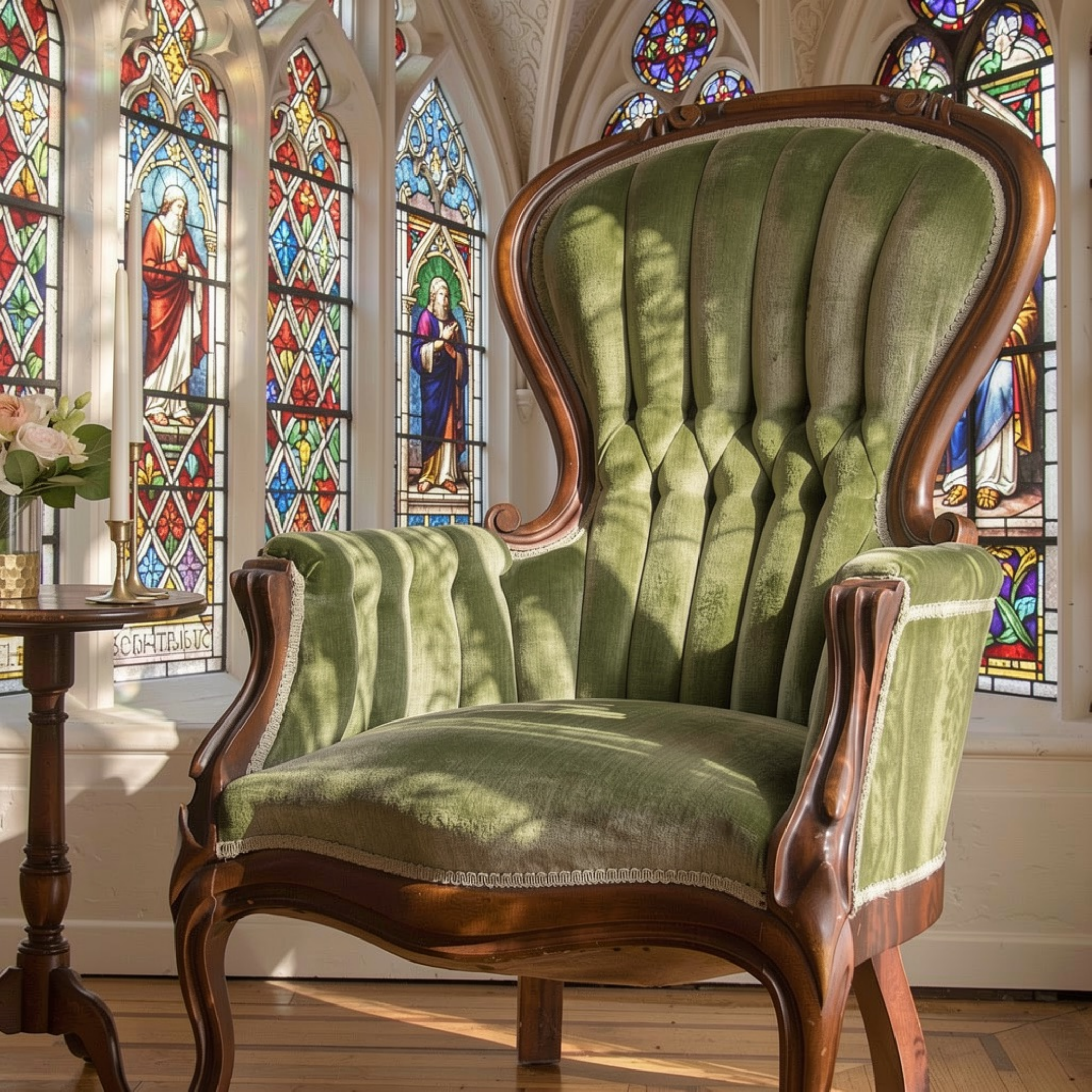} &
        \textbf{Transform the entire image into stained-glass cathedral-window style: segmented panels in rich emerald, gold, amber, and cobalt blue with sharp lead lines, translucent glass textures, and high-contrast directional lighting casting vibrant, geometric shadows}. Preserve spatial arrangement of chair, side table, candles, flowers, and wallpaper — only apply style transformation. &
        \includegraphics[width=0.11\textwidth]{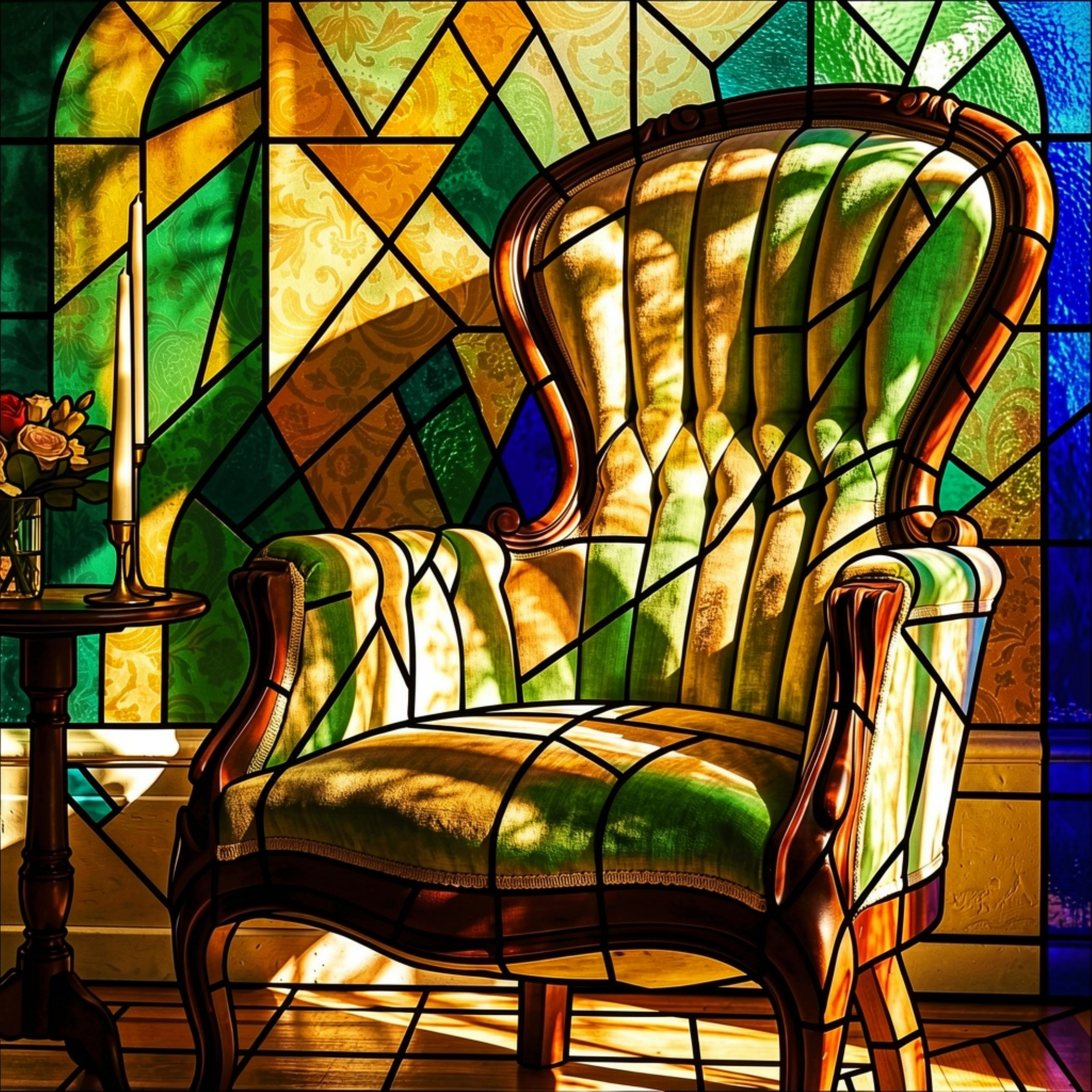} \\
        \midrule

        Remove the tiger in the water. &
        \includegraphics[width=0.11\textwidth]{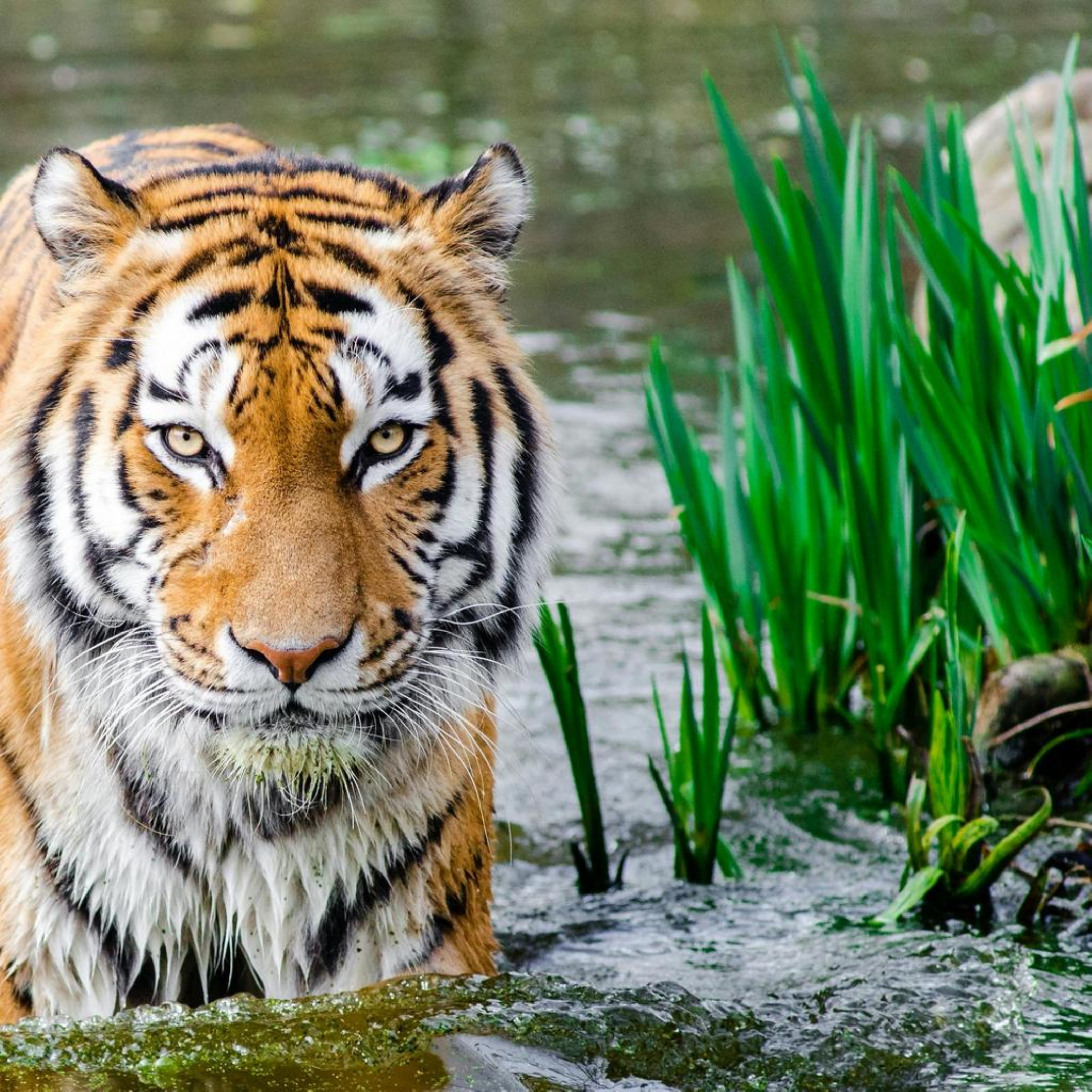} &
        Remove the tiger from the water, keeping its face and fur pattern intact. Maintain the background water and green reeds. Do not alter the lighting or composition; preserve the natural setting and the tiger’s gaze. &
        \includegraphics[width=0.11\textwidth]{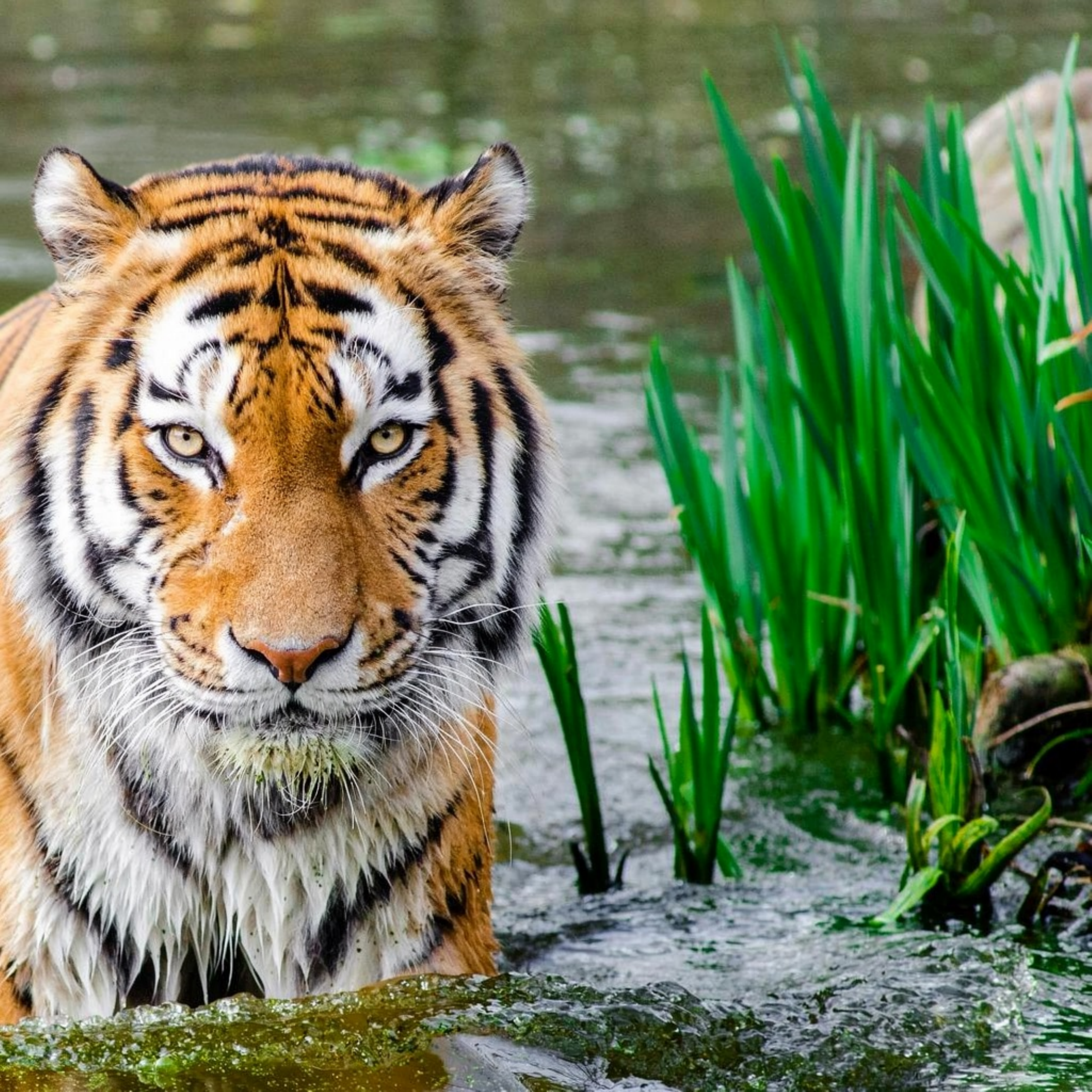} &
        Remove the tiger entirely from the foreground, \textbf{including its head, shoulders, and submerged lower body, while preserving the water’s surface and natural ripples}. Smoothly blend the area with surrounding water and green reeds on the right, maintaining lighting and texture consistency. Keep the blurred water background and reeds unchanged; ensure no wet fur or droplets remain. &
        \includegraphics[width=0.11\textwidth]{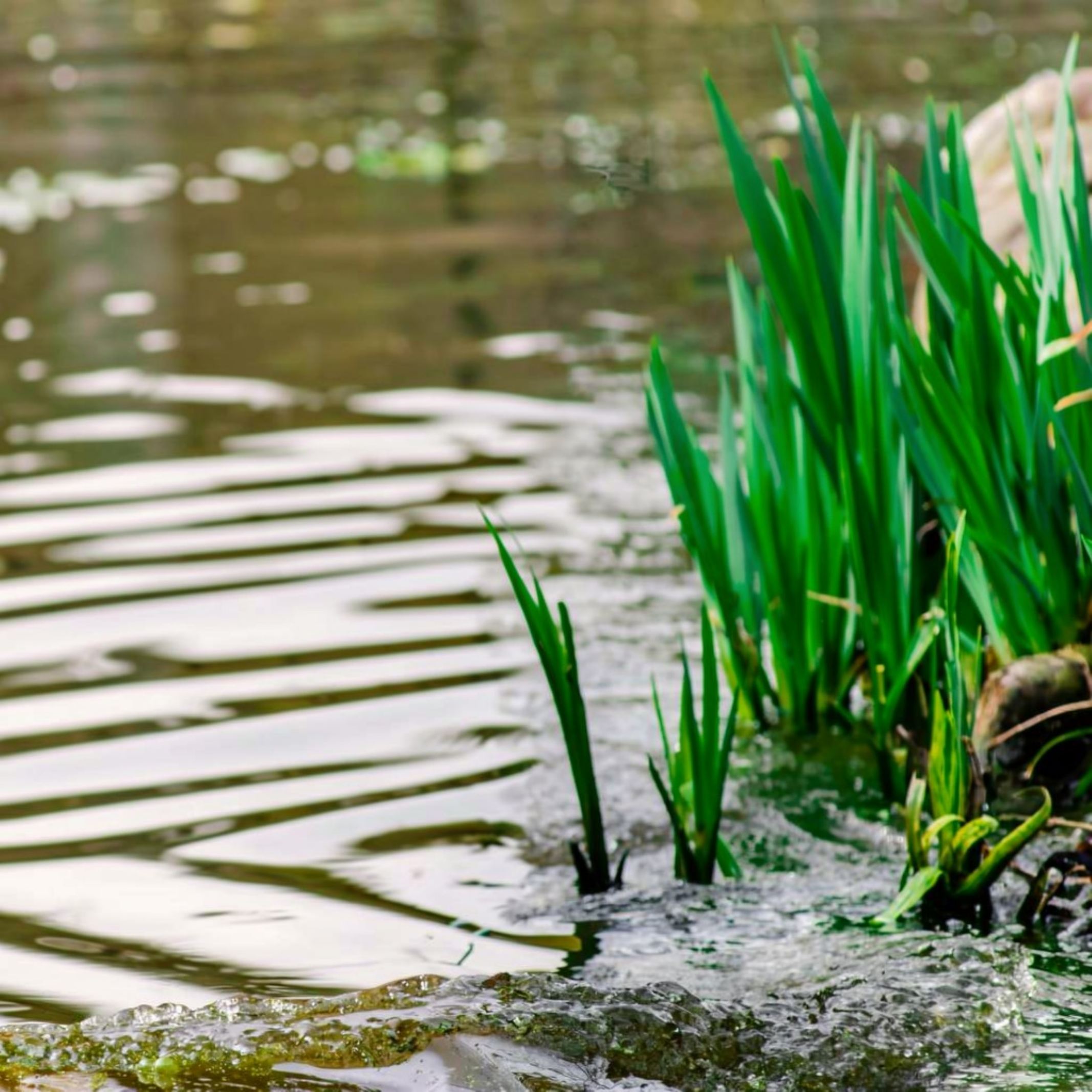} \\
        \midrule

        Add a small stone cottage in the middle of the flower field, near the bottom center of the image. &
        \includegraphics[width=0.11\textwidth]{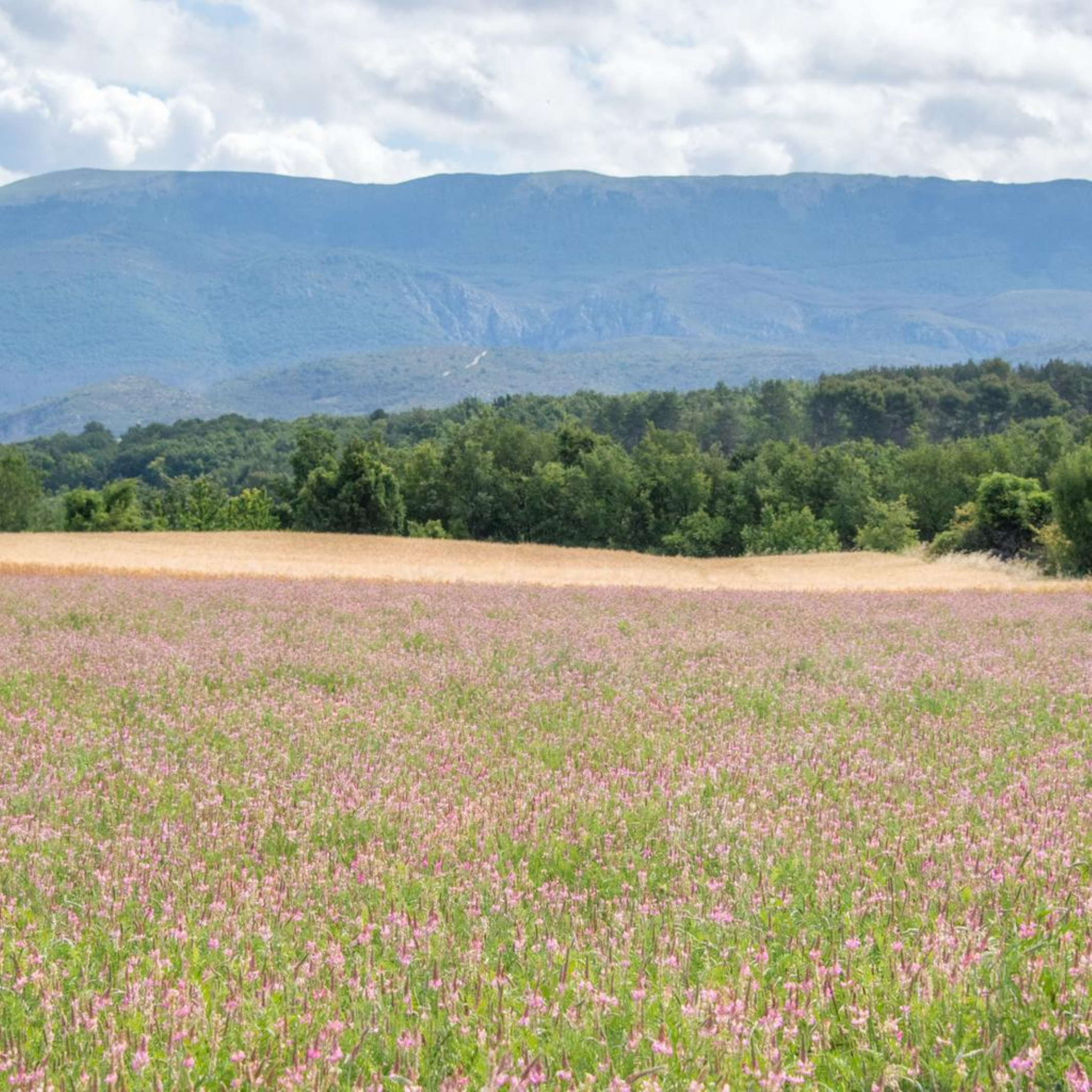} &
        Add a small stone cottage with a thatched roof, centered near the bottom of the flower field, blending into the greenery with no visible foundation or shadow. Keep the existing landscape, lighting, and distant mountains unchanged. &
        \includegraphics[width=0.11\textwidth]{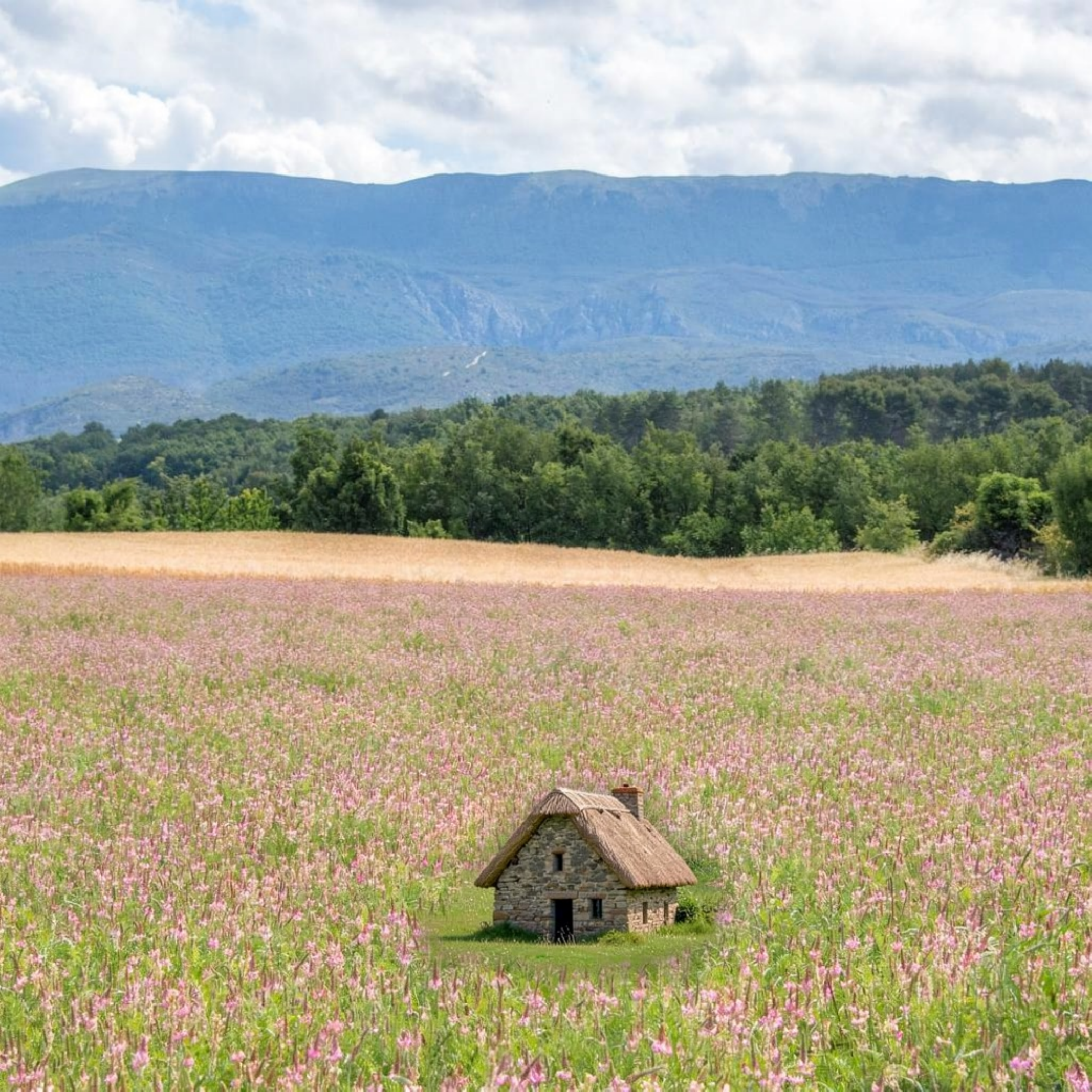} &
        \textbf{Add a medium-sized, rustic stone cottage with a thatched roof—grayish-brown walls textured with moss, a slightly tilted, crooked roof, single wooden-framed window, and a left-side chimney—positioned near bottom center, slightly right of vertical center, nestled among dense pink flowers}. Blend with natural lighting and shadows; maintain perspective alignment with horizon and trees. Keep surrounding flowers undisturbed, preserving the field’s depth and the distant forest/mountains under partly cloudy sky. &
        \includegraphics[width=0.11\textwidth]{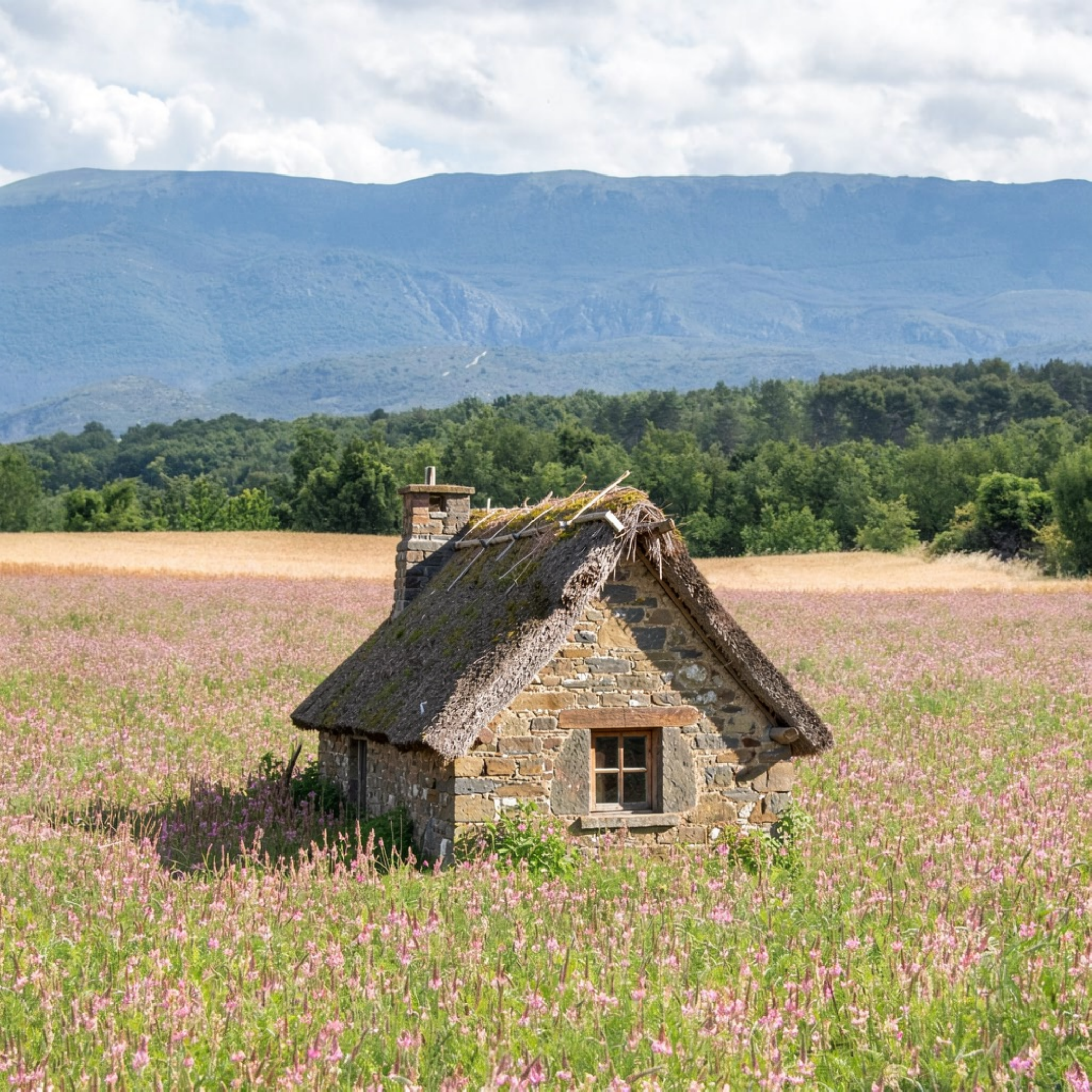} \\
        \midrule

        Change the surface color of the object to a solid shade of blue. &
        \includegraphics[width=0.11\textwidth]{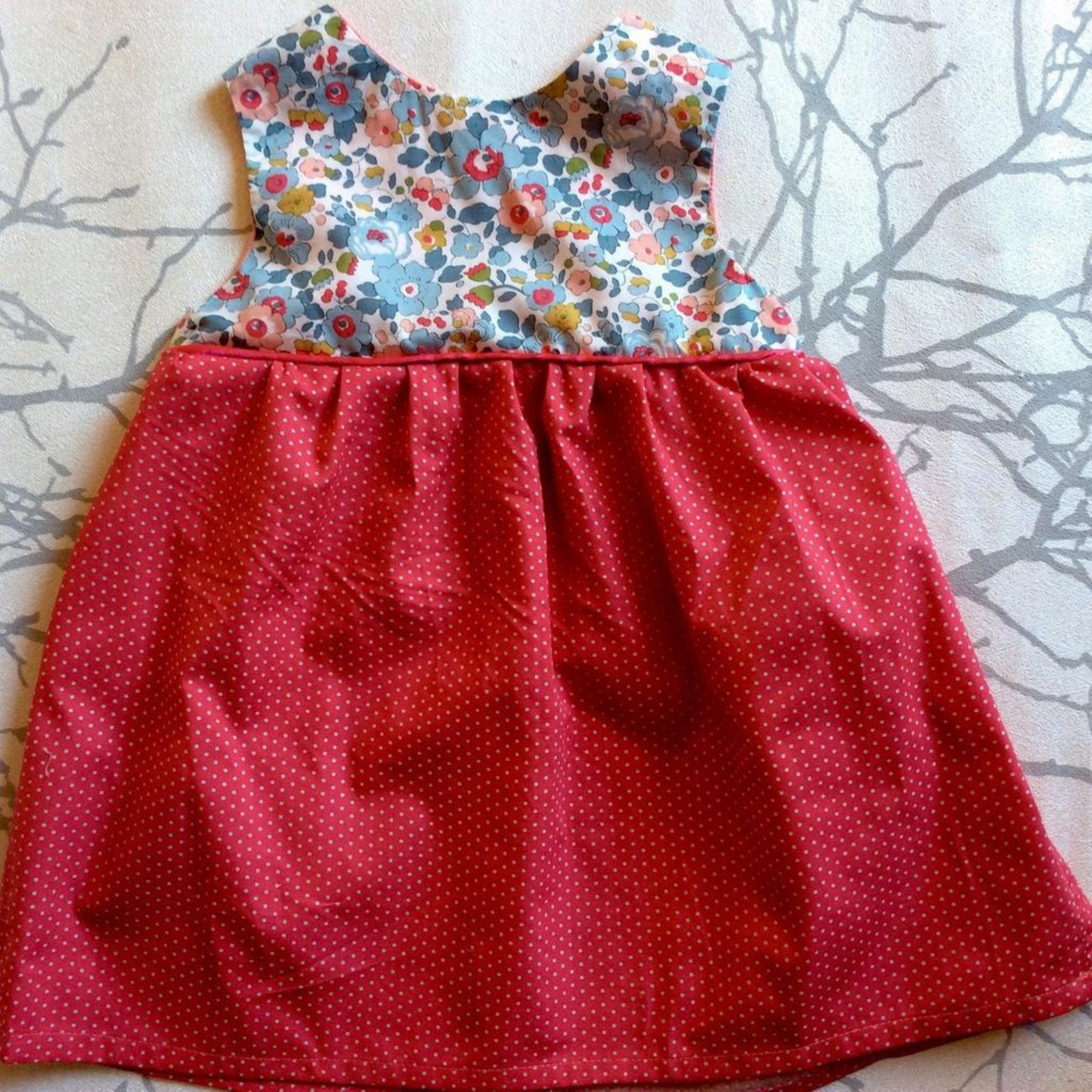} &
        Change the background surface from gray branches to a solid, uniform blue. Keep the dress’s floral top and red polka-dot skirt unchanged, maintaining their current patterns and colors. &
        \includegraphics[width=0.11\textwidth]{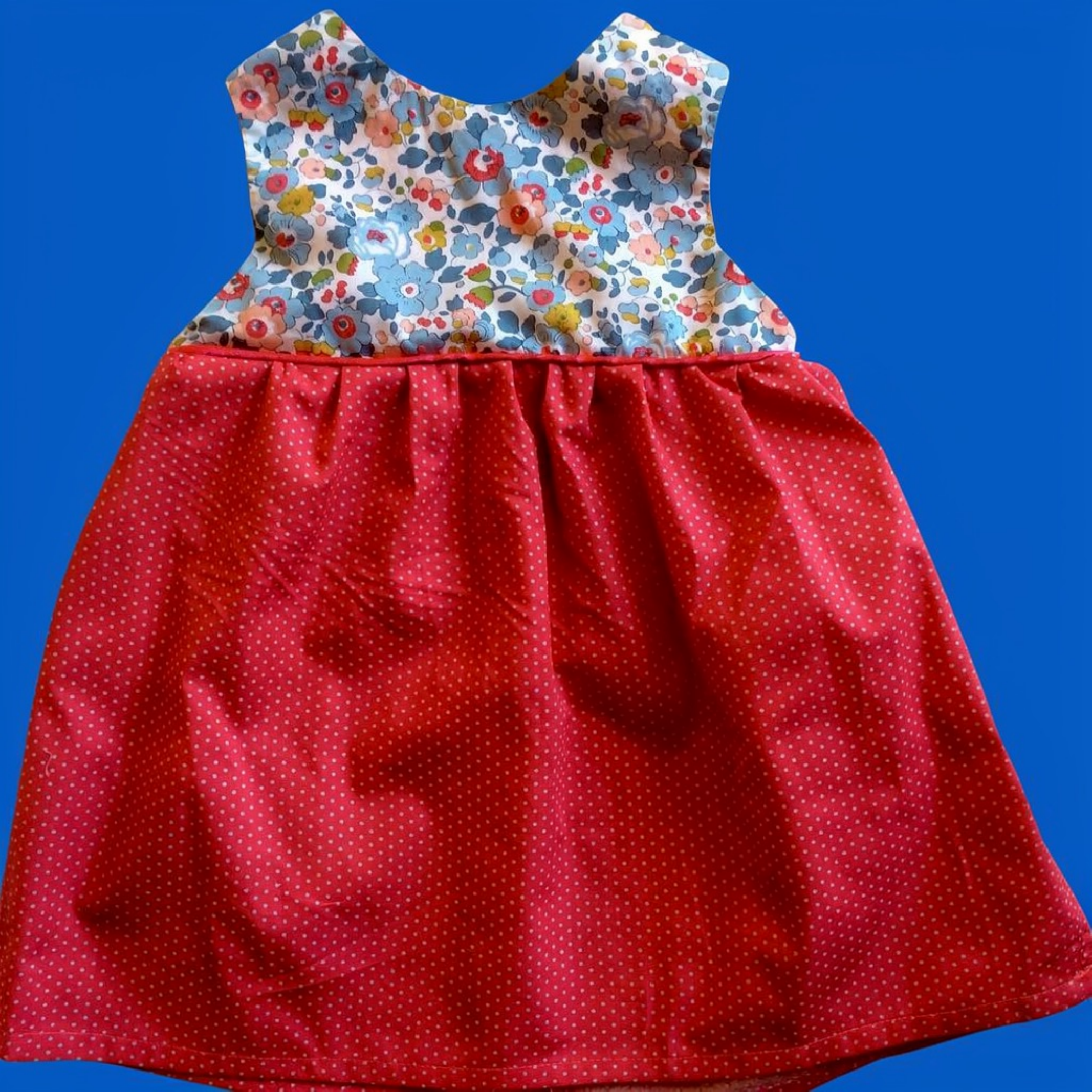} &
        \textbf{Change the entire dress surface to a uniform solid blue, removing all floral patterns and red-and-white polka dots from both top and bottom sections. Keep the dress’s shape, sleeveless design, rounded neckline, and gathered waist intact}. Maintain its flat lay on the light-colored surface with gray branch pattern. No other elements or colors should be altered. &
        \includegraphics[width=0.11\textwidth]{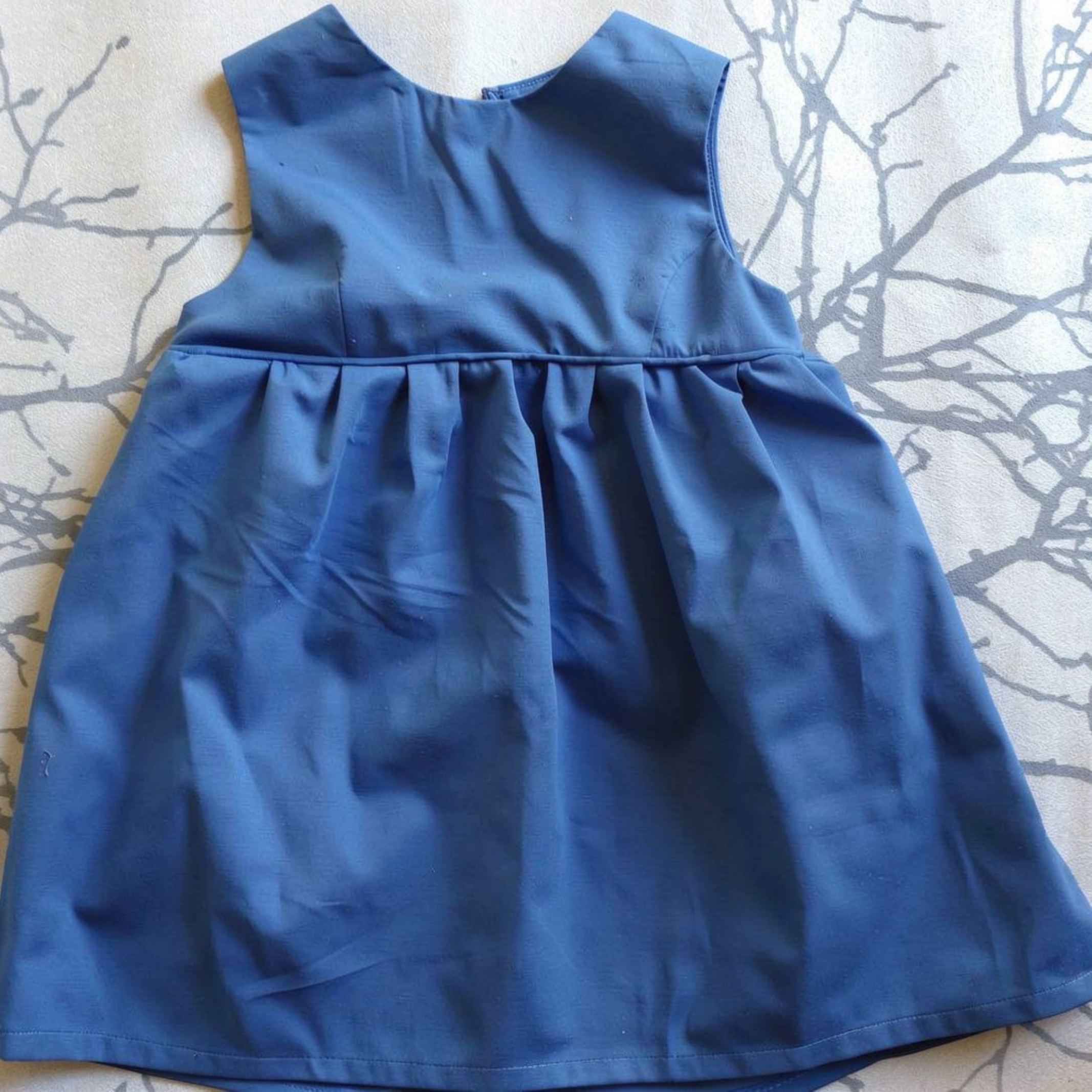} \\
        \bottomrule
    \end{tabular}
\end{table}

\clearpage

\begin{table}[H]
    \ContinuedFloat
    \centering
    \scriptsize
    \setlength{\tabcolsep}{3pt}
    \renewcommand{\arraystretch}{1.05}
    \caption[]{(Continued.) Part II.}
    \begin{tabular}{L{0.12\textwidth} C{0.12\textwidth} L{0.16\textwidth} C{0.12\textwidth} L{0.24\textwidth} C{0.12\textwidth}}
        \toprule
        \multirow{2}{*}{\textbf{Instruction}} &
        \multirow{2}{*}{\makecell[c]{\textbf{Original}\\\textbf{Image}}} &
        \multicolumn{2}{c}{\textbf{Qwen3-VL-4B}} &
        \multicolumn{2}{c}{\textbf{MAPE (Qwen3-VL-4B)}} \\
        \cmidrule(lr){3-4} \cmidrule(lr){5-6}
        & & \multicolumn{1}{c}{\textbf{Prompt}} & \textbf{Image} & \multicolumn{1}{c}{\textbf{Prompt}} & \textbf{Image} \\
        \midrule

        Replace the priest in the image with a large cactus plant. &
        \includegraphics[width=0.11\textwidth]{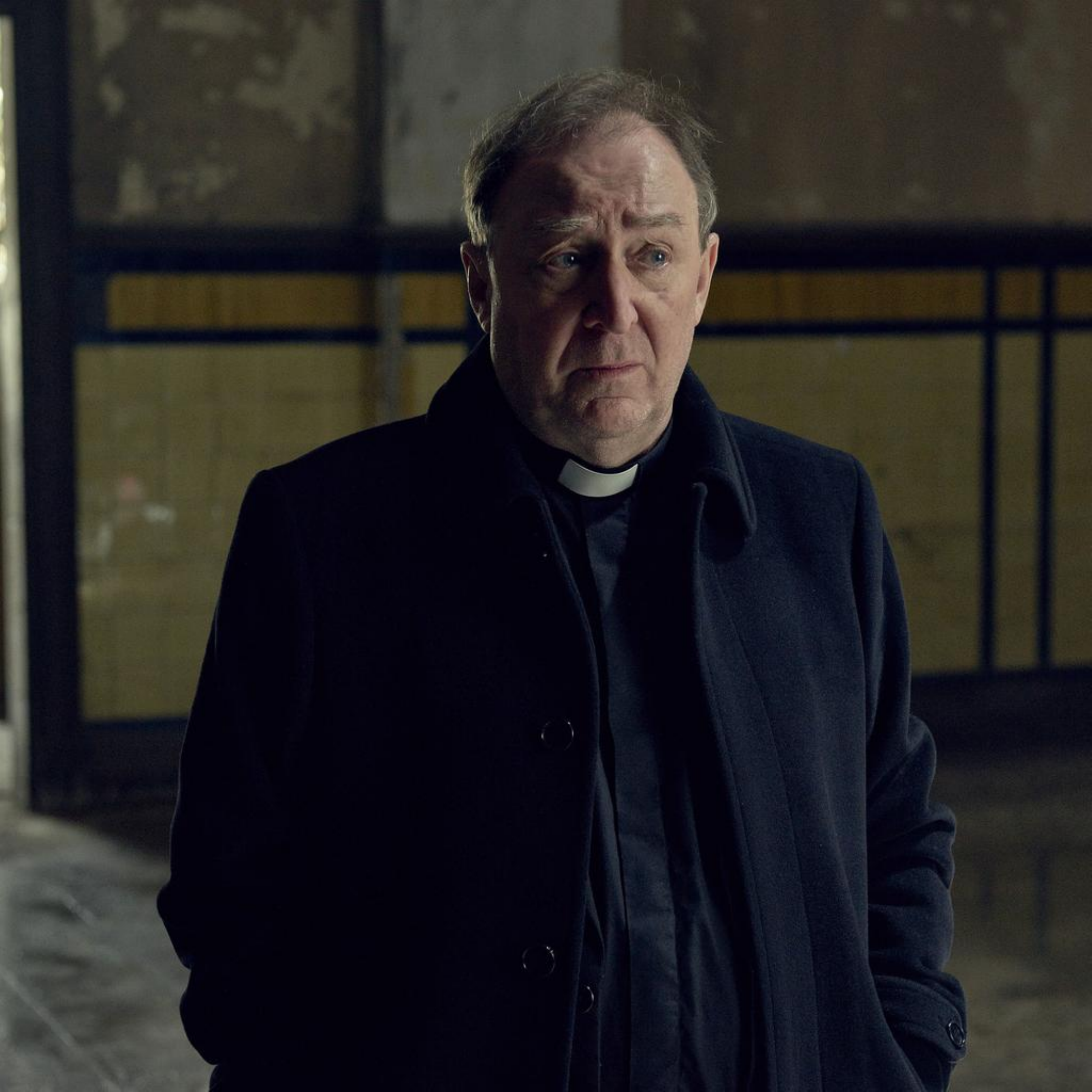} &
        Replace the priest with a large, spiky cactus plant, keeping the same dark coat and background architecture. Maintain the dim, moody lighting and composition, ensuring the cactus occupies the central frame with its textured, green body and sharp spines. &
        \includegraphics[width=0.11\textwidth]{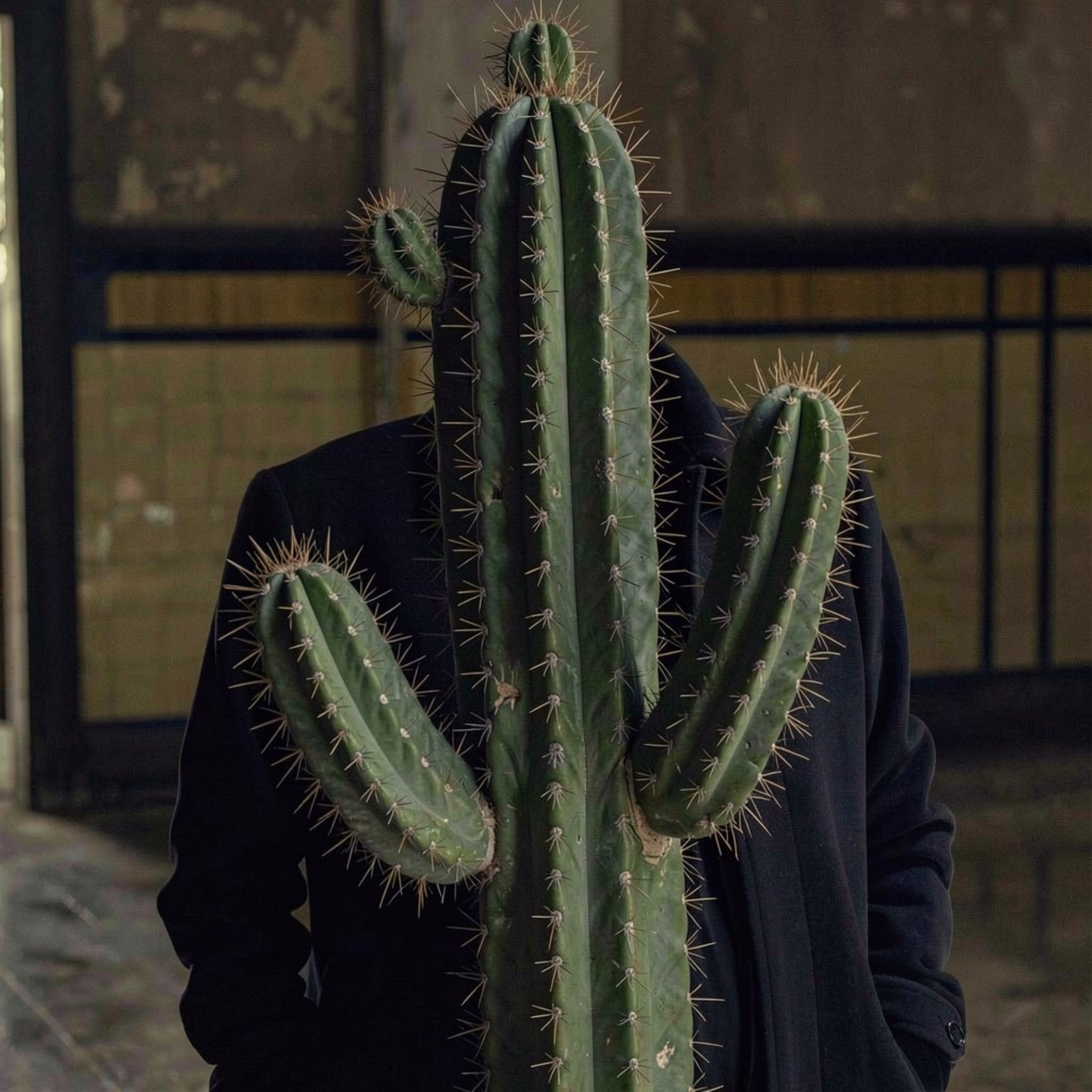} &
        \textbf{Replace the central priest—dark coat, clerical collar, upright posture—with a large, spiky green cactus, matching his height and width, centered in frame}. Maintain dim, moody industrial lighting and gritty cinematic style; preserve shadows and highlights. Keep background elements (yellow-striped wall, metal railings) unchanged. The cactus stands upright, facing forward, with rough, segmented texture and brownish tones, contrasting against the dark setting while retaining visual balance. &
        \includegraphics[width=0.11\textwidth]{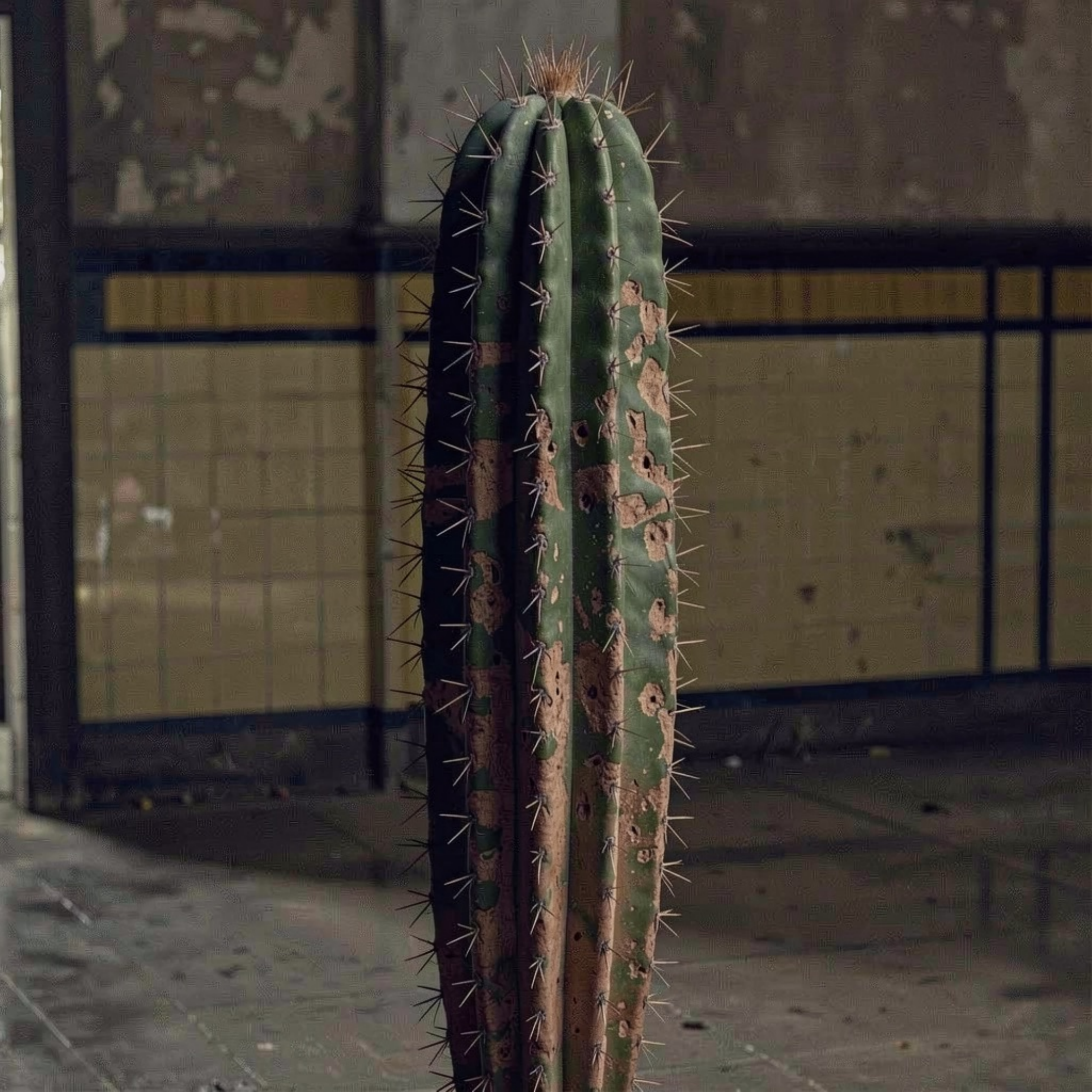} \\
        \midrule

        Remove the person on the left side of the image, and adjust the brightness of the background to make it appear lighter. &
        \includegraphics[width=0.11\textwidth]{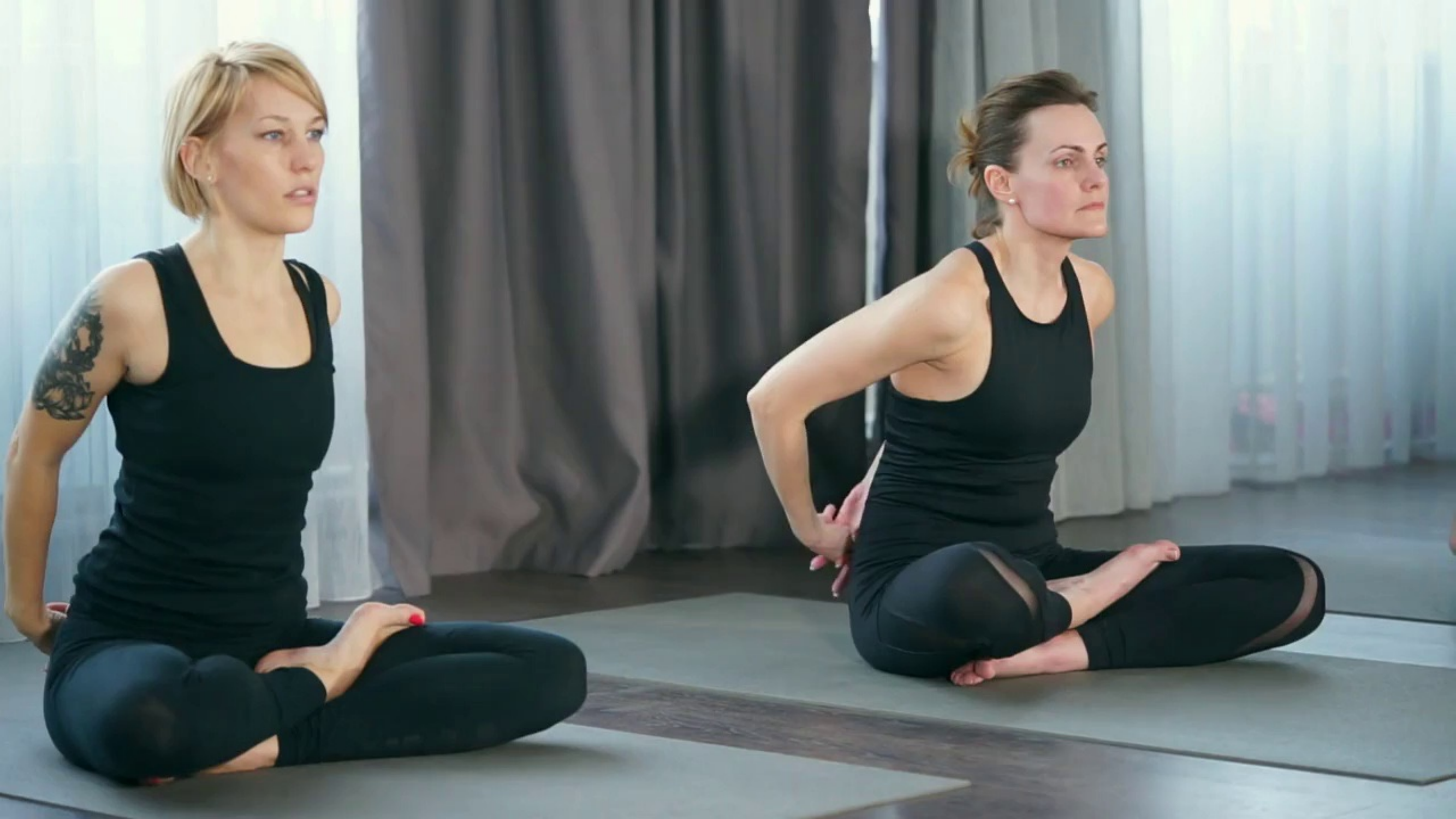} &
        Remove the woman on the left. Increase background brightness to lighten the scene, keeping the woman on the right and her pose unchanged. Maintain original lighting on her, and preserve the dark curtains and floor mat textures. &
        \includegraphics[width=0.11\textwidth]{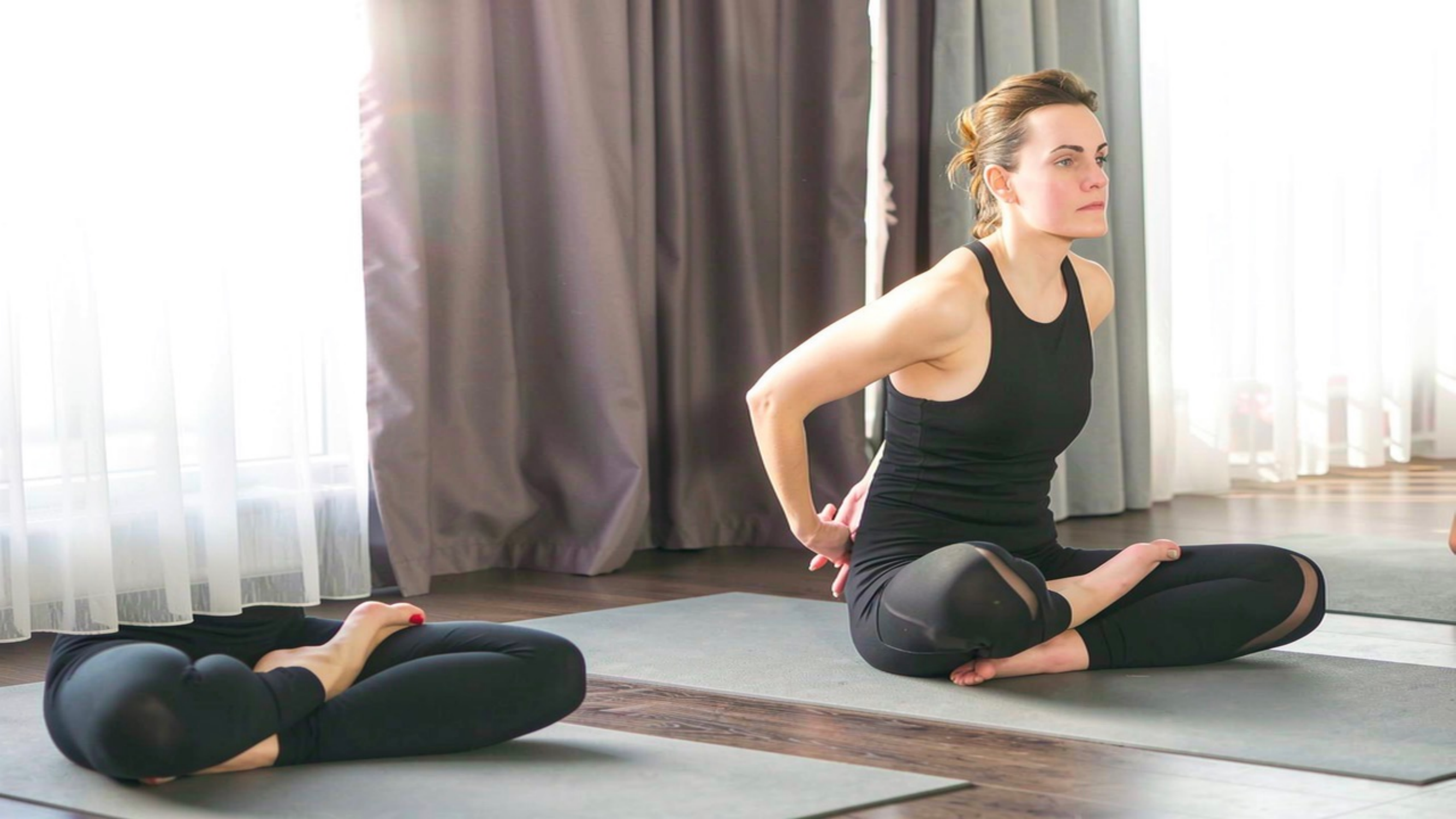} &
        \textbf{Remove the woman on the left—short blonde hair, black tank top, tattoo on left arm, red nail polish—sitting cross-legged on a yoga mat, from the left half of the frame}. Inpaint seamlessly using background texture of curtain and floor, blending edges smoothly. Brighten the entire background, especially the white sheer curtains and floor, to enhance contrast and create a luminous, airy atmosphere while preserving the right-side subject and studio lighting. &
        \includegraphics[width=0.11\textwidth]{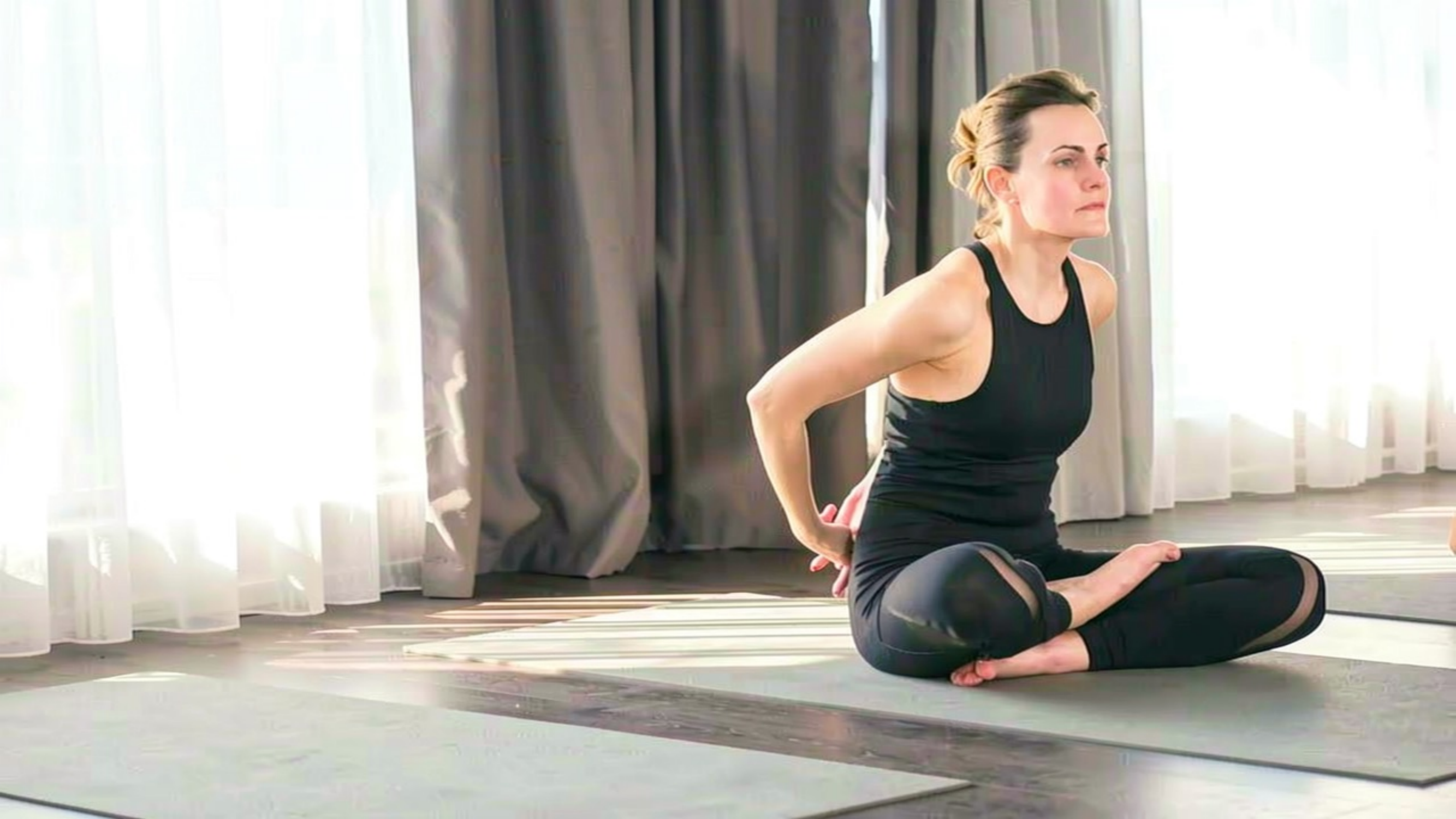} \\
        \midrule

        Extract the light blue off-shoulder dress worn by the woman in the sunflower field. &
        \includegraphics[width=0.11\textwidth]{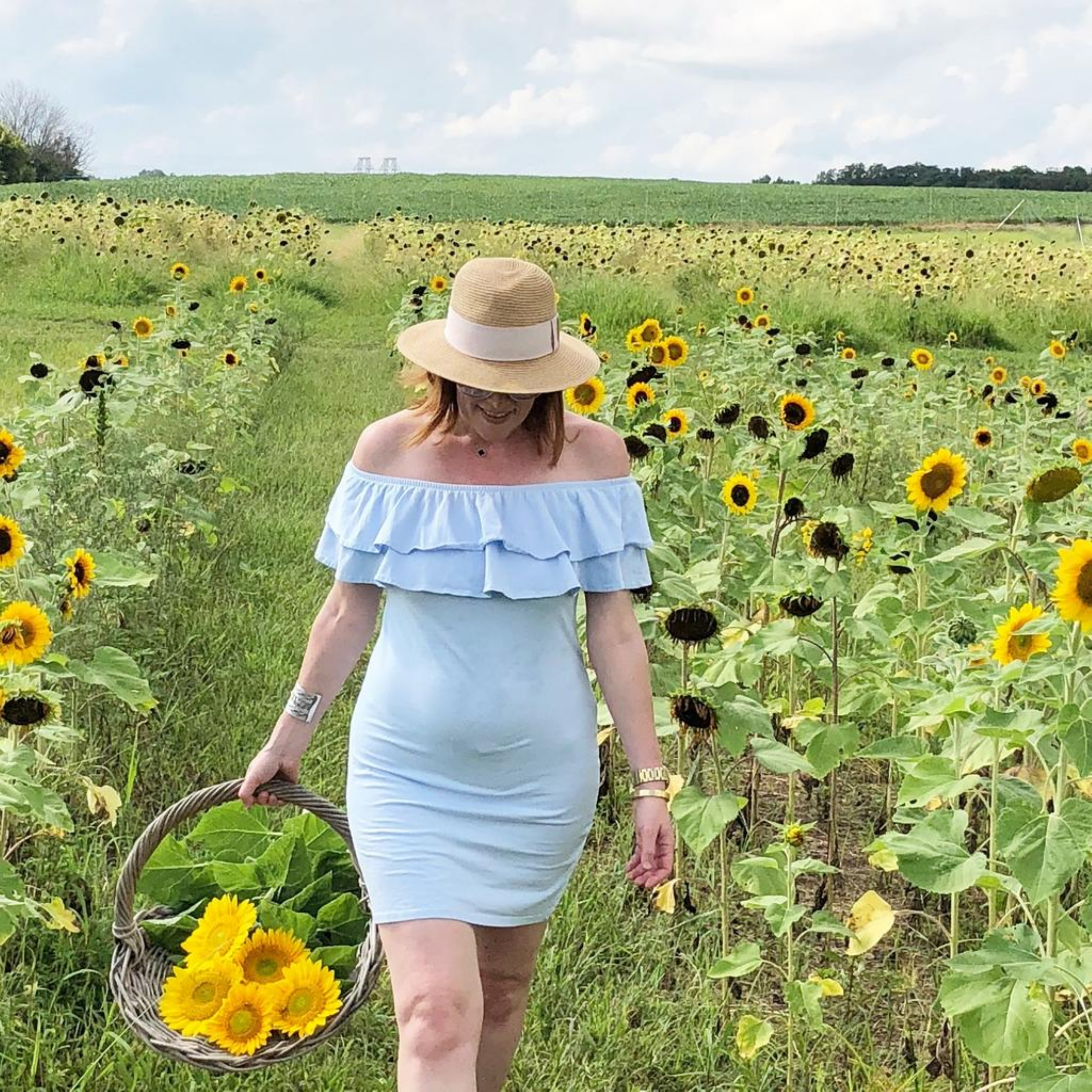} &
        Extract the light blue off-shoulder dress worn by the woman, keeping her straw hat, smiling face, and basket of sunflowers intact. Maintain the natural sunlight, field background, and her posture as she walks through the sunflowers. &
        \includegraphics[width=0.11\textwidth]{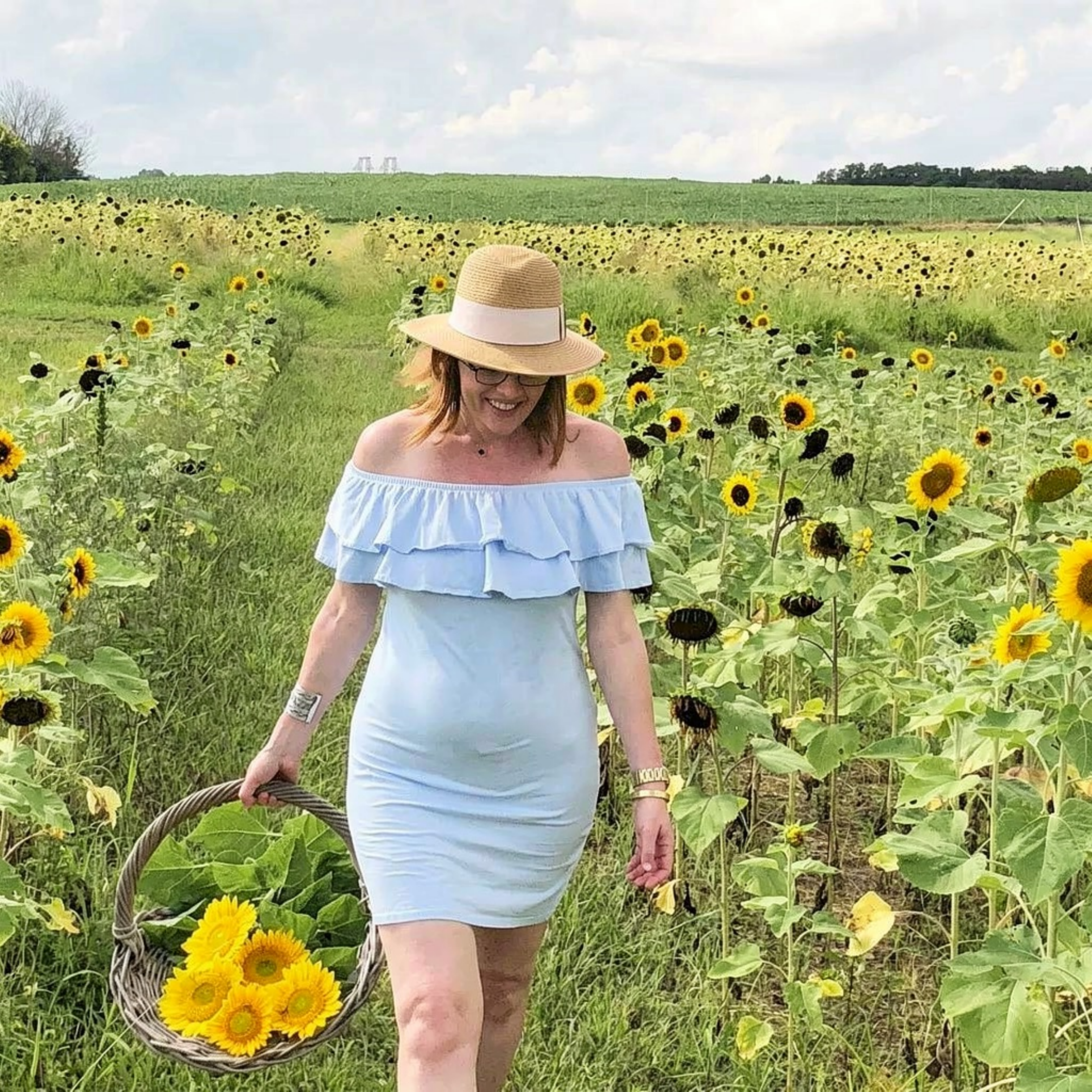} &
        Extract the light blue off-shoulder dress from the woman in the sunflower field, preserving ruffle texture, fabric drape, and natural seams with precise edge detection and soft transitions. \textbf{Isolate against a clean white background, excluding arms, hat, and basket}. Retain all folds, creases, and transparent edges; remove background sunflowers, sky, and greenery. Focus on dress from shoulders to knees. &
        \includegraphics[width=0.11\textwidth]{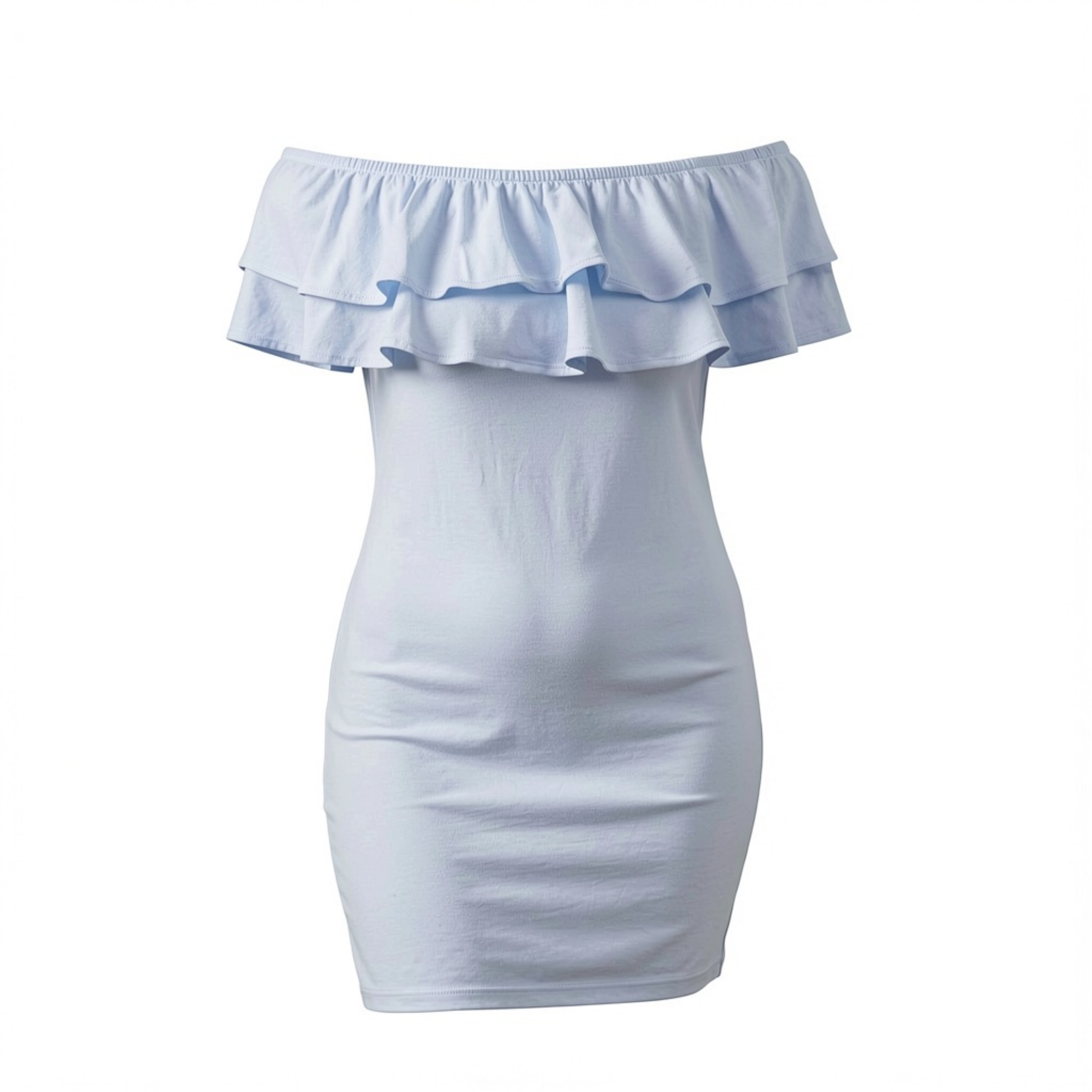} \\
        \midrule

        Extract the white building with a tower structure located on the left side of the image, situated on a beach during sunset. &
        \includegraphics[width=0.11\textwidth]{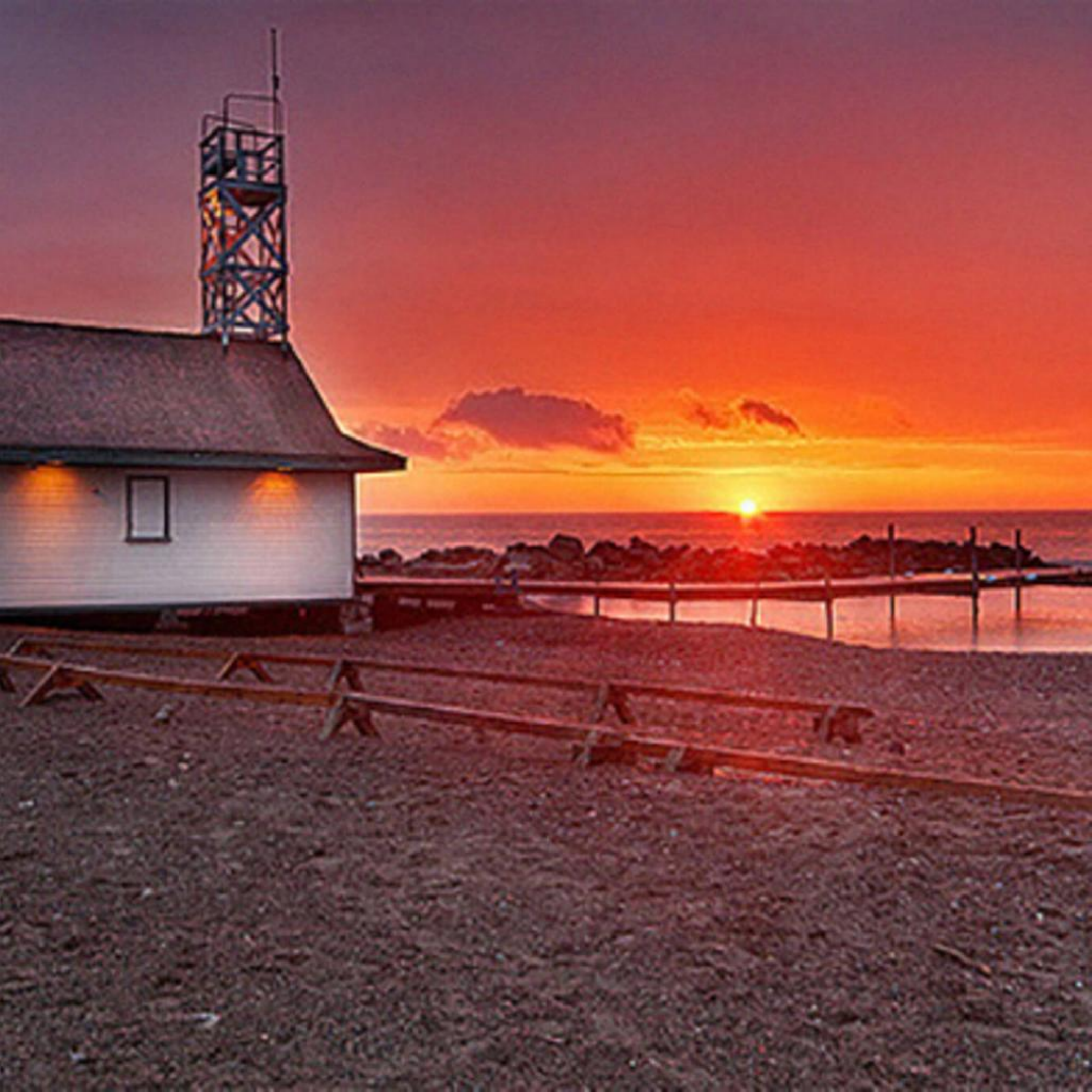} &
        Extract the white building with a tower structure on the left, keeping its roof, tower, and window intact; maintain the sunset lighting and beach foreground, removing all surrounding elements including the wooden benches and distant pier. &
        \includegraphics[width=0.11\textwidth]{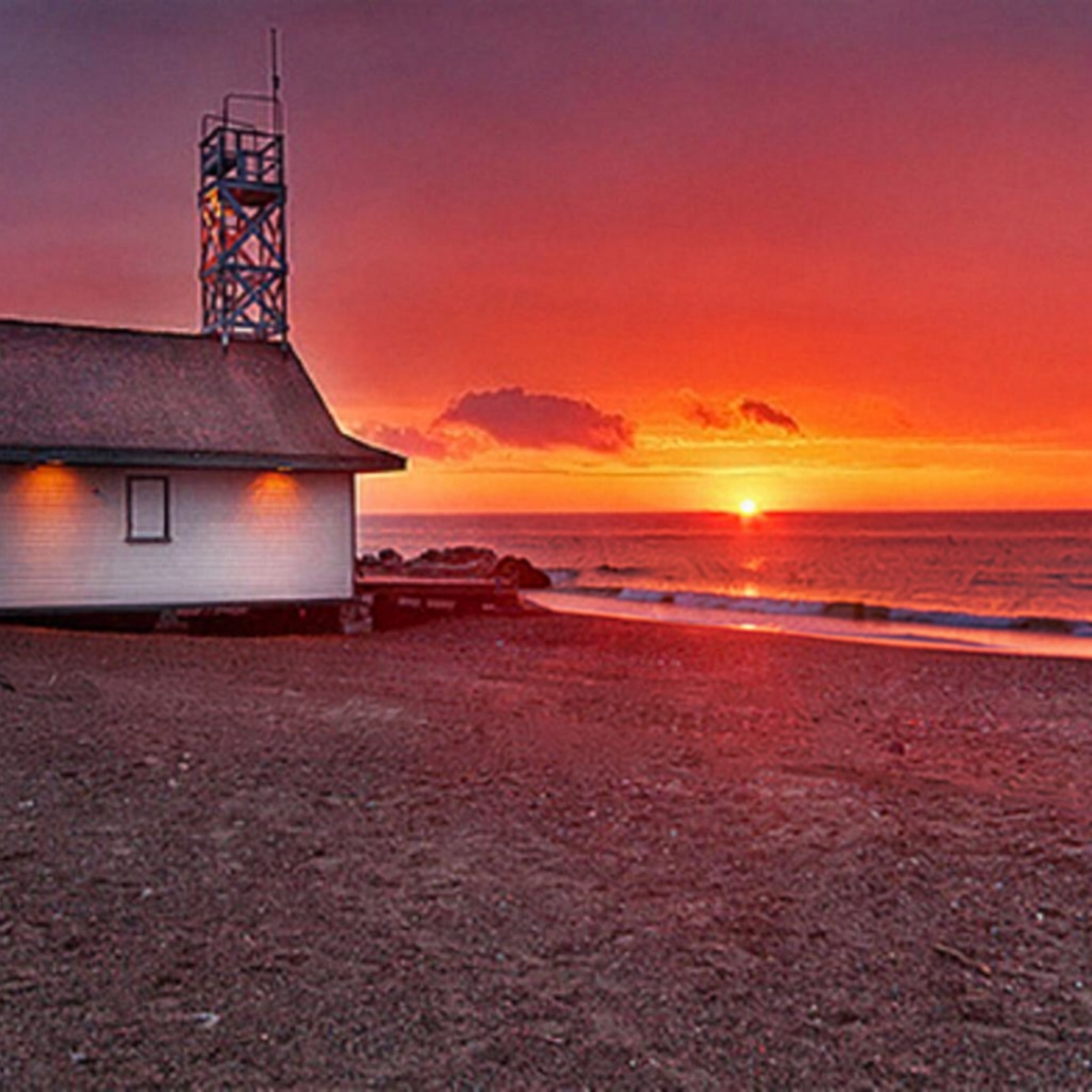} &
        Extract the white building with its tower structure — including sloped roof, single window, metallic lattice tower with railings and antennas, and warm interior lighting — from the left side of the image, preserving all fine details like wall texture and metal framework. \textbf{Remove the beach, sunset, rocky breakwater, benches, and sky. Isolate the building against a clean white background, retaining its full structure and illumination}. &
        \includegraphics[width=0.11\textwidth]{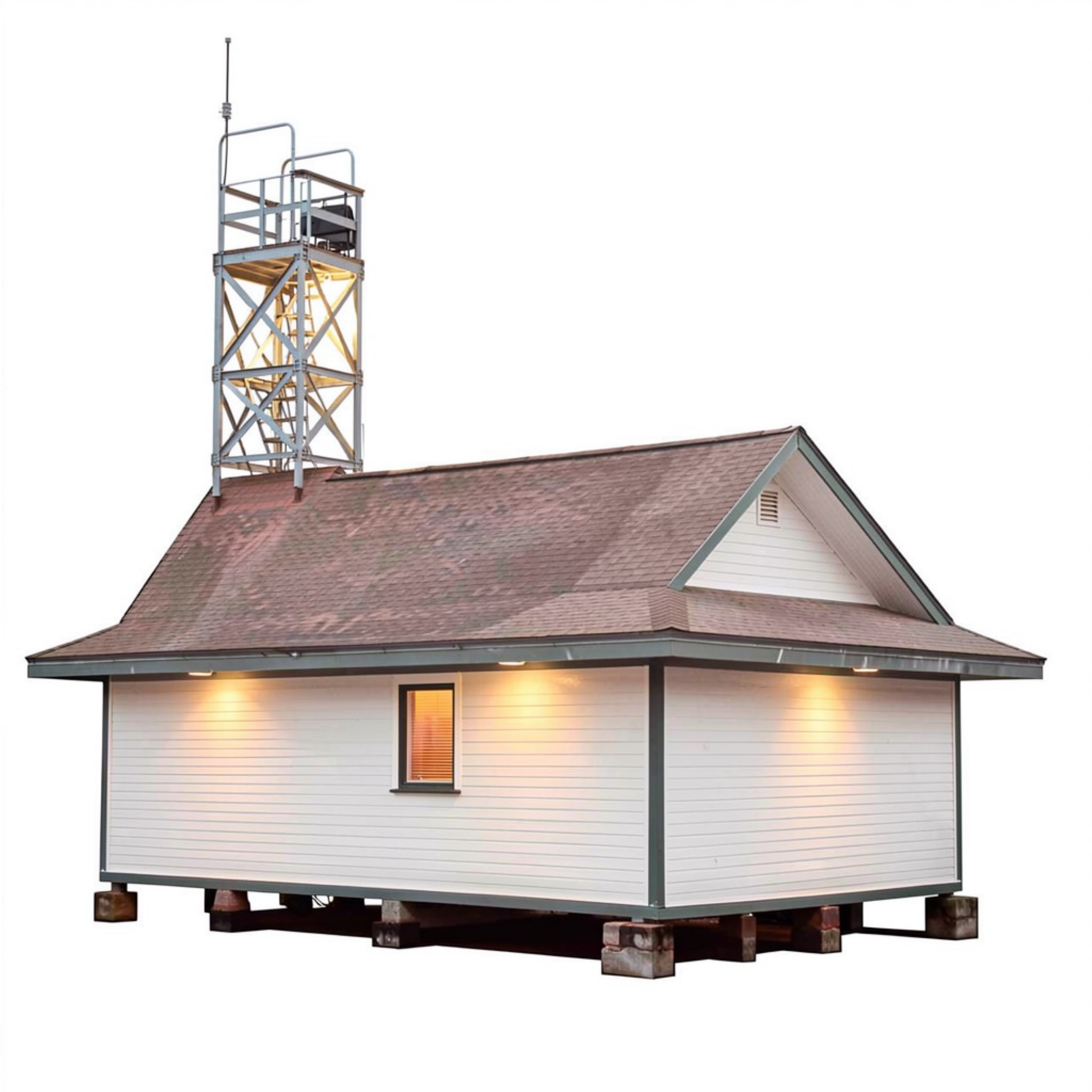} \\

        \bottomrule
    \end{tabular}
\end{table}

\clearpage

\begin{table}[H]
    \ContinuedFloat
    \centering
    \scriptsize
    \setlength{\tabcolsep}{3pt}
    \renewcommand{\arraystretch}{1.05}
    \caption[]{(Continued.) Part III.}
    \begin{tabular}{L{0.12\textwidth} C{0.12\textwidth} L{0.16\textwidth} C{0.12\textwidth} L{0.24\textwidth} C{0.12\textwidth}}
        \toprule
        \multirow{2}{*}{\textbf{Instruction}} &
        \multirow{2}{*}{\makecell[c]{\textbf{Original}\\\textbf{Image}}} &
        \multicolumn{2}{c}{\textbf{Qwen3-VL-4B}} &
        \multicolumn{2}{c}{\textbf{MAPE (Qwen3-VL-4B)}} \\
        \cmidrule(lr){3-4} \cmidrule(lr){5-6}
        & & \multicolumn{1}{c}{\textbf{Prompt}} & \textbf{Image} & \multicolumn{1}{c}{\textbf{Prompt}} & \textbf{Image} \\
        \midrule

        Extract the colorful domes and structure of the St. Basil's Cathedral architecture in the image &
        \includegraphics[width=0.11\textwidth]{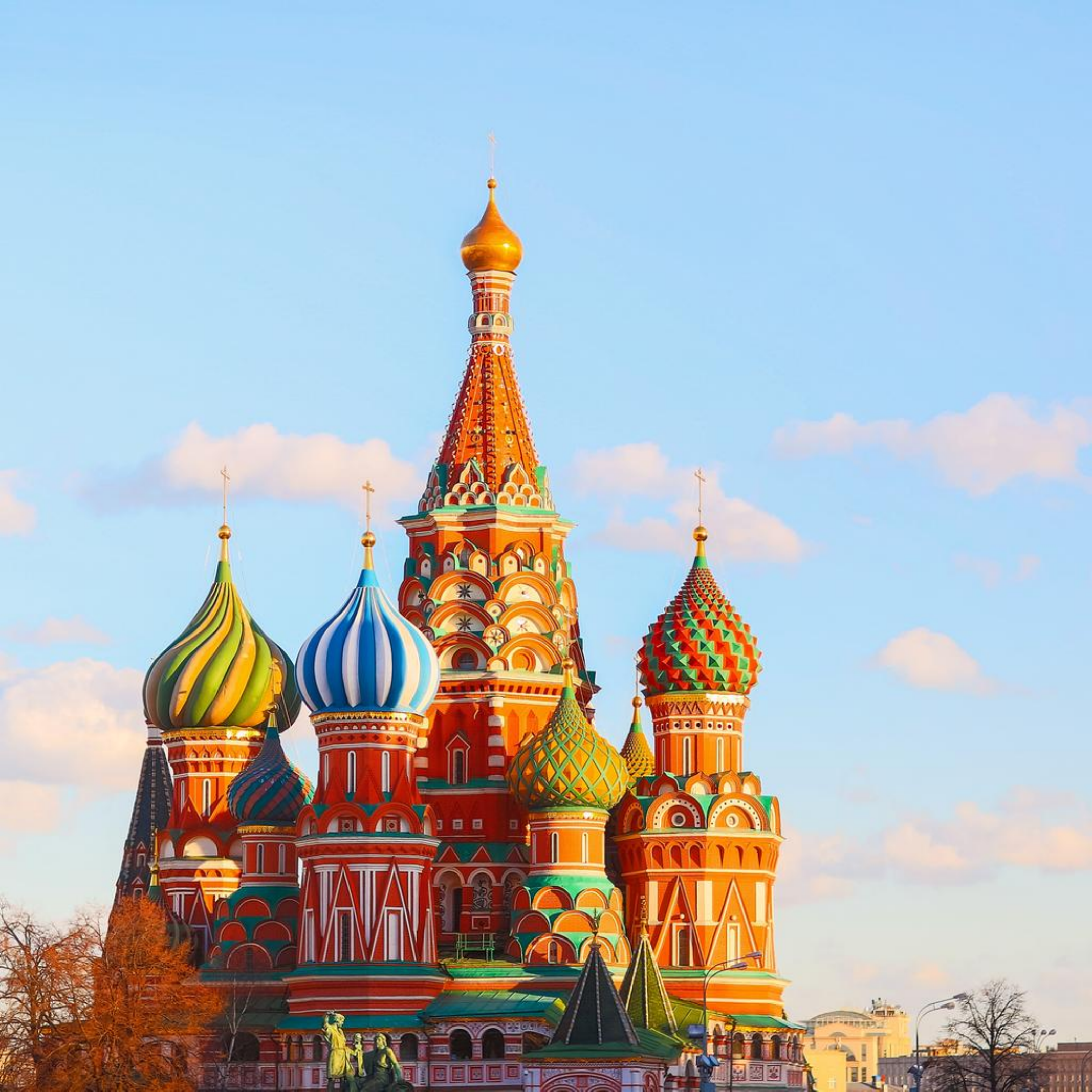} &
        Isolate the colorful domes and intricate architectural structure of St. Basil's Cathedral, keeping the vibrant patterns, golden spires, and ornate red-and-white facades intact. Maintain the original composition, lighting, and sky backdrop; remove all surrounding trees and buildings to focus solely on the cathedral's iconic design. &
        \includegraphics[width=0.11\textwidth]{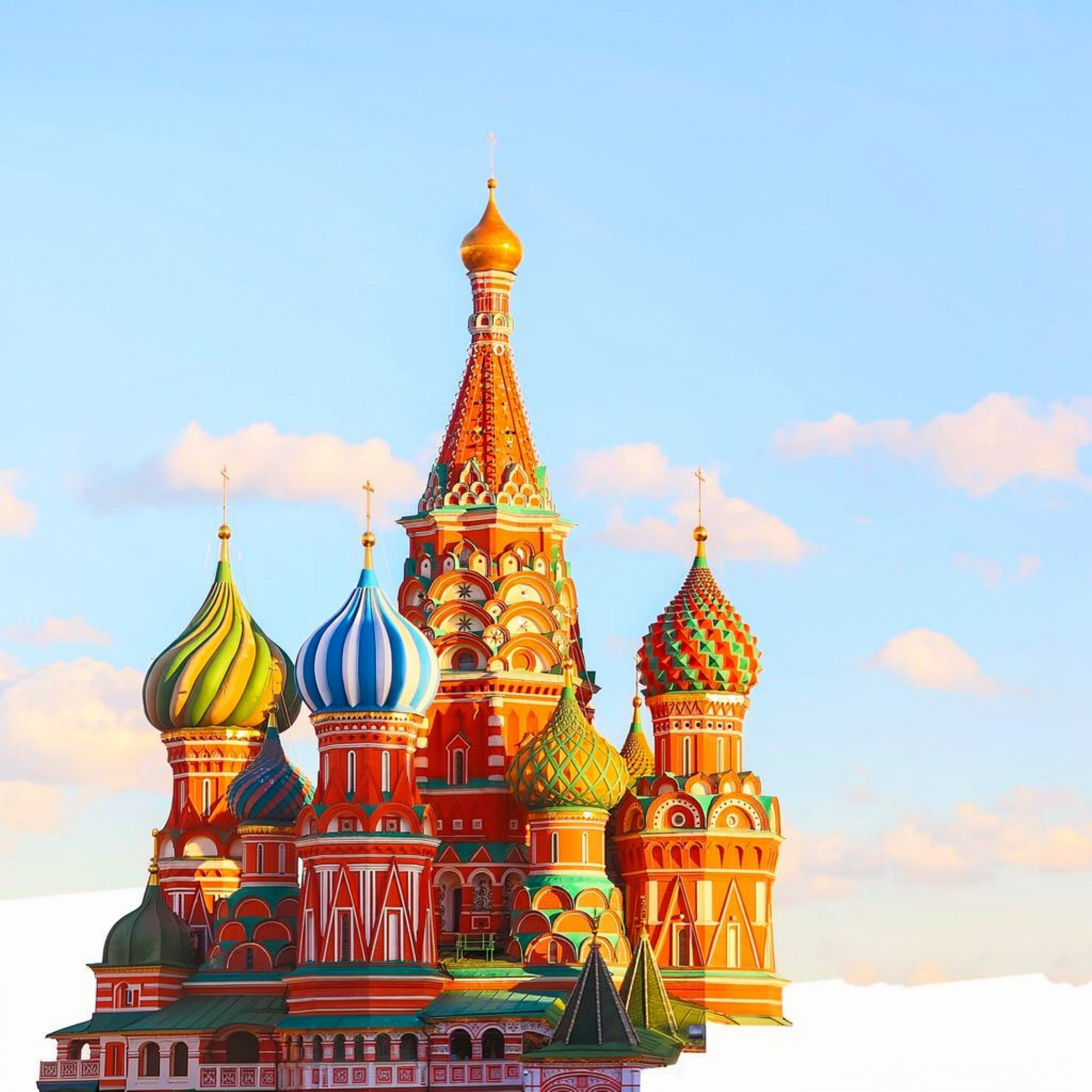} &
        Extract the entire St. Basil's Cathedral structure — \textbf{including all colorful onion-shaped domes (green, blue, gold, red), ornate brickwork, arched windows, and spires with golden finials — with highly detailed edge detection preserving textures and transparent edges}. Isolate against a clean white background, removing sky, trees, and distant buildings. Retain all intricate patterns, color gradients, and structural integrity. &
        \includegraphics[width=0.11\textwidth]{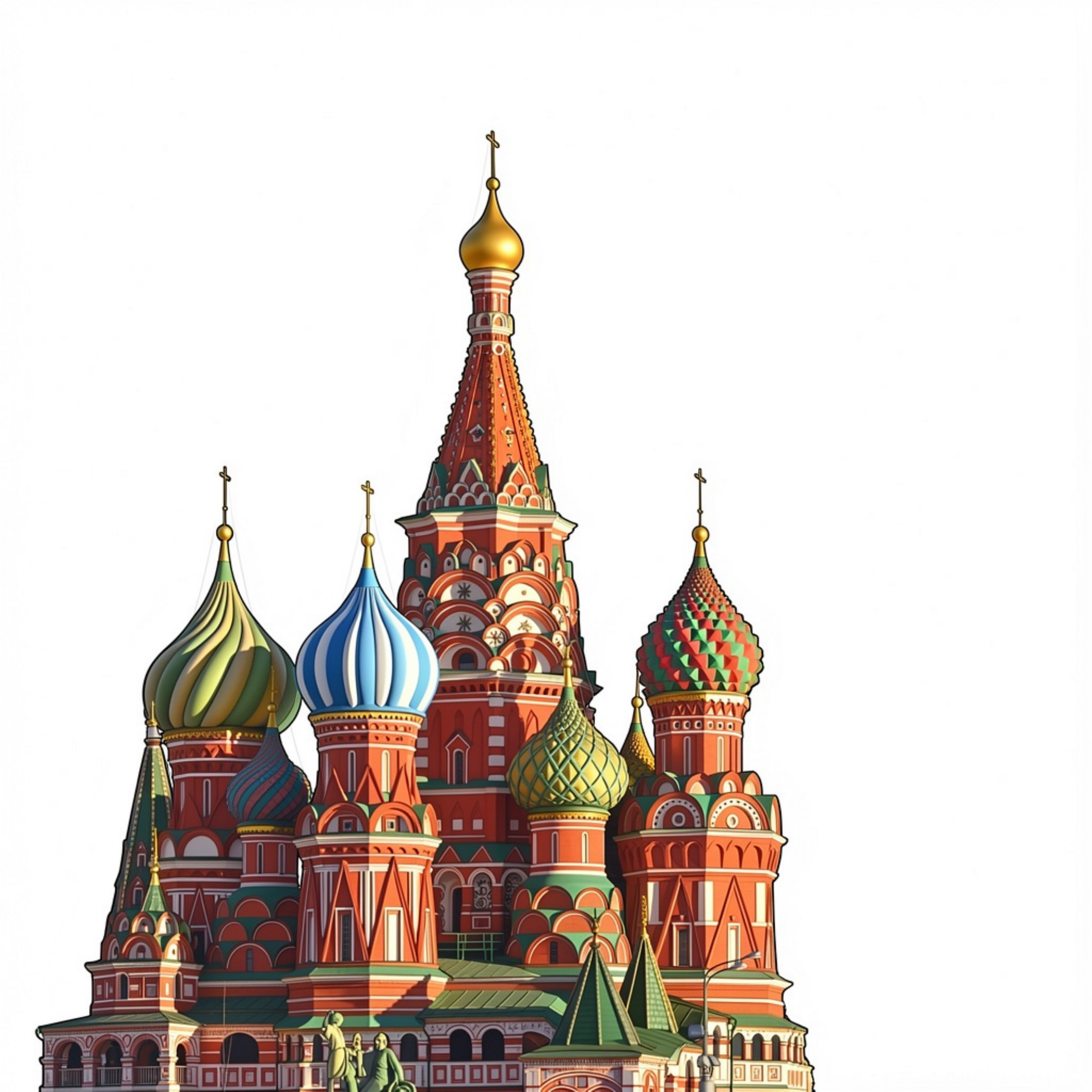} \\
        \midrule

        Extract the motorcycle from the image. &
        \includegraphics[width=0.11\textwidth]{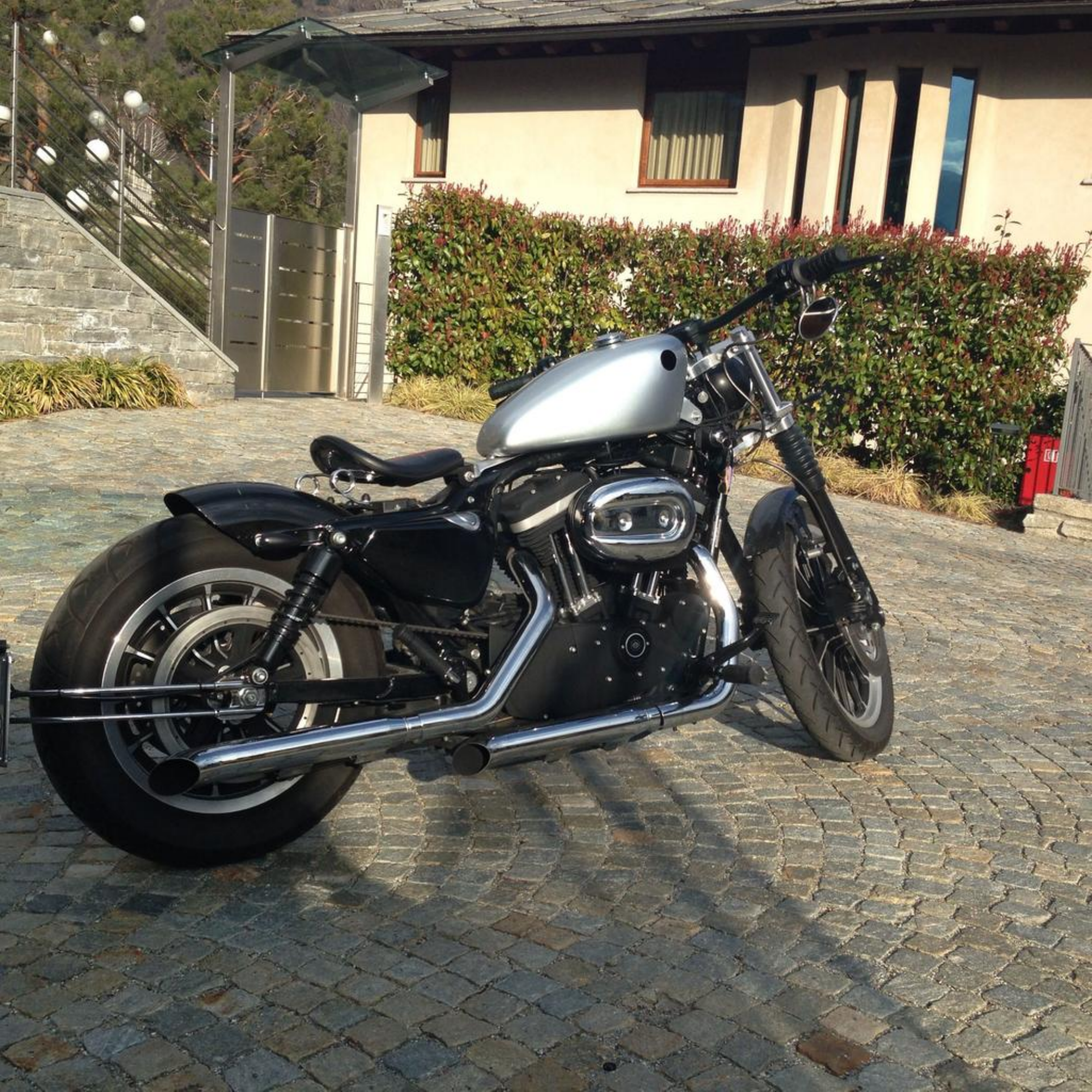} &
        Remove the motorcycle from the image, keeping the cobblestone driveway, house, hedge, and staircase intact. Do not alter the background elements or lighting conditions. &
        \includegraphics[width=0.11\textwidth]{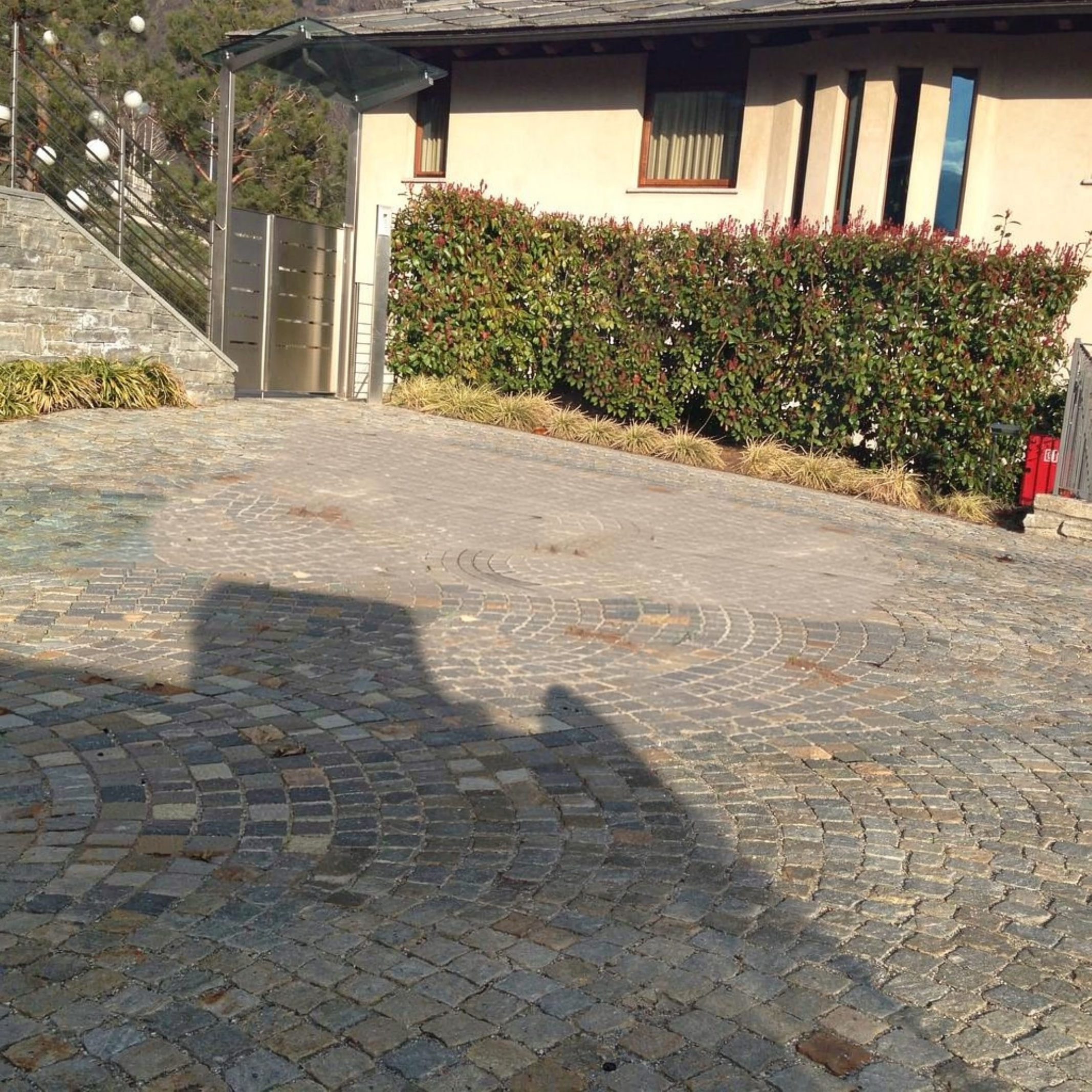} &
        Extract the entire motorcycle — \textbf{including wheels, engine, handlebars, seat, exhaust, and chrome surfaces} — from the image. Preserve all reflective chrome, tire tread, seat texture, and subtle shadows. Isolate against a clean white background, removing cobblestone, house, hedges, and gate. Maintain original details, lighting, and diagonal center-right positioning. &
        \includegraphics[width=0.11\textwidth]{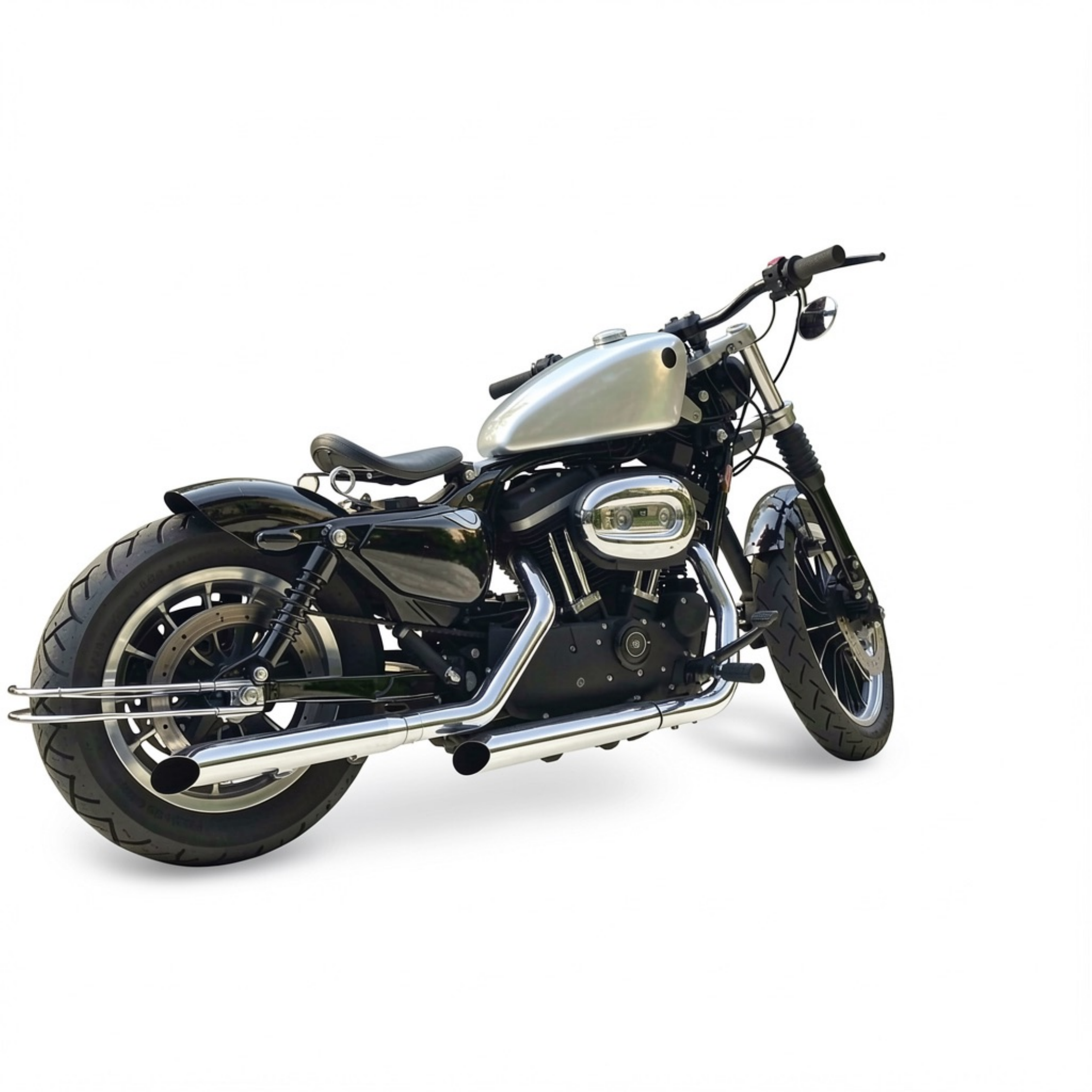} \\
        \bottomrule
    \end{tabular}
\end{table}

\clearpage
\section{Prompt Length Distribution Change}
\label{appendix:distribution}
\begin{figure}[t]
    \centering
    \includegraphics[width=\linewidth]{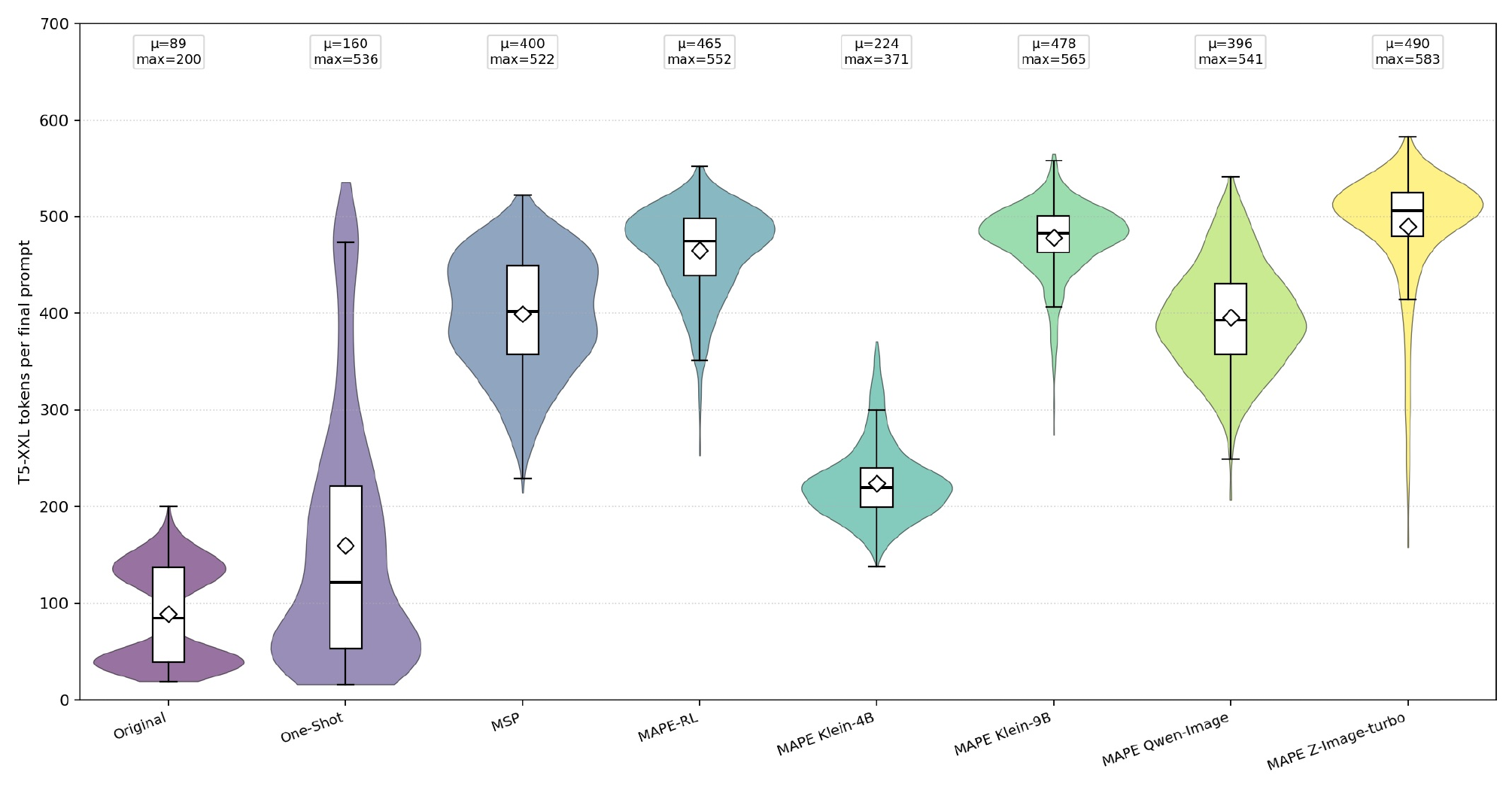}
    \caption{Final prompt length distribution change for different prompt enhancer frameworks and different training stages.}
    \label{fig:prompt_length}
\end{figure}
From \Cref{fig:prompt_length}, we can find that the prompt length increases from the original prompt, to one-shot SLMs, MSP and until MAPE - RL. This means that more test-time computation, the agentic pipeline and SFT can help adding more details to guide the downstream image generation models. However, after RL, the prompt length is different for each image generation model, which means that RL can help the agentic pipeline adapt to each image generation model, and some models prefer shorter prompts (e.g., FLUX.2-klein-4B) and some prefer longer prompts (Z-Image-turbo). This might be the result of difference between image generation models' capability and the dependence on the details during the pretraining stages. 

\section{Ablation Studies}
\label{appendix:ablation}
In \Cref{tab:unigenbench_appendix}, we show the full ablation studies for our MAPE framework, where we can find that MAPE can significantly improve the image generation model performance and achieve comparable performance to prompt enhancers with commercial LLMs. We can also find the architecture and each training stage of MAPE can bring their own benefits, including task decomposition, knowledge injection and downstream image generation model adaptation. 

\begin{table}
    \centering
    \scriptsize
    \caption{UniGenBench performance ablation study.}
    \label{tab:unigenbench_appendix}
        \begin{tabular}{c|c|c|c}
        \hline
        T2I Model & Prompt Enhancer & UniGen Short & UniGen Long \\
        \hline
        \multirow{11}{*}{Qwen-Image-2512} & \ding{55} & 0.7493 & 0.8869 \\
        & Gemini-3.1-Pro & 0.8271 & 0.8400 \\
        & Gemini-3.1-Pro (MSP) & 0.8669 & 0.8758 \\ \cline{2-4}
        & Qwen3-1.7B & 0.7645 & 0.8719 \\
        & Qwen3-1.7B (MSP) & 0.7617 & 0.8551 \\
        & MAPE - RL (Qwen3-1.7B) & 0.8268 & \textbf{0.8714} \\
        & MAPE (Qwen3-1.7B) & \textbf{0.8334} & 0.8624 \\\cline{2-4}
        & Qwen3-4B & 0.7055 & 0.7233 \\
        & Qwen3-4B (MSP) & 0.8119 & 0.8801 \\
        & MAPE - RL (Qwen3-4B) & 0.8473 & 0.8908 \\
        & MAPE (Qwen3-4B) & \textbf{0.8539} & \textbf{0.8923} \\
        \hline
        \multirow{11}{*}{Z-Image-turbo} & \ding{55} & 0.6931 & 0.8170 \\
        & Gemini-3.1-Pro & 0.6949 & 0.8170 \\
        & Gemini-3.1-Pro (MSP) & 0.7501 & 0.7674 \\ \cline{2-4}
        & Qwen3-1.7B & 0.6878 & 0.7947 \\
        & Qwen3-1.7B (MSP) & 0.7383 & 0.8231 \\
        & MAPE - RL (Qwen3-1.7B) & 0.7623 & 0.8321 \\
        & MAPE (Qwen3-1.7B) & \textbf{0.7716} & \textbf{0.8405} \\\cline{2-4}
        & Qwen3-4B & 0.7221 & 0.8081 \\
        & Qwen3-4B (MSP) & 0.7721 & 0.8290 \\
        & MAPE - RL (Qwen3-4B) & 0.8233 & 0.8502 \\
        & MAPE (Qwen3-4B) & \textbf{0.8356} & \textbf{0.8512} \\
        \hline
        \multirow{11}{*}{FLUX.2-klein-4B} & \ding{55} & 0.7489 & 0.8464 \\
        & Gemini-3.1-Pro & 0.8169 & 0.8630 \\
        & Gemini-3.1-Pro (MSP) & 0.8263 & 0.8392 \\ \cline{2-4}
        & Qwen3-1.7B & 0.6790 & 0.8170 \\
        & Qwen3-1.7B (MSP) & 0.6994 & 0.8028 \\
        & MAPE - RL (Qwen3-1.7B) & 0.7557 & 0.8183 \\
        & MAPE (Qwen3-1.7B) & \textbf{0.7710} & \textbf{0.8275} \\\cline{2-4}
        & Qwen3-4B & 0.4817 & 0.5037 \\
        & Qwen3-4B (MSP) & 0.7485 & 0.8365 \\
        & MAPE - RL (Qwen3-4B) & 0.7973 & 0.8477 \\
        & MAPE (Qwen3-4B) & \textbf{0.8042} & \textbf{0.8539} \\
        \hline
        \multirow{11}{*}{FLUX.2-klein-9B} & \ding{55} & 0.8058 & 0.8792 \\
        & Gemini-3.1-Pro & 0.8731 & 0.8908 \\
        & Gemini-3.1-Pro (MSP) & 0.8553 & 0.8704 \\ \cline{2-4}
        & Qwen3-1.7B & 0.7093 & 0.8518 \\
        & Qwen3-1.7B (MSP) & 0.7398 & 0.8382 \\
        & MAPE - RL (Qwen3-1.7B) & 0.8075 & 0.8561 \\
        & MAPE (Qwen3-1.7B) & \textbf{0.8268} & \textbf{0.8578} \\\cline{2-4}
        & Qwen3-4B & 0.4777 & 0.5154 \\
        & Qwen3-4B (MSP) & 0.7919 & 0.8683 \\
        & MAPE - RL (Qwen3-4B) & 0.8345 & \textbf{0.8737} \\
        & MAPE (Qwen3-4B) & \textbf{0.8460} & 0.8641 \\
        \hline
    \end{tabular}
    \vspace{-10pt}
\end{table}

\section{Details about GRPO and GDPO}
\label{appendix:rl_detail}
\subsection{GRPO for Scalar-Reward Optimization}
\label{appendix:grpo}
When supervision is available as a scalar reward, we optimize the prompt enhancer using Group Relative Policy Optimization (GRPO). GRPO is well suited to prompt enhancement because there is no single correct rewrite; instead, the quality of a rewritten prompt is determined by the image it produces. 

For an input state $s$, we sample a group of $G$ candidate prompts $p_1,...,p_G\sim\pi_{\theta_\text{old}}(\cdot|s)$, where $\pi_{\theta_\text{old}}$ is the behavior policy. Each prompt is executed by the downstream model to produce an output $y_i=G(s,p_i)$, which receives a scalar reward $r_i=R(y_i,s)$. If several scalar metrics are available, they may be combined into a single reward $r_i=\sum_{k=1}^K \lambda_kr_i^{(k)}$, where $\lambda_k\geq0$ are fixed coefficients. GRPO computes a group-relative normalized advantage using the group mean and standard deviation,
\begin{align}
    \mu=\frac{1}{G}\sum_{i=1}^G r_i,\quad \sigma=\sqrt{\frac{1}{G}\sum_{i=1}^G (r_i-\mu)^2+\epsilon},\quad A_i=\frac{r_i-\mu}{\sigma}
\end{align}
Let $\rho_i(\theta)=\frac{\pi_\theta(p_i|s)}{\pi_{\theta_\text{old}}(p_i|s)}$ be the importance ratio. The clipped GRPO objective for one input $s$ is
\begin{align}
    \mathcal{J}_{\text{GRPO}}(\theta;s)=\frac{1}{G}\sum_{i=1}^G \min(\rho_i(\theta)A_i,\text{clip}(\rho_i(\theta),1-\epsilon,1+\epsilon)A_i)-\beta \text{KL}(\pi_\theta(\cdot|s)||\pi_{\text{ref}}(\cdot|s))
\end{align}
This objective is agnostic to whether the underlying architecture is SAPE or MAPE. In the latter case, users can choose which modules in the router–rewriter–composer pipeline to be optimized through the likelihood of the final prompt $p$.

\subsection{GDPO for Multi-Reward Optimization}
\label{appendix:gdpo}
When supervision is inherently multi-reward, especially when using a VLM to score testpoints, we replace GRPO with Group reward-Decoupled Normalization Policy Optimization (GDPO). GDPO was introduced to address a specific failure mode of applying GRPO naively in multi-reward settings: if one first aggregates multiple rewards and then performs group normalization, distinct reward combinations can collapse to identical normalized advantages, reducing the resolution of the training signal and harming optimization. GDPO resolves this issue by decoupling group-wise normalization across reward dimensions before aggregation, and then applying an additional batch-wise normalization step to stabilize the final advantage scale.

For each sampled prompt  $p_i$, let the downstream evaluator return a reward vector $\mathbf{r}=(r^{(1)},r^{(2)},...,r^{(K)})\in\mathbb{R}^K$. For each reward dimension $k$, GDPO first computes group-wise statistics 
\begin{align}
    \mu^{(k)}=\frac{1}{G}\sum_{i=1}^G r_i^{(k)},\quad \sigma^{(k)}=\sqrt{\frac{1}{G}\sum_{i=1}^G (r_i^{(k)}-\mu^{(k)})^2+\epsilon},\quad A_i^{(k)}=\frac{r_i^{(k)}-\mu^{(k)}}{\sigma^{(k)}}. 
\end{align}
These per-reward normalized advantages are then aggregated: $\tilde{A_i}=\sum_{k=1}^K w_kA_i^{(k)}$, where $w_k\geq0$ are optional reward weights. Following GDPO, the aggregated advantage is then normalized again across the training batch, $\hat{A_i}=\frac{\tilde{A_i}-\mu_{\text{batch}}}{\sigma_{\text{batch}}}$, where $\mu_\text{batch}$ and $\sigma_{\text{batch}}$ are the batch mean and standard deviation of $\tilde{A_i}$. This second normalization step keeps the optimization scale stable as the number or magnitude of reward dimensions changes. Using the same importance ratio $\rho_i(\theta)=\frac{\pi_\theta(p_i|s)}{\pi_{\theta_\text{old}}(p_i|s)}$, the GDPO objective function is 
\begin{align}
    \mathcal{J}_\text{GDPO}(\theta;s)=\frac{1}{G}\sum_{i=1}^G \min(\rho_i(\theta)\hat{A_i},\text{clip}(\rho_i(\theta),1-\epsilon,1+\epsilon)\hat{A_i})-\beta \text{KL}(\pi_\theta(\cdot|s)||\pi_{\text{ref}}(\cdot|s))
\end{align}
Thus, GRPO and GDPO share the same clipped policy optimization backbone, but differ in how they construct the advantage signal: GRPO uses a single scalar reward and group-relative normalization, whereas GDPO decouples normalization across reward dimensions before aggregation and batch normalization.

Our framework cleanly disentangles prompt-enhancement architecture from reward-based policy optimization. SAPE and MAPE define two architectural instantiations of the prompt enhancer, while GRPO and GDPO define two optimization schemes corresponding to scalar and multi-reward supervision, respectively. In particular, MAPE realizes prompt enhancement as a router–rewriter–composer pipeline: the router chooses which semantic fields to refine, field-specific rewriters elaborate only the selected fields, and the composer constructs the final natural-language prompt. This modular design allows the framework to scale its reasoning structure with task difficulty while remaining trainable under a unified reinforcement learning formulation.

\section{Prompts}
\label{appendix:prompts}
\subsection{Prompts for MAPE-Gen}
\label{appendix:prompts_gen}
\begin{lstlisting}[caption={Prompt for MAPE-Gen router}, label={lst:prompt_MAPE-Gen_router}]
_ROUTER_SYSTEM_PROMPT = f"""You are a visual scene analysis expert. Your task is to analyze a user's image generation prompt and determine which visual domains need creative elaboration to produce a high-quality image.

The available domains are:
{chr(10).join(f'- `{k}`: {_FIBO_DOMAIN_SCHEMAS[k].split(chr(10))[0]}' for k in _ALL_FIBO_DOMAIN_KEYS)}

Rules:
1. `short_description` and `objects` are ALWAYS selected - they are mandatory.
2. `text_render` should ONLY be selected if the prompt explicitly requests text to appear in the image (e.g., a poster with a title, a sign with words). If no text is mentioned, do NOT select it.
3. Select domains that are underspecified or ambiguous in the original prompt and would benefit from creative elaboration by a specialist.
4. If the prompt already fully specifies a domain (e.g., "oil painting in impressionist style" fully specifies style_medium and artistic_style), you may still select it if additional refinement would help.
5. For most prompts, select 5-8 domains. Only skip domains that are truly irrelevant.

Output ONLY a valid JSON object with these keys:
- "selected_domains": a list of domain key strings from the available domains
- "reasoning": an object mapping each selected domain key to a brief (1-sentence) justification

Do not include any text before or after the JSON object."""

_ROUTER_USER_TEMPLATE = (
    "Analyze the following image generation prompt and determine which visual "
    "domains need specification:\n\n{user_prompt}"
)
\end{lstlisting}

\begin{lstlisting}[caption={Prompt for MAPE-Gen domain rewriter}, label={lst:prompt_MAPE-Gen_rewriter}]
_DOMAIN_REWRITER_SYSTEM_PROMPT_TEMPLATE = """You are a specialist Visual Art Director focused on the "{domain_name}" aspect of image scenes.

Your task: Given a user's image generation prompt, generate a detailed JSON specification for the "{domain_name}" domain ONLY.

Schema for this domain:
{domain_schema}

Guidelines:
1. Be creative yet faithful to the original prompt's intent.
2. Infer and add details not explicitly mentioned to produce a high-quality, visually compelling result.
3. Output ONLY a valid JSON object with the key "{domain_key}" containing the specification.
4. Do not include any text before or after the JSON object.
5. When describing human body parts, positions, or actions, describe from the PERSON'S OWN PERSPECTIVE.
6. For human subjects, prefer compositions that frame them prominently (medium shot, close-up) unless the prompt implies a wide scene.
7. Default to photorealistic style unless the prompt requests otherwise.

Router reasoning for selecting this domain: {router_reasoning}"""

_DOMAIN_REWRITER_USER_TEMPLATE = (
    "Generate a detailed JSON specification for the \"{domain_name}\" domain "
    "based on this image generation request:\n\n{user_prompt}"
)
\end{lstlisting}

\begin{lstlisting}[caption={Prompt for MAPE-Gen composer}, label={lst:prompt_MAPE-Gen_composer}]
_COMPOSER_SYSTEM_PROMPT = """\
You are an expert prompt engineer for text-to-image generation models. \
Your task is to take a collection of domain-specific JSON specifications \
for an image scene and compose them into a single, detailed, flowing \
natural language prompt.

Rules:
1. Capture ALL visual details from every domain specification: objects, \
their attributes, positions, relationships, background, lighting, \
composition, colors, mood, camera angle, depth of field, and artistic style.
2. Output ONLY the natural language prompt. No explanations, no preamble, no formatting marks.
3. The output should be a single continuous paragraph, rich in visual detail, \
suitable for directly feeding into a text-to-image model.
4. Preserve any text rendering requirements (text content, position, style).
5. Be specific and concrete. Use vivid, descriptive language.
6. Do NOT include quality tags like "8K", "masterpiece", "best quality".
7. Ensure consistency across all domain specifications - resolve any contradictions \
by favoring the more specific or detailed description."""

_COMPOSER_USER_TEMPLATE = (
    "Compose the following domain-specific JSON specifications into a single "
    "detailed natural language prompt for a text-to-image model.\n\n"
    "Original prompt: {original_prompt}\n\n"
    "Domain specifications:\n{domain_json_str}"
)
\end{lstlisting}

\begin{lstlisting}[caption={Prompt for UniGenBench scoring}, label={lst:prompt_MAPE-Gen_scoring}]
TESTPOINT_PROMPT_TEMPLATE = (
    'You are a precise and objective English-language image description system. '
    'I will provide you with a prompt for image generation, as well as the '
    'corresponding generated image. You will be given a set of evaluation criteria '
    '(checkpoints) and their explanations. You must evaluate whether the generated '
    'image fulfills the requirements implied by each checkpoint in the prompt.\n\n'
    'For each image, follow the steps below in order:\n\n'
    '1. The prompt for the generated image is: \u300c{prompt}\u300d. '
    'You are to analyze the image content in detail from the angles specified in '
    '{testpoint}. Detailed definitions of these checkpoints are provided here: '
    '{explanation}. The specific description of each checkpoint in the context of '
    'the prompt is: {test_explanation}. You must analyze whether the image meets '
    'the requirements for each checkpoint individually.\n\n'
    '2. Based on the above analysis, determine whether the generated image satisfies '
    'each checkpoint. If the image meets the requirements of a checkpoint, assign a '
    'score of 1; otherwise, assign a score of 0.\n\n'
    'Constraints:\n'
    '- Only describe content that is directly visible.\n'
    '- Focus solely on visually verifiable details.\n'
    '- Omit any uncertain or ambiguous elements.\n\n'
    'Please strictly follow the output format below:\n\n'
    '<description>\n'
    '    <prompt>{prompt}</prompt>\n'
    '    <checkpoint>{testpoint}</checkpoint>\n'
    '    <analysis>A list using square brackets `[]`, where each element is a string '
    'of detailed analysis corresponding to one checkpoint. **Ensure the list length '
    'matches the number of checkpoints**.</analysis>\n'
    '    <score>A list using square brackets `[]`, where each element is a binary '
    'score (0 or 1). **Ensure the list length matches the number of checkpoints**.</score>\n'
    '</description>'
)
\end{lstlisting}

\subsection{Prompts for MAPE-Edit}
\label{appendix:prompts_edit}
\begin{lstlisting}[caption={Prompt for MAPE-Edit router}, label={lst:prompt_MAPE-Edit_router}]
_ROUTER_SYSTEM_PROMPT = f"""You are an expert image editing analyst. Given an original image and an editing instruction, your task is to:

1. Decide whether the instruction needs prompt rewriting or is already clear enough for the editing model.

2. If rewriting is needed, classify the editing instruction into one or more action types from this list:
   {json.dumps(ROUTER_ACTION_TYPES)}

3. Analyze the image to identify the target object(s), region(s), and the specific changes requested.

4. Extract structured information to guide specialized rewriters.

**When to set no_rewrite to true (use the original instruction as-is):**
- The instruction is already clear, specific, and unambiguous (e.g. "Remove the sheep in the foreground")
- The edit is a straightforward operation on an obvious target (remove, simple color/attribute change, well-known style transfer)
- Adding more detail would not help the editing model and could confuse it
- The target object is already clearly identified in the instruction

**When to set no_rewrite to false (rewrite the instruction):**
- The instruction is vague or ambiguous (e.g. "Extract the vehicle" - which vehicle?)
- The instruction involves complex spatial relationships that need grounding
- Multiple objects or operations need coordination
- The target is not clearly identified and visual analysis would add value

Action type definitions:
- add: Adding new objects to the scene
- adjust: Changing attributes (color, material, texture, size) of existing objects
- remove: Removing objects from the scene
- replace: Replacing one object with another
- background: Changing the background while preserving foreground
- style: Changing the artistic style or visual tone
- action: Changing poses, expressions, or motions of characters
- extract: Isolating an object onto a clean background

If the instruction involves multiple editing operations, list ALL applicable action types \
(e.g. ["add", "remove"] for an instruction that adds one object and removes another). \
Do NOT use "compose" or "hybrid" as an action type.

Output ONLY a valid JSON object with these keys:
- "no_rewrite": boolean - true if the original instruction should be used as-is, false if rewriting is needed
- "action_types": list of applicable action type strings (usually 1, use 2+ for multi-operation edits)
- "target_objects": list of objects being edited (description + approximate location)
- "edit_details": object describing the specific changes requested
- "preservation_requirements": what must remain unchanged
- "reasoning": brief explanation of the classification and the no_rewrite decision

Do not include any text before or after the JSON object."""

_ROUTER_USER_TEMPLATE = (
    "Analyze the following editing instruction for the provided image "
    "and determine the action type(s) and editing details.\n\n"
    "Editing instruction: {instruction}"
)
\end{lstlisting}

\begin{lstlisting}[caption={Prompt for MAPE-Edit domain rewriter}, label={lst:prompt_MAPE-Edit_rewriter}]
_REWRITER_SCHEMAS = {
    "add": {
        "description": "Specialize in adding new objects to a scene",
        "output_schema": {
            "object_to_add": "Detailed description of the object to add",
            "placement": "Exact location in the image (e.g., center-right, foreground)",
            "relative_size": "Size relative to existing elements",
            "integration": "How the object should blend (lighting, shadows, perspective)",
            "attributes": "Color, texture, material, pose details",
        },
    },
    "adjust": {
        "description": "Specialize in modifying attributes of existing objects",
        "output_schema": {
            "target_object": "Which object(s) to modify",
            "attribute_type": "What attribute to change (color, material, texture, size)",
            "current_value": "Current state of the attribute (from the image)",
            "desired_value": "Target state as specified by the instruction",
            "change_scope": "Exact region/boundary of the change",
        },
    },
    "remove": {
        "description": "Specialize in removing objects from a scene",
        "output_schema": {
            "target_object": "Which object(s) to remove",
            "object_location": "Precise location in the image",
            "fill_strategy": "How to fill the gap (background continuation, inpainting)",
            "edge_handling": "How to handle edges and transitions",
        },
    },
    "replace": {
        "description": "Specialize in replacing one object with another",
        "output_schema": {
            "original_object": "Object being replaced (description, location)",
            "replacement_object": "New object to place (description, attributes)",
            "spatial_consistency": "Size, position, orientation matching requirements",
            "attribute_matching": "Color, lighting, style consistency needs",
        },
    },
    "background": {
        "description": "Specialize in changing image backgrounds",
        "output_schema": {
            "new_background": "Detailed description of the target background",
            "foreground_preservation": "Which foreground elements to preserve exactly",
            "blending_requirements": "Edge handling, lighting consistency, perspective matching",
            "atmospheric_matching": "Color grading, mood, depth of field adjustments",
        },
    },
    "style": {
        "description": "Specialize in style transfer and artistic transformation",
        "output_schema": {
            "target_style": "Detailed description of the desired style",
            "transfer_scope": "Global vs. selective style application",
            "content_preservation": "What content/structure must be maintained",
            "style_parameters": "Color palette, brushwork, texture, lighting changes",
        },
    },
    "action": {
        "description": "Specialize in changing poses, motions, and expressions",
        "output_schema": {
            "target_subject": "Which character/subject to modify",
            "current_pose": "Current pose/expression (from the image)",
            "target_pose": "Desired pose/expression/motion",
            "identity_preservation": "Facial features, clothing, accessories to maintain",
            "anatomical_constraints": "Natural body mechanics and proportions",
        },
    },
    "extract": {
        "description": "Specialize in isolating objects from their background",
        "output_schema": {
            "target_object": "Which object to extract",
            "object_boundary": "Edge detail level and handling approach",
            "background_type": "Target background (typically white/clean)",
            "preservation_details": "Fine details to preserve (hair, transparent edges)",
        },
    },
    "compose": {
        "description": "Specialize in coordinating multiple editing operations",
        "output_schema": {
            "operations": "List of individual operations with their details",
            "operation_order": "Recommended execution sequence",
            "cross_operation_consistency": "How operations should interact",
            "global_coherence": "Overall image consistency requirements",
        },
    },
}

_REWRITER_SYSTEM_TEMPLATE = """You are a specialist image editing analyst focused on "{action_type}" edits.

Your task: Given an original image and an editing instruction, generate a detailed JSON specification for a {action_type} edit.

Your output should contain these fields:
{schema_str}

Additionally, always include:
- "global_layout_perception": Describe the overall scene layout with key objects and positions
- "local_object_perception": Describe the target object(s) in detail (shape, color, texture, state)
- "edit_area_localization": Specify exactly which regions will be modified
- "edited_image_imagination": Describe the expected appearance after editing

Context from router: {router_context}

Output ONLY a valid JSON object. Do not include any text before or after the JSON object."""

_REWRITER_USER_TEMPLATE = (
    "Generate a detailed editing specification for this {action_type} edit.\n\n"
    "Editing instruction: {instruction}"
)
\end{lstlisting}

\begin{lstlisting}[caption={Prompt for MAPE-Edit composer}, label={lst:prompt_MAPE-Edit_composer}]
_COMPOSER_SYSTEM_PROMPT_DEFAULT = """You are an expert prompt engineer for image editing models. \
Your task is to take structured editing specifications and compose them into a clear, \
detailed natural language editing prompt.

Rules:
1. Incorporate ALL details from the specification: target objects, desired changes, \
spatial information, preservation requirements.
2. Structure the output as a chain-of-thought blueprint with these sections:
   - Global Layout: Describe the overall scene
   - Target Region: Identify what will be modified
   - Edit Specification: Detail the exact changes
   - Expected Result: Describe the final appearance
3. Be specific about locations, sizes, colors, and relationships.
4. Include preservation requirements (what should NOT change).
5. Output ONLY the editing prompt text. No JSON, no explanations."""

_COMPOSER_SYSTEM_PROMPT_FLUX2 = """You are an expert prompt engineer for FLUX.2 image editing models. \
Your task is to take structured editing specifications and compose them into one concise \
editing instruction (50-80 words, ~30 for simple edits).

Rules:
1. Distill ALL the editing specifications into a single, direct instruction.
2. Use clear, analytical language - avoid flowery or vague words \
(no "whimsical," "cascading," "ethereal," etc.).
3. Specify what changes AND what stays the same (face, lighting, composition).
4. Reference actual image elements identified in the specifications.
5. Turn negatives into positives ("don't change X" -> "keep X").
6. Make abstractions concrete ("futuristic" -> "glowing cyan neon, metallic panels").
7. Do NOT structure the output with sections or bullet points - write a single \
flowing instruction paragraph.
8. Output ONLY the final instruction in plain text. No JSON, no explanations, no commentary."""

_COMPOSER_SYSTEM_PROMPT_QWEN_IMGEDIT = """You are a professional edit prompt composer for Qwen-Image-Edit. \
Your task is to take structured editing specifications and compose them into a direct and specific \
edit prompt.

Rules:
1. Keep the composed prompt **direct and specific** - one clear instruction.
2. Preserve the core intention of the original instruction; only enhance clarity and visual feasibility.
3. For Add/Delete/Replace tasks: if already clear, refine grammar only. If vague, add minimal \
details (category, color, size, position). For replacement, specify "Replace Y with X" and \
describe key visual features of X.
4. For Style tasks: describe the style concisely using key visual features.
5. For Human/ID editing: emphasize maintaining the person's visual consistency \
(ethnicity, gender, hairstyle, expression, outfit). Changes must be natural, never exaggerated.
6. All text content must be enclosed in English double quotes.
7. Do NOT structure the output with sections or bullet points - write a single flowing instruction.
8. Output ONLY the final instruction in plain text. No JSON, no explanations, no commentary."""

_COMPOSER_SYSTEM_PROMPTS = {
    "default": _COMPOSER_SYSTEM_PROMPT_DEFAULT,
    "flux2-klein": _COMPOSER_SYSTEM_PROMPT_FLUX2,
    "qwen-image-edit": _COMPOSER_SYSTEM_PROMPT_QWEN_IMGEDIT,
}

_COMPOSER_USER_TEMPLATE = (
    "Compose the following editing specifications into a detailed editing prompt.\n\n"
    "Original instruction: {instruction}\n\n"
    "Editing specifications:\n{spec_json_str}"
)
\end{lstlisting}

\begin{lstlisting}[caption={Prompt for ImgEdit scoring}, label={lst:prompt_MAPE-Edit_scoring}]
IMGEDIT_SCORING_RUBRICS = {
    "replace": (
        '\nYou are a data rater specializing in grading image replacement edits. '
        'You will be given two images (before and after editing) and the corresponding editing instructions. '
        'Your task is to evaluate the replacement editing effect on a 5-point scale from three perspectives:\n\n'
        'Prompt Compliance\n'
        '1 Target not replaced, or an unrelated object edited.\n'
        '2 Only part of the target replaced, or wrong class/description used.\n'
        '3 Target largely replaced but other objects altered, remnants visible, or count/position clearly wrong.\n'
        '4 Correct object fully replaced; only minor attribute errors (colour, size, etc.).\n'
        '5 Perfect replacement: all and only the specified objects removed; new objects\' class, number, position, scale, pose and detail exactly match the prompt.\n\n'
        'Visual Naturalness\n'
        '1 Image heavily broken or new object deformed / extremely blurred.\n'
        '2 Obvious seams, smears, or strong mismatch in resolution or colour; background not restored.\n'
        '3 Basic style similar, but lighting or palette clashes; fuzzy edges or noise are noticeable.\n'
        '4 Style almost uniform; tiny edge artefacts visible only on close inspection; casual viewers see no edit.\n'
        '5 Completely seamless; new objects blend fully with the scene, edit area undetectable.\n\n'
        'Physical & Detail Integrity\n'
        '1 Floating, interpenetration, severe perspective/light errors; key original elements ruined; background heavily warped.\n'
        '2 Missing shadows/occlusion; large background shifts or holes.\n'
        '3 Lighting, perspective and contact surfaces mostly correct; small but tolerable errors; background adjusted locally.\n'
        '4 New objects interact realistically with scene (shadows, reflections, texture) and preserve existing details; background change minimal.\n'
        '5 Physically flawless and enhances realism: accurate highlights, shadows, reflections, ambient effects; background untouched.\n'
        'The second and third score should no higher than first score!!!\n\n'
        'Example Response Format:\n'
        'Brief reasoning: A short explanation of the score based on the criteria above, no more than 20 words.\n'
        'Prompt Compliance: A number from 1 to 5.\n'
        'Visual Naturalness: A number from 1 to 5.\n'
        'Physical & Detail Integrity: A number from 1 to 5.\n'
        'editing instruction is :.\n\n'
        'Below are the images before and after editing:\n'
    ),
    "add": (
        '\nYou are a data rater specializing in grading image addition edits. '
        'You will be given two images (before and after editing) and the corresponding editing instructions. '
        'Your task is to evaluate the added object(s) on a 5-point scale from three perspectives:\n\n'
        'Prompt Compliance\n'
        '1 Nothing added or the added content is corrupt.\n'
        '2 Added object is a wrong class or unrelated to the prompt.\n'
        '3 Correct class, but key attributes (position, colour, size, count, etc.) are wrong.\n'
        '4 Main attributes correct; only minor details off or 1-2 small features missing.\n'
        '5 Every stated attribute correct and scene logic reasonable; only microscopic flaws.\n\n'
        'Visual Naturalness\n'
        '1 Image badly broken or full of artefacts.\n'
        '2 Obvious paste marks; style, resolution, or palette strongly mismatch.\n'
        '3 General style similar, but lighting or colours clearly clash; noticeable disharmony.\n'
        '4 Style almost uniform; small edge issues visible only when zoomed.\n'
        '5 Perfect blend; no visible difference between added object and original image.\n\n'
        'Physical & Detail Coherence\n'
        '1 Severe physical errors (floating, wrong perspective/light); key original elements blocked; background heavily distorted.\n'
        '2 Contact or occlusion handled poorly; minor background shifts, jaggies or noise; background visibly changed.\n'
        '3 Lighting, perspective, and contact mostly correct; remaining flaws small and acceptable; limited background change.\n'
        '4 Shadows, reflections, and material response believable; no loss of original detail; background changes are minute.\n'
        '5 Added object enhances overall realism: precise highlights, shadows, ambient effects; background essentially untouched.\n'
        'The second and third score should no higher than first score!!!\n\n'
        'Example Response Format:\n'
        'Brief reasoning: A short explanation of the score based on the criteria above, no more than 20 words.\n'
        'Prompt Compliance: A number from 1 to 5.\n'
        'Visual Naturalness: A number from 1 to 5.\n'
        'Physical & Detail Coherence: A number from 1 to 5.\n'
        'editing instruction is :.\n\n'
        'Below are the images before and after editing:\n'
    ),
    "adjust": (
        '\nYou are a data rater specializing in grading attribute alteration edits. '
        'You will be given two images (before and after editing) and the corresponding editing instructions. '
        'Your task is to evaluate the attribute change on a 5-point scale from three perspectives:\n\n'
        'Prompt Compliance\n'
        '1 Target not adjusted, wrong object touched, or geometry changed.\n'
        '2 Right object but wrong attribute value/direction; only part edited; other objects also altered; slight stretch/crop.\n'
        '3 Mainly correct object and attribute, yet large hue/brightness/texture error; minor collateral edits; visible jaggies/distortion.\n'
        '4 All requested objects adjusted, only their attributes changed; shape kept; small inaccuracy in colour, material or amount.\n'
        '5 Exactly and only the requested objects adjusted; colour, material, gloss etc. match the prompt perfectly; shape 100% intact; zero unintended edits.\n\n'
        'Visual Seamlessness\n'
        '1 Massive colour spill, mosaics or heavy noise; image nearly unusable.\n'
        '2 Clear smears/bleeding on edges; abrupt resolution or tone shift; highlights/shadows clipped; background gaps.\n'
        '3 Overall palette OK but local tone or grain conflicts; soft edges; noticeable disharmony.\n'
        '4 Style unified, transitions smooth; only slight edge artefacts visible when zoomed.\n'
        '5 No detectable edit traces; colours/materials fuse with scene lighting; edit area practically invisible.\n\n'
        'Physical & Detail Fidelity\n'
        '1 Object floating, interpenetrating, or severe perspective/light mismatch; background badly warped.\n'
        '2 Missing shadows/highlights; wrong reflection direction; background visibly discoloured or distorted.\n'
        '3 Light, perspective and contact surface largely correct; minor acceptable flaws; background only locally affected.\n'
        '4 Adjusted material interacts believably with scene; shadows, highlights, reflections handled well; original details preserved.\n'
        '5 High physical realism: fine micro-highlights, diffuse bounce, subsurface effects present; overall scene realism improved.\n'
        'The second and third score should no higher than first score!!!\n\n'
        'Example Response Format:\n'
        'Brief reasoning: A short explanation of the score based on the criteria above, no more than 20 words.\n'
        'Prompt Compliance: A number from 1 to 5.\n'
        'Visual Seamlessness: A number from 1 to 5.\n'
        'Physical & Detail Fidelity: A number from 1 to 5.\n'
        'editing instruction is :.\n\n'
        'Below are the images before and after editing:\n'
    ),
    "remove": (
        '\nYou are a data rater specializing in grading object removal edits. '
        'You will be given two images (before and after editing) and the corresponding editing instructions. '
        'Your task is to evaluate the removal quality on a 5-point scale from three perspectives:\n\n'
        'Prompt Compliance\n'
        '1 Nothing removed, or an unrelated object edited.\n'
        '2 Target only partly removed, or a different instance/class deleted, or another object appears in the gap.\n'
        '3 Target mostly removed but extra objects also deleted, or fragments of the target remain.\n'
        '4 Only the specified objects removed, but a few tiny/background items deleted by mistake, or the count is wrong.\n'
        '5 Perfect: all and only the requested objects removed; every other element untouched.\n\n'
        'Visual Naturalness\n'
        '1 Image badly broken (large holes, strong artefacts).\n'
        '2 Clear erase marks; colour/resolution mismatch; background not restored.\n'
        '3 General look acceptable yet lighting/colour/style still clash; blur or noise visible.\n'
        '4 Style consistent; minor edge issues visible only when zoomed.\n'
        '5 Seamless: removal is virtually impossible to spot.\n\n'
        'Physical & Detail Integrity\n'
        '1 Severe physical errors (floating items, wrong perspective/light); key scene elements damaged; background heavily warped.\n'
        '2 Large un-filled gaps or obvious background shifts.\n'
        '3 Lighting, perspective and contacts mostly correct; flaws small and tolerable; background adjusted locally.\n'
        '4 Background reconstruction clean; existing details preserved; only minute changes outside the removal area.\n'
        '5 Physically flawless and even enhances realism: accurate light/shadow/texture infill, high-quality micro-details.\n'
        'The second and third score should no higher than first score!!!\n\n'
        'Example Response Format:\n'
        'Brief reasoning: A short explanation of the score based on the criteria above, no more than 20 words.\n'
        'Prompt Compliance: A number from 1 to 5.\n'
        'Visual Naturalness: A number from 1 to 5.\n'
        'Physical & Detail Integrity: A number from 1 to 5.\n'
        'editing instruction is :.\n\n'
        'Below are the images before and after editing:\n'
    ),
    "style": (
        '\nYou are a data rater specializing in grading style transfer edits. '
        'You will be given an input image, a reference style, and the styled result. '
        'Your task is to evaluate the style transfer on a 5-point scale from three perspectives:\n\n'
        'Style Fidelity\n'
        '1 Target style absent or clearly wrong.\n'
        '2 Style shows in a few areas only, or mixed with unrelated styles.\n'
        '3 Key traits (palette, brushwork, texture) present but patchy or inconsistent.\n'
        '4 Style reproduced across almost the whole image; only small local mismatches.\n'
        '5 Full, faithful transfer: colour, texture, brushwork, lighting all match the exemplar over the entire image.\n\n'
        'Content Preservation\n'
        '1 Major objects or layout lost/distorted; original scene barely recognisable.\n'
        '2 Main subject recognisable, but size, perspective or key parts clearly wrong/missing.\n'
        '3 Overall structure correct; some local warping or minor omissions.\n'
        '4 Nearly all geometry intact; only slight, non-distracting deformation.\n'
        '5 All objects and spatial relations kept; only stylistic, harmless distortion.\n\n'
        'Rendering Quality\n'
        '1 Heavy noise, banding, pixel damage or blur; image unusable.\n'
        '2 Visible seams, aliasing, colour drift; low resolution or chaotic strokes.\n'
        '3 Moderate quality: local blur/noise/texture breaks, but generally acceptable.\n'
        '4 Sharp, coherent strokes; tiny artefacts visible only when zoomed.\n'
        '5 High resolution, no artefacts; strokes, textures and colour transitions look fully natural.\n'
        'The second and third score should no higher than first score!!!\n\n'
        'Example Response Format:\n'
        'Brief reasoning: A short explanation of the score based on the criteria above, no more than 20 words.\n'
        'Style Fidelity: A number from 1 to 5.\n'
        'Content Preservation: A number from 1 to 5.\n'
        'Rendering Quality: A number from 1 to 5.\n'
        'editing instruction is :.\n\n'
        'Below are the images before and after editing:\n'
    ),
    "action": (
        '\nYou are a data rater specializing in grading action or expression change edits. '
        'You will be given two images (before and after editing) and the editing instruction. '
        'Your task is to evaluate the motion or expression change on a 5-point scale from three perspectives:\n\n'
        'Action / Expression Fidelity\n'
        '1 No visible change, or wrong action / expression.\n'
        '2 Partial or clearly incorrect pose; only some body parts change; expression direction wrong.\n'
        '3 Main idea present but details off (angle, side, intensity, missing gesture).\n'
        '4 Requested pose / expression achieved with just minor inaccuracy (small angular drift, timing nuance).\n'
        '5 Exact match to prompt: every limb, gesture, and facial muscle aligns with the described action.\n\n'
        'Identity Preservation\n'
        '1 Person unrecognisable; face or body replaced.\n'
        '2 Strong drift: key facial features, hairstyle or clothing heavily altered.\n'
        '3 Mostly same identity; moderate changes in some features but still recognisable.\n'
        '4 Identity clearly the same; only subtle stylisation or lighting differences.\n'
        '5 Perfect preservation of face, hairstyle, skin tone, clothing and accessories.\n\n'
        'Visual & Anatomical Coherence\n'
        '1 Severe artifacts: broken or duplicated limbs, extreme distortion, heavy noise/blur.\n'
        '2 Noticeable cut-out halos, proportion errors, lighting or perspective clearly off.\n'
        '3 Generally plausible; minor joint or shading issues; small noise/blur acceptable.\n'
        '4 Clean render; anatomy, lighting, depth and edges consistent; flaws only on close inspection.\n'
        '5 Flawless realism or stylistic coherence; perfect anatomy, lighting, shadows and texture continuity.\n'
        'The second and third score should no higher than first score!!!\n\n'
        'Example Response Format:\n'
        'Brief reasoning: A short explanation of the score based on the criteria above, no more than 20 words.\n'
        'Action Fidelity: A number from 1 to 5.\n'
        'Identity Preservation: A number from 1 to 5.\n'
        'Visual & Anatomical Coherence: A number from 1 to 5.\n'
        'editing instruction is :.\n\n'
        'Below are the images before and after editing:\n'
    ),
    "extract": (
        '\nYou are a data rater specializing in grading object cut-out quality. '
        'You will be given an image with the object extracted on a white background. '
        'Your task is to evaluate the cut-out accuracy on a 5-point scale from three perspectives:\n\n'
        'Object Selection & Identity\n'
        '1 Wrong object or multiple objects extracted.\n'
        '2 Correct class but only part of the object, or obvious intrusions from other items.\n'
        '3 Object largely correct yet small pieces missing / extra, identity still recognisable.\n'
        '4 Full object with clear identity; only tiny mis-crop (e.g., tip of antenna).\n'
        '5 Exact requested object, complete and unmistakably the same instance (ID).\n\n'
        'Mask Precision & Background Purity\n'
        '1 Large background remnants, holes in mask, or non-white backdrop dominates.\n'
        '2 Noticeable jagged edges, colour fringes, grey/colour patches in white area.\n'
        '3 Acceptable mask; minor edge softness or faint halo visible on close look.\n'
        '4 Clean, smooth edges; white (#FFFFFF) background uniform, tiny artefacts only when zoomed.\n'
        '5 Crisp anti-aliased contour, zero spill or halo; backdrop perfectly pure white throughout.\n\n'
        'Object Integrity & Visual Quality\n'
        '1 Severe blur, compression, deformation, or missing parts; unusable.\n'
        '2 Moderate noise, colour shift, or slight warping; details clearly degraded.\n'
        '3 Overall intact with minor softness or noise; colours mostly preserved.\n'
        '4 Sharp detail, accurate colours; negligible artefacts.\n'
        '5 Pristine: high-resolution detail, true colours, no artefacts or distortion.\n'
        'The second and third score should no higher than first score!!!\n\n'
        'Example Response Format:\n'
        'Brief reasoning: A short explanation of the score based on the criteria above, no more than 20 words.\n'
        'Object Identity: A number from 1 to 5.\n'
        'Mask Precision: A number from 1 to 5.\n'
        'Visual Quality: A number from 1 to 5.\n'
        'editing instruction is :.\n\n'
        'Below is the extracted object image:\n'
    ),
    "background": (
        '\nYou are a data rater specializing in grading background editing. '
        'You will be given two images (before and after editing) and the editing instruction. '
        'Your task is to evaluate the background change on a 5-point scale from three perspectives:\n\n'
        'Instruction Compliance\n'
        '1 No change, or background unrelated to prompt, or foreground also replaced/distorted.\n'
        '2 Background partly replaced or wrong style/content; foreground noticeably altered.\n'
        '3 Main background replaced but elements missing/extra, or faint spill onto subject edges.\n'
        '4 Requested background fully present; foreground intact except minute artefacts or small prompt mismatch (e.g. colour tone).\n'
        '5 Background exactly matches prompt (content, style, placement); all foreground pixels untouched.\n\n'
        'Visual Seamlessness (Edge & Texture Blend)\n'
        '1 Large tearing, posterisation, extreme blur/noise; edit area obvious at a glance.\n'
        '2 Clear cut-out halos, colour-resolution gap, or heavy smudge strokes.\n'
        '3 Blend acceptable but visible on closer look: slight edge blur, grain or palette shift.\n'
        '4 Nearly invisible seams; textures and sharpness aligned, only minor issues when zoomed in.\n'
        '5 Indistinguishable composite: edges, textures, resolution and colour grading perfectly continuous.\n\n'
        'Physical Consistency (Lighting, Perspective, Depth)\n'
        '1 Severe mismatch: wrong horizon, conflicting light direction, floating subject, warped geometry.\n'
        '2 Noticeable but not extreme inconsistencies in light, shadows or scale; depth cues off.\n'
        '3 Overall believable; small errors in shadow length, perspective or ambient colour.\n'
        '4 Lighting, scale, depth, and camera angle well matched; only subtle discrepancies.\n'
        '5 Physically flawless: foreground and new background share coherent light, shadows, reflections, perspective and atmospheric depth, enhancing overall realism.\n'
        'The second and third score should no higher than first score!!!\n\n'
        'Example Response Format:\n'
        'Brief reasoning: A short explanation of the score based on the criteria above, no more than 20 words.\n'
        'Instruction Compliance: A number from 1 to 5.\n'
        'Visual Seamlessness: A number from 1 to 5.\n'
        'Physical Consistency: A number from 1 to 5.\n'
        'editing instruction is :.\n\n'
        'Below are the images before and after editing:\n'
    ),
    "compose": (
        '\nYou are a data rater specializing in grading hybrid image edits '
        '(involving multiple operations on multiple objects). '
        'You will be given two images (before and after editing) and the editing instruction. '
        'Your task is to evaluate the overall editing quality on a 5-point scale from three perspectives:\n\n'
        'Instruction Compliance\n'
        '1 Neither object nor operations match the prompt; wrong items edited or shapes distorted.\n'
        '2 Only one object correctly edited, or both edited but with wrong/partial operations; collateral changes to other items.\n'
        '3 Both target objects touched, each with the requested operation broadly correct but missing details (e.g., wrong colour value, incomplete removal).\n'
        '4 Both objects receive the exact operations; tiny deviations in amount, position, or parameter. No unintended edits elsewhere.\n'
        '5 Perfect execution: each object fully reflects its specified operation, all other scene elements untouched.\n\n'
        'Visual Naturalness (Seamlessness)\n'
        '1 Large artefacts, obvious cut-outs, heavy blur/noise; edits conspicuous at a glance.\n'
        '2 Clear edge halos, colour or resolution mismatch, awkward scaling.\n'
        '3 Acceptable but visible on close look: slight edge softness, minor palette or focus shift.\n'
        '4 Edits blend smoothly; seams hard to spot, textures and sharpness largely consistent.\n'
        '5 Indistinguishable composite: colour grading, grain, resolution and style fully match the original image.\n\n'
        'Physical Consistency & Fine Detail\n'
        '1 Severe lighting/perspective mismatch, missing or wrong shadows; objects appear floating or warped.\n'
        '2 Noticeable but tolerable inconsistencies in illumination, scale, or depth cues.\n'
        '3 Generally plausible; small errors in shadow length, reflection angle, or texture alignment.\n'
        '4 Lighting, perspective, and material response closely match; only subtle flaws visible when zoomed.\n'
        '5 Physically flawless: shadows, highlights, reflections, depth and texture perfectly integrated, enhancing overall realism.\n'
        'The second and third score should no higher than first score!!!\n\n'
        'Example Response Format:\n'
        'Brief reasoning: A short explanation of the score based on the criteria above, no more than 20 words.\n'
        'Instruction Compliance: A number from 1 to 5.\n'
        'Visual Naturalness: A number from 1 to 5.\n'
        'Physical Consistency & Fine Detail: A number from 1 to 5.\n'
        'editing instruction is :.\n\n'
        'Below are the images before and after editing:\n'
    ),
}
\end{lstlisting}

\section{Related Works}
\label{appendix:related}
\textbf{Generative Models for Image Synthesis.}
Image synthesis has been extensively studied over the past decade, evolving from early autoregressive \cite{van2016conditional,parmar2018image,ramesh2021zero} and GAN-based \cite{goodfellow2020generative,karras2019style,radford2015unsupervised} approaches to more scalable and expressive generative paradigms. In recent years, progress has been largely driven by diffusion-based models \cite{ho2020denoising,nichol2021improved,song2020score}, beginning with Latent Diffusion Models (LDMs) \cite{rombach2022high}, which enable large-scale image synthesis by operating in a compressed latent space. Building on this foundation, recent advances \cite{peebles2023scalable,esser2024scaling} have shifted toward flow-matching objectives \cite{lipman2022flow} and transformer-based architectures. More recent models, such as FLUX~\cite{flux-2-2025}, Z-Image~\cite{cai2025z} and Qwen-Image \cite{wu2025qwen}, further enhance instruction following and fine-grained visual fidelity.

Image editing has progressed alongside generation, typically building on pretrained generative backbones. FLUX.1-Kontext \cite{labs2025flux} and FLUX.2-klein \cite{flux-2-2025} support strong instruction alignment and multi-turn editing. Qwen-Image-Edit \cite{wu2025qwen} adopts a dual-stream vision-language architecture for precise edits, and ChronoEdit \cite{wu2025chronoedit} leverages temporal priors from large pretrained video generative models to enable physically consistent edits.

Across both tasks, prompt quality is a key determinant of performance: richer, well-grounded descriptions improve semantic alignment, attribute binding, and edit faithfulness \cite{betker2023improving,gutflaish2025generating}. This motivates us to treat prompt formulation as a trainable component of the visual pipeline.

\textbf{Prompt Enhancement.} Prompt enhancement has emerged as an effective way to improve generative model performance without modifying the underlying generator itself. Prior work in NLP formulates prompt design as an optimization problem, where prompts are automatically searched, evolved, or selected to maximize downstream performance. Representative examples include Automatic Prompt Engineer \cite{zhou2022large}, which generates and ranks candidate instructions, and Promptbreeder \cite{fernandopromptbreeder}, which evolves prompts through self-referential mutation and selection. In text-to-image generation, prompt rewriting methods similarly aim to bridge the gap between underspecified human instructions and the richer textual descriptions preferred by visual generators. Existing approaches use LLMs to ground complex semantics or train prompt rewriters that expand prompts toward better alignment and compositional fidelity \cite{lianllm,wang2025promptenhancer}. However, most prompt enhancers remain monolithic: they rewrite the prompt in a single stage, often relying on prompting or powerful commercial models rather than post-training small enhancers with task-aware rewards. In contrast, APE studies SLMs as trainable prompt-enhancement agents, and MAPE further decomposes enhancement into role-specialized decisions for selecting, refining, and composing semantic fields. 

\textbf{Multi-Agent Systems (MASs).} Multi-agent systems (MASs) have become a prominent paradigm for extending LLM capabilities by decomposing complex tasks into coordinated interactions among specialized agents. Recent surveys characterize LLM-based MASs along dimensions such as agent profiles, communication protocols, memory, tool use, and coordination structure, identifying role specialization and inter-agent feedback as central design principles \cite{guo2024large,li2024survey,tran2025multi}. Representative frameworks include CAMEL \cite{li2023camel}, which studies autonomous cooperation through role-playing agents, AutoGen \cite{wu2024autogen}, which provides a flexible abstraction for multi-agent conversation among LLMs, humans, and tools, and workflow-driven systems such as MetaGPT \cite{hong2023metagpt}, ChatDev \cite{qian2024chatdev}, and AgentVerse \cite{chen2023agentverse}, which show that structured role assignment and coordination can improve performance and controllability on complex tasks. Beyond fixed workflows, multi-agent reasoning has also been explored through debate-based collaboration, where multiple agents exchange and refine candidate solutions, as well as through recent studies emphasizing diversity of thought and collaborative reasoning behavior in multi-agent settings \cite{hegazy2024diversity,hu2025multi}. Inspired by this literature, we view prompt enhancement as a naturally decomposable problem and instantiate it as a multi-agent pipeline with specialized roles for field selection, field-specific rewriting, and final prompt composition.

%% file: ref.bib
@inproceedings{zhou2022large,
  title={Large language models are human-level prompt engineers},
  author={Zhou, Yongchao and Muresanu, Andrei Ioan and Han, Ziwen and Paster, Keiran and Pitis, Silviu and Chan, Harris and Ba, Jimmy},
  booktitle={The eleventh international conference on learning representations},
  year={2022}
}

@inproceedings{fernandopromptbreeder,
  title={Promptbreeder: Self-Referential Self-Improvement via Prompt Evolution},
  author={Fernando, Chrisantha and Banarse, Dylan Sunil and Michalewski, Henryk and Osindero, Simon and Rockt{\"a}schel, Tim},
  booktitle={Forty-first International Conference on Machine Learning}
}

@article{lianllm,
  title={LLM-grounded Diffusion: Enhancing Prompt Understanding of Text-to-Image Diffusion Models with Large Language Models},
  author={Lian, Long and Li, Boyi and Yala, Adam and Darrell, Trevor},
  journal={Transactions on Machine Learning Research}
}

@article{wang2025promptenhancer,
  title={Promptenhancer: A simple approach to enhance text-to-image models via chain-of-thought prompt rewriting},
  author={Wang, Linqing and Xing, Ximing and Cheng, Yiji and Zhao, Zhiyuan and Li, Donghao and Hang, Tiankai and Tao, Jiale and Wang, Qixun and Li, Ruihuang and Chen, Comi and others},
  journal={arXiv preprint arXiv:2509.04545},
  year={2025}
}

@article{li2023camel,
  title={Camel: Communicative agents for" mind" exploration of large language model society},
  author={Li, Guohao and Hammoud, Hasan and Itani, Hani and Khizbullin, Dmitrii and Ghanem, Bernard},
  journal={Advances in neural information processing systems},
  volume={36},
  pages={51991--52008},
  year={2023}
}

@inproceedings{wu2024autogen,
  title={Autogen: Enabling next-gen LLM applications via multi-agent conversations},
  author={Wu, Qingyun and Bansal, Gagan and Zhang, Jieyu and Wu, Yiran and Li, Beibin and Zhu, Erkang and Jiang, Li and Zhang, Xiaoyun and Zhang, Shaokun and Liu, Jiale and others},
  booktitle={First conference on language modeling},
  year={2024}
}

@inproceedings{hong2023metagpt,
  title={MetaGPT: Meta programming for a multi-agent collaborative framework},
  author={Hong, Sirui and Zhuge, Mingchen and Chen, Jonathan and Zheng, Xiawu and Cheng, Yuheng and Wang, Jinlin and Zhang, Ceyao and Wang, Zili and Yau, Steven Ka Shing and Lin, Zijuan and others},
  booktitle={The twelfth international conference on learning representations},
  year={2023}
}

@inproceedings{qian2024chatdev,
  title={Chatdev: Communicative agents for software development},
  author={Qian, Chen and Liu, Wei and Liu, Hongzhang and Chen, Nuo and Dang, Yufan and Li, Jiahao and Yang, Cheng and Chen, Weize and Su, Yusheng and Cong, Xin and others},
  booktitle={Proceedings of the 62nd annual meeting of the association for computational linguistics (volume 1: Long papers)},
  pages={15174--15186},
  year={2024}
}

@inproceedings{guo2024large,
  title={Large Language Model based Multi-Agents: A Survey of Progress and Challenges.},
  author={Guo, T and Chen, X and Wang, Y and Chang, R and Pei, S and Chawla, NV and Wiest, O and Zhang, X},
  booktitle={33rd International Joint Conference on Artificial Intelligence (IJCAI 2024)},
  year={2024},
  organization={IJCAI; Cornell arxiv}
}

@article{shao2024deepseekmath,
  title={Deepseekmath: Pushing the limits of mathematical reasoning in open language models},
  author={Shao, Zhihong and Wang, Peiyi and Zhu, Qihao and Xu, Runxin and Song, Junxiao and Bi, Xiao and Zhang, Haowei and Zhang, Mingchuan and Li, YK and Wu, Yang and others},
  journal={arXiv preprint arXiv:2402.03300},
  year={2024}
}

@article{liu2026gdpo,
  title={Gdpo: Group reward-decoupled normalization policy optimization for multi-reward rl optimization},
  author={Liu, Shih-Yang and Dong, Xin and Lu, Ximing and Diao, Shizhe and Belcak, Peter and Liu, Mingjie and Chen, Min-Hung and Yin, Hongxu and Wang, Yu-Chiang Frank and Cheng, Kwang-Ting and others},
  journal={arXiv preprint arXiv:2601.05242},
  year={2026}
}

@article{gutflaish2025generating,
  title={Generating an Image From 1,000 Words: Enhancing Text-to-Image With Structured Captions},
  author={Gutflaish, Eyal and Kachlon, Eliran and Zisman, Hezi and Hacham, Tal and Sarid, Nimrod and Visheratin, Alexander and Huberman, Saar and Davidi, Gal and Bukchin, Guy and Goldberg, Kfir and others},
  journal={arXiv preprint arXiv:2511.06876},
  year={2025}
}

@inproceedings{rombach2022high,
  title={High-resolution image synthesis with latent diffusion models},
  author={Rombach, Robin and Blattmann, Andreas and Lorenz, Dominik and Esser, Patrick and Ommer, Bj{\"o}rn},
  booktitle={Proceedings of the IEEE/CVF conference on computer vision and pattern recognition},
  pages={10684--10695},
  year={2022}
}

@inproceedings{podellsdxl,
  title={SDXL: Improving Latent Diffusion Models for High-Resolution Image Synthesis},
  author={Podell, Dustin and English, Zion and Lacey, Kyle and Blattmann, Andreas and Dockhorn, Tim and M{\"u}ller, Jonas and Penna, Joe and Rombach, Robin},
  booktitle={The Twelfth International Conference on Learning Representations}
}

@article{betker2023improving,
  title={Improving image generation with better captions},
  author={Betker, James and Goh, Gabriel and Jing, Li and Brooks, Tim and Wang, Jianfeng and Li, Linjie and Ouyang, Long and Zhuang, Juntang and Lee, Joyce and Guo, Yufei and others},
  journal={Computer Science. https://cdn. openai. com/papers/dall-e-3. pdf},
  volume={2},
  number={3},
  pages={8},
  year={2023}
}

@article{saharia2022photorealistic,
  title={Photorealistic text-to-image diffusion models with deep language understanding},
  author={Saharia, Chitwan and Chan, William and Saxena, Saurabh and Li, Lala and Whang, Jay and Denton, Emily L and Ghasemipour, Kamyar and Gontijo Lopes, Raphael and Karagol Ayan, Burcu and Salimans, Tim and others},
  journal={Advances in neural information processing systems},
  volume={35},
  pages={36479--36494},
  year={2022}
}

@article{li2024survey,
  title={A survey on LLM-based multi-agent systems: workflow, infrastructure, and challenges},
  author={Li, Xinyi and Wang, Sai and Zeng, Siqi and Wu, Yu and Yang, Yi},
  journal={Vicinagearth},
  volume={1},
  number={1},
  pages={9},
  year={2024},
  publisher={Springer}
}

@article{tran2025multi,
  title={Multi-agent collaboration mechanisms: A survey of llms},
  author={Tran, Khanh-Tung and Dao, Dung and Nguyen, Minh-Duong and Pham, Quoc-Viet and O'Sullivan, Barry and Nguyen, Hoang D},
  journal={arXiv preprint arXiv:2501.06322},
  year={2025}
}

@inproceedings{chen2023agentverse,
  title={Agentverse: Facilitating multi-agent collaboration and exploring emergent behaviors},
  author={Chen, Weize and Su, Yusheng and Zuo, Jingwei and Yang, Cheng and Yuan, Chenfei and Chan, Chi-Min and Yu, Heyang and Lu, Yaxi and Hung, Yi-Hsin and Qian, Chen and others},
  booktitle={The Twelfth International Conference on Learning Representations},
  year={2023}
}

@article{hegazy2024diversity,
  title={Diversity of thought elicits stronger reasoning capabilities in multi-agent debate frameworks},
  author={Hegazy, Mahmood},
  journal={arXiv preprint arXiv:2410.12853},
  year={2024}
}

@article{hu2025multi,
  title={Multi-Agent Debate for LLM Judges with Adaptive Stability Detection},
  author={Hu, Tianyu and Tan, Zhen and Wang, Song and Qu, Huaizhi and Chen, Tianlong},
  journal={arXiv preprint arXiv:2510.12697},
  year={2025}
}

@inproceedings{
zheng2026diffusionnft,
title={Diffusion{NFT}: Online Diffusion Reinforcement with Forward Process},
author={Kaiwen Zheng and Huayu Chen and Haotian Ye and Haoxiang Wang and Qinsheng Zhang and Kai Jiang and Hang Su and Stefano Ermon and Jun Zhu and Ming-Yu Liu},
booktitle={The Fourteenth International Conference on Learning Representations},
year={2026},
url={https://openreview.net/forum?id=VJZ477R89F}
}

@inproceedings{liuflow,
  title={Flow-GRPO: Training Flow Matching Models via Online RL},
  author={Liu, Jie and Liu, Gongye and Liang, Jiajun and Li, Yangguang and Liu, Jiaheng and Wang, Xintao and Wan, Pengfei and ZHANG, Di and Ouyang, Wanli},
  booktitle={The Thirty-ninth Annual Conference on Neural Information Processing Systems}
}

@article{cai2025z,
  title={Z-image: An efficient image generation foundation model with single-stream diffusion transformer},
  author={Cai, Huanqia and Cao, Sihan and Du, Ruoyi and Gao, Peng and Hoi, Steven and Hou, Zhaohui and Huang, Shijie and Jiang, Dengyang and Jin, Xin and Li, Liangchen and others},
  journal={arXiv preprint arXiv:2511.22699},
  year={2025}
}

@article{yang2025qwen3,
  title={Qwen3 technical report},
  author={Yang, An and Li, Anfeng and Yang, Baosong and Zhang, Beichen and Hui, Binyuan and Zheng, Bo and Yu, Bowen and Gao, Chang and Huang, Chengen and Lv, Chenxu and others},
  journal={arXiv preprint arXiv:2505.09388},
  year={2025}
}

@article{wu2025qwen,
  title={Qwen-image technical report},
  author={Wu, Chenfei and Li, Jiahao and Zhou, Jingren and Lin, Junyang and Gao, Kaiyuan and Yan, Kun and Yin, Sheng-ming and Bai, Shuai and Xu, Xiao and Chen, Yilei and others},
  journal={arXiv preprint arXiv:2508.02324},
  year={2025}
}

@misc{flux-2-2025,
    author={Black Forest Labs},
    title={{FLUX.2: Frontier Visual Intelligence}},
    year={2025},
    howpublished={\url{https://bfl.ai/blog/flux-2}},
}

@inproceedings{yeimgedit,
  title={ImgEdit: A Unified Image Editing Dataset and Benchmark},
  author={Ye, Yang and He, Xianyi and Li, Zongjian and Lin, Bin and Yuan, Shenghai and Yan, Zhiyuan and Hou, Bohan and Yuan, Li},
  booktitle={The Thirty-ninth Annual Conference on Neural Information Processing Systems Datasets and Benchmarks Track}
}

@misc{OpenAI_GPT4_2026,
  author = {{OpenAI}},
  title = {GPT-4.1},
  year = {2026},
  url = {https://chat.openai.com},
}

@misc{gemini2026pro,
  author = {{Google}},
  title = {{Gemini 3.1 Pro}},
  year = {2026},
  howpublished = {\url{https://gemini.google.com/}},
  note = {Accessed: April 28, 2026},
  type = {Large language model}
}

@inproceedings{esser2024scaling,
  title={Scaling rectified flow transformers for high-resolution image synthesis},
  author={Esser, Patrick and Kulal, Sumith and Blattmann, Andreas and Entezari, Rahim and M{\"u}ller, Jonas and Saini, Harry and Levi, Yam and Lorenz, Dominik and Sauer, Axel and Boesel, Frederic and others},
  booktitle={Forty-first international conference on machine learning},
  year={2024}
}

@article{wu2025chronoedit,
  title={Chronoedit: Towards temporal reasoning for image editing and world simulation},
  author={Wu, Jay Zhangjie and Ren, Xuanchi and Shen, Tianchang and Cao, Tianshi and He, Kai and Lu, Yifan and Gao, Ruiyuan and Xie, Enze and Lan, Shiyi and Alvarez, Jose M and others},
  journal={arXiv preprint arXiv:2510.04290},
  year={2025}
}

@article{labs2025flux,
  title={FLUX. 1 Kontext: Flow Matching for In-Context Image Generation and Editing in Latent Space},
  author={Labs, Black Forest and Batifol, Stephen and Blattmann, Andreas and Boesel, Frederic and Consul, Saksham and Diagne, Cyril and Dockhorn, Tim and English, Jack and English, Zion and Esser, Patrick and others},
  journal={arXiv preprint arXiv:2506.15742},
  year={2025}
}

@inproceedings{parmar2018image,
  title={Image transformer},
  author={Parmar, Niki and Vaswani, Ashish and Uszkoreit, Jakob and Kaiser, Lukasz and Shazeer, Noam and Ku, Alexander and Tran, Dustin},
  booktitle={International conference on machine learning},
  pages={4055--4064},
  year={2018},
  organization={PMLR}
}

@article{van2016conditional,
  title={Conditional image generation with pixelcnn decoders},
  author={Van den Oord, Aaron and Kalchbrenner, Nal and Espeholt, Lasse and Vinyals, Oriol and Graves, Alex and others},
  journal={Advances in neural information processing systems},
  volume={29},
  year={2016}
}

@article{goodfellow2020generative,
  title={Generative adversarial networks},
  author={Goodfellow, Ian and Pouget-Abadie, Jean and Mirza, Mehdi and Xu, Bing and Warde-Farley, David and Ozair, Sherjil and Courville, Aaron and Bengio, Yoshua},
  journal={Communications of the ACM},
  volume={63},
  number={11},
  pages={139--144},
  year={2020},
  publisher={ACM New York, NY, USA}
}

@inproceedings{karras2019style,
  title={A style-based generator architecture for generative adversarial networks},
  author={Karras, Tero and Laine, Samuli and Aila, Timo},
  booktitle={Proceedings of the IEEE/CVF conference on computer vision and pattern recognition},
  pages={4401--4410},
  year={2019}
}

@article{ho2020denoising,
  title={Denoising diffusion probabilistic models},
  author={Ho, Jonathan and Jain, Ajay and Abbeel, Pieter},
  journal={Advances in neural information processing systems},
  volume={33},
  pages={6840--6851},
  year={2020}
}

@inproceedings{nichol2021improved,
  title={Improved denoising diffusion probabilistic models},
  author={Nichol, Alexander Quinn and Dhariwal, Prafulla},
  booktitle={International conference on machine learning},
  pages={8162--8171},
  year={2021},
  organization={PMLR}
}

@article{song2020score,
  title={Score-based generative modeling through stochastic differential equations},
  author={Song, Yang and Sohl-Dickstein, Jascha and Kingma, Diederik P and Kumar, Abhishek and Ermon, Stefano and Poole, Ben},
  journal={arXiv preprint arXiv:2011.13456},
  year={2020}
}

@inproceedings{ramesh2021zero,
  title={Zero-shot text-to-image generation},
  author={Ramesh, Aditya and Pavlov, Mikhail and Goh, Gabriel and Gray, Scott and Voss, Chelsea and Radford, Alec and Chen, Mark and Sutskever, Ilya},
  booktitle={International conference on machine learning},
  pages={8821--8831},
  year={2021},
  organization={Pmlr}
}

@article{radford2015unsupervised,
  title={Unsupervised representation learning with deep convolutional generative adversarial networks},
  author={Radford, Alec and Metz, Luke and Chintala, Soumith},
  journal={arXiv preprint arXiv:1511.06434},
  year={2015}
}

@article{lipman2022flow,
  title={Flow matching for generative modeling},
  author={Lipman, Yaron and Chen, Ricky TQ and Ben-Hamu, Heli and Nickel, Maximilian and Le, Matt},
  journal={arXiv preprint arXiv:2210.02747},
  year={2022}
}

@inproceedings{peebles2023scalable,
  title={Scalable diffusion models with transformers},
  author={Peebles, William and Xie, Saining},
  booktitle={Proceedings of the IEEE/CVF international conference on computer vision},
  pages={4195--4205},
  year={2023}
}

@article{snell2024scaling,
  title={Scaling llm test-time compute optimally can be more effective than scaling model parameters},
  author={Snell, Charlie and Lee, Jaehoon and Xu, Kelvin and Kumar, Aviral},
  journal={arXiv preprint arXiv:2408.03314},
  year={2024}
}

@article{kirstain2023pick,
  title={Pick-a-pic: An open dataset of user preferences for text-to-image generation},
  author={Kirstain, Yuval and Polyak, Adam and Singer, Uriel and Matiana, Shahbuland and Penna, Joe and Levy, Omer},
  journal={Advances in neural information processing systems},
  volume={36},
  pages={36652--36663},
  year={2023}
}

@inproceedings{hessel2021clipscore,
  title={Clipscore: A reference-free evaluation metric for image captioning},
  author={Hessel, Jack and Holtzman, Ari and Forbes, Maxwell and Le Bras, Ronan and Choi, Yejin},
  booktitle={Proceedings of the 2021 conference on empirical methods in natural language processing},
  pages={7514--7528},
  year={2021}
}

@article{wu2023human,
  title={Human preference score v2: A solid benchmark for evaluating human preferences of text-to-image synthesis},
  author={Wu, Xiaoshi and Hao, Yiming and Sun, Keqiang and Chen, Yixiong and Zhu, Feng and Zhao, Rui and Li, Hongsheng},
  journal={arXiv preprint arXiv:2306.09341},
  year={2023}
}

@article{xu2023imagereward,
  title={Imagereward: Learning and evaluating human preferences for text-to-image generation},
  author={Xu, Jiazheng and Liu, Xiao and Wu, Yuchen and Tong, Yuxuan and Li, Qinkai and Ding, Ming and Tang, Jie and Dong, Yuxiao},
  journal={Advances in Neural Information Processing Systems},
  volume={36},
  pages={15903--15935},
  year={2023}
}

@article{wang2025unified,
  title={Unified reward model for multimodal understanding and generation},
  author={Wang, Yibin and Zang, Yuhang and Li, Hao and Jin, Cheng and Wang, Jiaqi},
  journal={arXiv preprint arXiv:2503.05236},
  year={2025}
}

@article{wang2025unigenbenchpp,
  title={UniGenBench++: A unified semantic evaluation benchmark for text-to-image generation},
  author={Wang, Yibin and Li, Zhimin and Zang, Yuhang and Bu, Jiazi and Zhou, Yujie and Xin, Yi and He, Junjun and Wang, Chunyu and Lu, Qinglin and Jin, Cheng and others},
  journal={arXiv preprint arXiv:2510.18701},
  year={2025}
}
